\documentclass[lettersize,journal]{IEEEtran}
\usepackage{amsmath,amsfonts}
\usepackage{algorithm}
\usepackage{array}
\usepackage[caption=false,font=normalsize,labelfont=sf,textfont=sf]{subfig}
\usepackage{textcomp}
\usepackage{stfloats}
\usepackage{url}
\usepackage{verbatim}
\usepackage{graphicx}
\usepackage{cite}

\usepackage{amssymb}
\usepackage[noend]{algpseudocode}
\usepackage{multirow}
\usepackage[switch]{lineno}
\usepackage{color}
\usepackage{pifont}       
\usepackage{svg}
\usepackage{subcaption}
\graphicspath{{./imgs/}{./figures/}}

\begin{document}

\title{TextDoctor: Unified Document Image Inpainting via Patch Pyramid Diffusion Models}

\author{Wanglong Lu, Lingming Su, Jingjing Zheng, Vinícius Veloso de Melo, Farzaneh Shoeleh, John Hawkin, \\Terrence Tricco, Hanli Zhao$^{*}$, Xianta Jiang
\thanks{W. Lu, L. Su, and H. Zhao are with the College of Computer Science and Artificial Intelligence, Wenzhou University, Wenzhou 325035, China.}
\thanks{J. Zheng is with the Department of Mathematics, University of British Columbia,  British Columbia, V6T 1Z2, Canada.}
\thanks{W. Lu, V. Melo, F. Shoeleh, and J. Hawkin are with the AI Analytics Team, Nasdaq, St. John's, A1A 0L9, Canada.}
\thanks{W. Lu, T. Tricco, and X. Jiang are with the Department of Computer Science, Memorial University of Newfoundland, St. John's, NL A1B 3X5, Canada.}

\thanks{$^{*}$ Corresponding author. E-mail: hanlizhao@wzu.edu.cn}
}




\maketitle

\begin{abstract}
Digital versions of real-world text documents often suffer from issues like environmental corrosion of the original document, low-quality scanning, or human interference. Existing document restoration and inpainting methods typically struggle with generalizing to unseen document styles and handling high-resolution images. To address these challenges, we introduce TextDoctor, a novel unified document image inpainting method. Inspired by human reading behavior, TextDoctor restores fundamental text elements from patches and then applies diffusion models to entire document images instead of training models on specific document types. To handle varying text sizes and avoid out-of-memory issues, common in high-resolution documents, we propose using structure pyramid prediction and patch pyramid diffusion models. These techniques leverage multiscale inputs and pyramid patches to enhance the quality of inpainting both globally and locally. 
Extensive qualitative and quantitative experiments on seven public datasets validated that TextDoctor outperforms state-of-the-art methods in restoring various types of high-resolution document images.
\end{abstract}

\begin{IEEEkeywords}
Document image inpainting, high-resolution image processing, patch pyramid diffusion models, denoising diffusion probabilistic models
\end{IEEEkeywords}

\section{Introduction}

\IEEEPARstart{D}{ocument} image inpainting focuses on restoring text information from degraded photographed or scanned documents that were compromised by environmental corrosion or human interference.
This task is challenging due to the diverse textures, structures, and sizes of text within a document, which can significantly impact downstream applications~\cite{PEN2021106722_KBS,SUN2022109210_KBS} and diminish the visual appeal and legibility of the documents~\cite{LIU2024112046_KBS,HUANG2023110346_KBS,HEDJAZI2021106789_KBS}.

Despite the impressive performance of existing the state-of-the-art (SOTA) document image restoration~\cite{DocRes,DocDiff} and text image inpainting methods~\cite{zhu2024gsdm,Sun_TSINIT}, several critical challenges remain largely unexplored. 
These include unified inpainting models for various document styles and efficient processing of high-resolution document images.
The above models trained on one type of document (e.g., checks) often struggle when applied to another type (e.g., forms). Additionally, these methods typically evaluate document image patches without considering the high-resolution nature of entire documents, which can lead to potential out-of-memory issues during full-document inference.

\begin{figure}[t]
	\centering
	\includegraphics[width=0.49\textwidth]{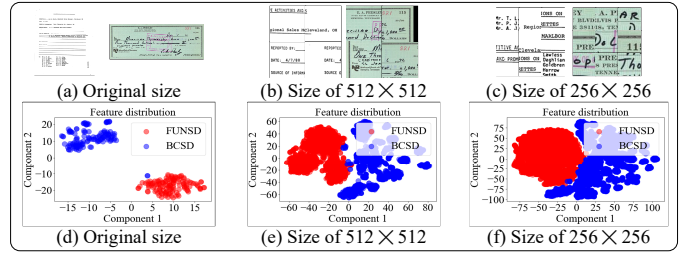}
	\caption{
    The motivation of unified high-resolution document image inpainting. 
    Fine-grain document patches share more visual similarities between FUNSD (a) (left) and BCSD (a) (right) datasets:
    (a) document images, (b) and (c) cropped patches; Fine-grain patches intersect more in feature space using t-SNE~\cite{TSNE} from FUNSD (\textcolor{red}{red}) and BCSD (\textcolor{blue}{blue}): (d) distributions of document images, (e) and (f) distributions of image patches.}\label{fig:fig_ideas}  
\end{figure}

\begin{figure}[t]
	\centering
	\includegraphics[width=0.49\textwidth]{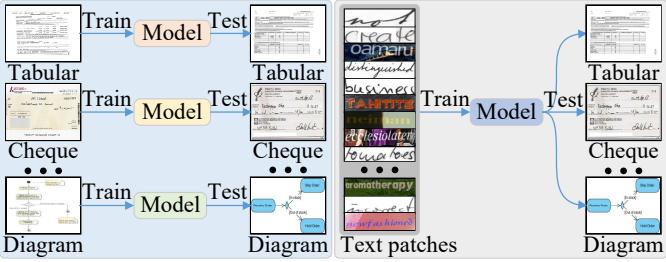}
	\caption{The training and testing pipelines of SOTA methods~\cite{DocDiff} (left) and TextDoctor (right). Once trained on the image patches, our TextDoctor can be applied to perform high-quality document image inpainting across various document types and resolutions. }\label{fig:fig_difference}  
\end{figure}

Human reading behavior~\cite{liu2005reading,reading2014} typically involves scanning individual text elements.
After learning to understand words or letters, humans can initially grasp the general context and then gain full comprehension by closely examining text elements. 
As shown in Fig.~\ref{fig:fig_ideas}, one observes that fine-grained patches containing basic text elements share similar visual patterns and feature distributions across different types of documents. 
Building on these observations, we explore a fundamental question in computer vision: Can we tackle the generalization and out-of-memory challenges in high-resolution document image inpainting, inspired by human reading behavior? 
As shown in Fig.~\ref{fig:fig_difference}, our approach involves training diffusion-based inpainting models to recover text elements from corrupted image patches and then applying them to entire document images. 
By leveraging prior knowledge of text elements, models can focus on text information and avoid overfitting visual patterns specific to particular documents.

However, three obstacles need to be addressed: First, as shown in Fig.~\ref{fig:fig_ideas} (d-f), the feature distribution of trained patches does not align well with document images, leading to suboptimal inpainting performance when directly applying the trained model to document images.
Second, the diverse text sizes, structures, textures, and colors in documents complicate the inpainting process due to models' limited receptive fields.
Third, inference on high-resolution document images can cause potential out-of-memory issues.

To address these challenges, we propose \textbf{TextDoctor}, a novel unified document image inpainting method, utilizing structure pyramid prediction and patch pyramid diffusion models to achieve high-quality inpainting across various document types and resolutions.
For the first challenge, our method employs patch-based inference to partition the document into a sequence of patches. 
TextDoctor restores these patches, which are then merged into the complete inpainted document image, ensuring consistent visual patterns between inputs from training and inference.
Second, by using multiscale inputs and pyramid patches, we enhance both global and local text and background information, 
improving the quality and fidelity of inpainted images.
Third, processing each small patch independently reduces memory usage, effectively preventing out-of-memory issues common in high-resolution images~\cite{DocDiff}.

To our knowledge, we are the first to develop a unified method for inpainting text within unseen high-resolution documents by taking advantage of patch pyramids. 
We demonstrate that TextDoctor is compatible with existing network architectures, including CNN-based~\cite{Zeng2021}, Transformer-based~\cite{lu2023grig,zhu2024gsdm}, and Mamba-based~\cite{ruan2024vm} networks.
Our paper makes four key contributions:
(1) We introduce a novel solution to address generalization and high-memory consumption challenges in high-resolution document image inpainting, inspired by human reading behavior.
(2) We propose \textbf{TextDoctor} that can perform effective inpainting across various document types and resolutions without additional training on unseen document types.
(3) We propose structure pyramid prediction and patch pyramid diffusion models to ensure effective restoration across various document types and resolutions.
(4) We conduct extensive experiments on seven document datasets using various network architectures, demonstrating that TextDoctor achieves competitive or superior performance compared to SOTA methods, even without fine-tuning or training on these datasets, unlike the SOTA approaches that require such adjustments.

\section{Related work}
\textbf{Document image restoration.}
Numerous methods have been proposed to address various types of degradation for natural image restoration~\cite{wang2024promptrr,WANG2024109956}. 
However, these methods lack prior knowledge of text information. 
To solve this problem, some studies designed to enhance the readability of document images by leveraging document-specific characteristics~\cite{DocRes,cicchetti2024naf}. 
Dewarping techniques~\cite{Das_2019_ICCV,Geometric_eccv2022} focus on eliminating geometric distortions such as curves and crumples. 
Deshadowing methods~\cite{Li_2023_ICCV,Lin_2020_CVPR} aim to remove shadows commonly found in photographed document images, resulting in clearer, shadow-free documents. 
Appearance enhancement~\cite{Appearance_2024}, also known as illumination correction, seeks to improve the overall lighting quality of document images. 
Deblurring techniques~\cite{DEGAN,Hradi2015ConvolutionalNN,DocDiff} target the removal of blurriness to restore image clarity, while binarization methods~\cite{YANG2024109989,souibgui2022docentr,KHAMEKHEMJEMNI2022108370} involve segmenting foreground text from document images, which is crucial for applications centered on text content that is often obscured by stains or weak contrast.

While these methods address specific aspects of document restoration, they often lack generalization across data domains and fail to resolve out-of-memory issues with high-resolution images.
In contrast, TextDoctor is trained on image patches and achieves high-quality inpainting on high-resolution images without fine-tuning for different document styles.

\begin{figure*}[t]
	\centering
    \includegraphics[width=1.0\textwidth]{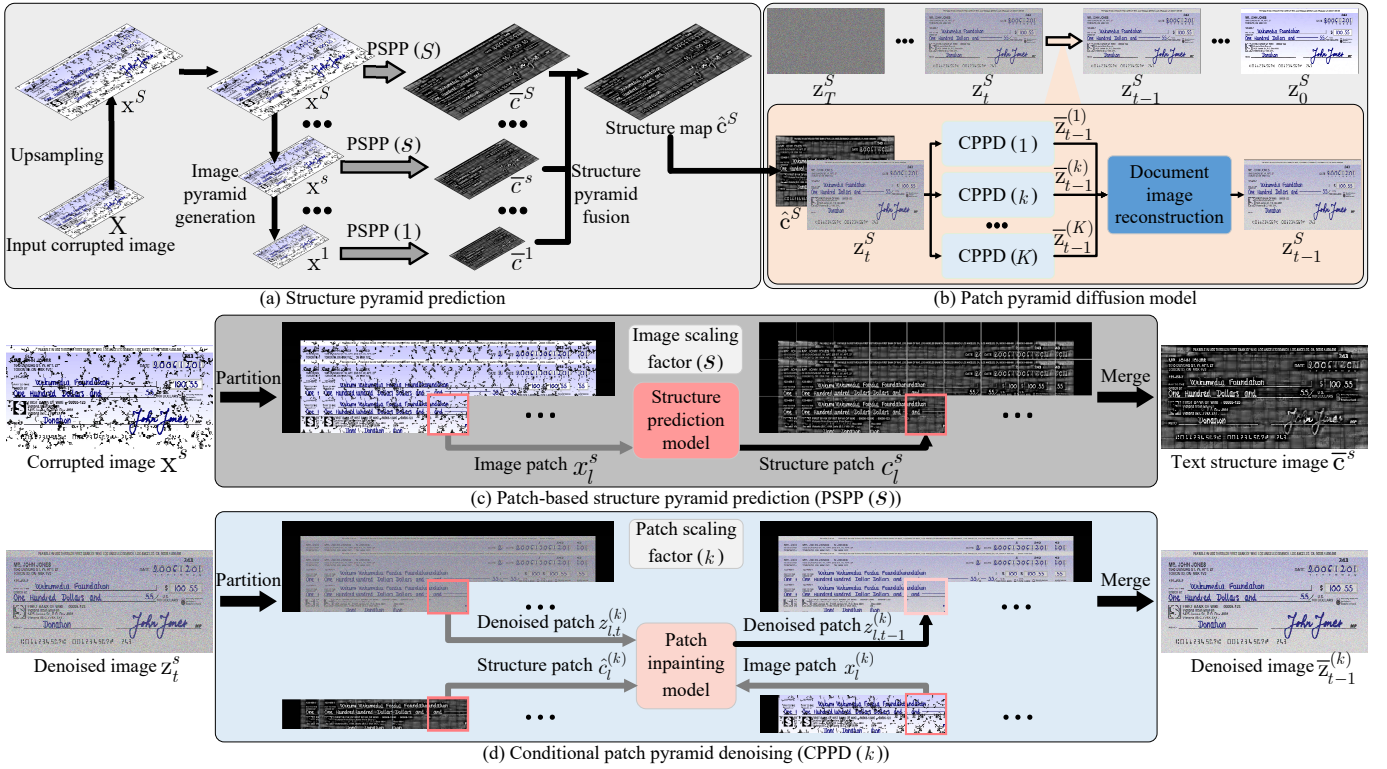}
	\caption{The inference pipeline of TextDoctor. (a) First, we use PSPP to perform structure pyramid prediction from an upsampled corrupted image. (b) Then, we utilize the predicted structures to guide the patch pyramid denoising process and get the inpainted image, using CPPD. 
   (c) Patch-based structure pyramid prediction (PSPP).
   (d) Conditional patch pyramid denoising  (CPPD).}\label{fig:fig_framework_full}  
\end{figure*}

\textbf{Document image inpainting.} 
Corrosion issues, such as graffiti and low-quality scanning, hinder text comprehension, highlighting the need for inpainting missing text parts.
Image inpainting is a longstanding challenge in computer vision~\cite{TensorSVD}, with numerous studies addressing high-quality~\cite{WANG2022102321,LIU2024112046_KBS}, diversity~\cite{lu2022inpainting}, editing~\cite{FACEMUG}, data efficiency~\cite{lu2023grig}, low resolution~\cite{HUANG2023110346_KBS}, and high resolution~\cite{yi2020contextual,liu2023coordfill} problems. While these approaches offer valuable insights for facial and natural images, they often lack prior knowledge, such as text structural information.
Thus, their inpainted images may look visually acceptable but fail to restore the semantic structure of the text~\cite{zhu2024gsdm}.

For text image inpainting~\cite{Sun_TSINIT,zhu2024gsdm} or restoration~\cite{liu2023textdiff}, incorporating text-related domain knowledge improves the restoration quality of corroded text images. 
However, these methods face significant challenges when applied to full document images.
One issue is the feature gap between training data (text image patches) and full document images, which affects performance. Moreover, document images often exceed 1K or 2K resolutions, causing out-of-memory problems on limited GPU resources. 
TextDoctor is designed to tackle these challenges for high-resolution document inpainting.

\section{Methodology}
\subsection{Overview}
Our unified document image inpainting algorithm is inspired by human reading behavior~\cite{liu2005reading,reading2014}, which captures visual features of documents in a coarse-to-fine manner.
Fig.~\ref{fig:fig_difference} (right) shows the training and testing pipeline of TextDoctor. 
It begins by training patch-based diffusion models to recover text elements (e.g., characters and digits) and non-text backgrounds from corrupted image patches, and then applies the patch-based inference of TextDoctor to the entire document image.

As shown in Fig.~\ref{fig:fig_framework_full}, we introduce \textit{structure pyramid prediction} to capture structural features at multiple scales, and \textit{patch pyramid diffusion models} to utilize these structural details for denoising the entire document using pyramid patches.
This approach enhances robustness across diverse document styles and effectively handles high-resolution document images avoiding out-of-memory issues.

\subsection{Proposed structure pyramid prediction}\label{sec:multiscale}
Direct inference on high-resolution images may struggle to capture various sizes of text due to limited receptive fields, potentially leading to out-of-memory issues.
Therefore, we first train a structure prediction model $f_{\bar{\theta}}(\cdot)$ to capture the text structure prior from corrupted patches.
Then, we employ our novel patch-based structure pyramid prediction (PSPP) for entire corrupted documents using $f_{\bar{\theta}}(\cdot)$.


As shown in Fig.~\ref{fig:fig_framework_full} (a), our {structure pyramid prediction} includes three key steps to capture structures by processing on multiple scales:
(1) {Image pyramid generation}: The document image is first resized into several candidate images at various scales.
(2) {Patch-based structure pyramid prediction (PSPP)}: For each scale, the high-resolution image is split into smaller patches, which are processed individually by the structure prediction model $f_{\bar{\theta}}(\cdot)$. All processed patches are then merged to reconstruct a candidate structure map. 
(3) {Structure pyramid fusion}: We aggregate all the predicted structure maps with various scales at the scale $S$.

\textbf{Image pyramid generation.}
As shown in Fig.~\ref{fig:fig_framework_full} (a), given a document image $\mathbf{x}^{\bar{s}} \in \mathbb{R}^{3\times (\bar{s}\times 
u)\times (\bar{s}\times v)} = \mathbf{X} \in \mathbb{R}^{3\times H\times W}$, we upsample and downsample (e.g., bicubic) the input to get a sequence of images $[\mathbf{x}^1,\ldots,\mathbf{x}^s,\ldots,\mathbf{x}^S]$, where $\mathbf{x}^1 \in \mathbb{R}^{3\times  u \times v}$; $s \in [1,\ldots,S]$; $s$ is the image scaling factor for the side length.  $\bar{s}$ is the original scale factor of the $\mathbf{X}$.
$m = \lfloor S / \bar{s} \rfloor$ is the upsampling scale factor;
Thus, we have multiscale inputs to emphasize text with various sizes.

\textbf{Patch structure prediction.} 
Given a degraded text image patch $\boldsymbol{x} \in \mathbb{R}^{3\times h\times w}$, our structure prediction model $f_{\bar{\theta}}(\cdot)$ predicts the structure information, denoted as $\boldsymbol{c} \in \mathbb{R}^{1\times h\times w} = f_{\bar{\theta}}(\boldsymbol{x})$. To train $f_{\bar{\theta}}(\cdot)$~\cite{Sun_TSINIT,zhu2024gsdm}, we apply several loss functions (L1, binary-cross entropy, perceptual content, and style loss functions) to constrain the pixel and semantic similarities.

\textbf{Patch-based structure pyramid prediction.}
As shown in Fig.~\ref{fig:fig_framework_full} (c), we partition each corrupted image $\mathbf{x}^s \in \mathbb{R}^{3\times (s\times u)\times (s\times v)}$ into partially overlapped patches as:
\begin{equation}\label{equ:partition}
        \begin{aligned}
        \mathcal{X}^s &= [\boldsymbol{x}^s_1,\ldots,\boldsymbol{x}^s_l,\ldots,\boldsymbol{x}^s_L] = \mathcal{S}_{split}(\mathbf{x}^s), \boldsymbol{x}^s_l \in \mathbb{R}^{3\times h\times w},
        \end{aligned}
\end{equation}
where  $ \mathcal{S}_{split}(\cdot)$ is a function, which performs shifted crop sampling on image space; $L = \left(\left\lfloor \frac{(s\times u)-h}{d_h} \right\rfloor+ 1\right) \times \left(\left\lfloor\frac{(s\times v)-w}{d_w} \right\rfloor + 1\right)$, $d_h$ and $d_w$ represent the stride in the vertical and horizontal directions, respectively. 
We apply the patch structure prediction model $f_{\bar{\theta}}(\cdot)$ to each patch $ \mathcal{C}^s= [\boldsymbol{c}^s_1,\ldots,\boldsymbol{c}^s_l,\ldots,\boldsymbol{c}^s_L] = [f_{\bar{\theta}}(\boldsymbol{x}^s_1),\ldots,f_{\bar{\theta}}(\boldsymbol{x}^s_l),\ldots,f_{\bar{\theta}}(\boldsymbol{x}^s_L)] = \mathcal{D}(\mathcal{X}^s)$. The $\boldsymbol{c}^s_l \in \mathbb{R}^{1\times h\times w}$ and $\boldsymbol{x}^s_l \in \mathbb{R}^{3\times h\times w}$ are the predicted structure patch and corrupted image patch, respectively.

To aggregate all predicted structure information at the scale $s$, a merge operation $\mathcal{R}(\cdot)$ is applied by averaging the overlapped patch regions to reconstruct the large-scale structure map, as  $\bar{\mathbf{c}}^s = \mathcal{R}(\mathcal{C}^s) \in \mathbb{R}^{1\times (s\times u)\times (s\times v)}$.

\textbf{Structure pyramid fusion.}
To restore document images, we utilize structural information from different scales, integrating both large and small text elements into the final high-resolution structure map.
As shown in Fig.~\ref{fig:fig_framework_full} (a), given predicted structure maps from various resolutions $\bar{\mathbf{c}}^1\ldots,\bar{\mathbf{c}}^s,\ldots,\bar{\mathbf{c}}^S$, the structure pyramid fusion is defined as $\hat{\mathbf{c}}^S = \frac{1}{S} \sum_{s=1}^S \operatorname{IP}_S(\bar{\mathbf{c}}^s)$,
where $\operatorname{IP}_S(\cdot)$ is an interpolation function (e.g., bicubic) that resizes the image size to align with $\bar{\mathbf{c}}^S$; 
We set $\hat{\mathbf{c}}^S$ as the final predicted structure map.

\subsection{Proposed patch pyramid diffusion models}\label{sec:pyramid}

To ensure generalization across different document styles, effectively capture multi-size text information, and prevent out-of-memory issues, we design our unified document image inpainting via patch pyramid diffusion models.

First, we train a patch-based diffusion inpainting model $g_\theta(\cdot)$ using the predicted structures to perform the denoising process to subtract noise at each denoising step. 
As shown in Fig.~\ref{fig:fig_framework_full} (b), we design two main steps to inpaint the entire corrupted document image:
(1) {Conditional patch pyramid denoising (CPPD)}:  The input images are divided into smaller patches using 
pyramid patches, controlled by patch scaling factor $k$. 
For each factor $k$, denoising is performed using conditional information.
(2) {Document image reconstruction}: The images denoised from pyramid patches are averaged to reconstruct the final complete document image.

\textbf{Patch inpainting using diffusion models.} 
Based on diffusion models (DMs)~\cite{NEURIPS2020DDPM}, $g_{\theta}(\cdot)$ utilizes conditional information to get a denoised text image patch.
DMs involve a forward diffusion process and a reverse denoising process.
The diffusion process incrementally adds Gaussian noise to the clean text image patch $\boldsymbol{z}_0 \in \mathbb{R}^{3\times h\times w} $ and corrupts it into an approximately pure Gaussian noise $\boldsymbol{z}_T$ using a variance schedule $\beta_1,\ldots, \beta_T$: $q(\boldsymbol{z}_{1:T}|\boldsymbol{z}_0) = \prod_{t=1}^{T} q(\boldsymbol{z}_t|\boldsymbol{z}_{t-1},\boldsymbol{z}_0 )$. Each step is defined as: $q(\boldsymbol{z}_t|\boldsymbol{z}_{t-1},\boldsymbol{z}_0) = \mathcal{N}(\boldsymbol{z}_t; \sqrt{1 - \beta_{t}} \boldsymbol{z}_{t-1}, \beta_t \mathbf{I}),$
where $\boldsymbol{z}_t$ can be approximated by $\boldsymbol{z}_t=\sqrt{\bar{\alpha}_t} \boldsymbol{z}_0+ \sqrt{1-\bar{\alpha}_t} \boldsymbol{\epsilon}$, with $\bar{\alpha}_t=\prod_{s=1}^t \alpha_s$, $\alpha_t=1-\beta_t$, and $\boldsymbol{\epsilon} \sim \mathcal{N}(\mathbf{0}, \mathbf{I})$. 

Conditioned on the corrupted image patch $\boldsymbol{x}$ and predicted text structure $\boldsymbol{c}$, the reverse denoising process aims to recover the cleaner version $\boldsymbol{z}_{t-1}$ from $\boldsymbol{z}_{t}$ by estimating the noise using learned Gaussian transition: $p_{\theta}(\boldsymbol{z}_{0:T} | \boldsymbol{x},\boldsymbol{c}) = p(\boldsymbol{z}_T) \prod_{t=1}^{T} p_{\theta}(\boldsymbol{z}_{t-1} | \boldsymbol{z}_t,\boldsymbol{x},\boldsymbol{c})$.
We followed the deterministic reverse process~\cite{song2020denoising} with zero variance. Each denoising step is expressed as:
\begin{equation}\label{equ:denoise_process}
    \begin{aligned}
        p_{\theta}(\boldsymbol{z}_{t-1}|\boldsymbol{z}_t,\boldsymbol{x},\boldsymbol{c}) &= q(\boldsymbol{z}_{t-1}|\boldsymbol{z}_t,g_\theta(\boldsymbol{z}_t, t, \boldsymbol{x},\boldsymbol{c})).
    \end{aligned}
\end{equation}
Given the trained  $g_{\theta}$, the denoising step is implemented as:
\begin{equation}\label{equ:dm_infer1}
	\begin{aligned}
        \boldsymbol{z}_{t-1} &= \sqrt{\bar{\alpha}_{t-1}}\hat{\boldsymbol{z}}_{0}+ \sqrt{1 - \bar{\alpha}_{t-1}} \frac{\boldsymbol{z}_{t} - \sqrt{\bar{\alpha}_t} \hat{\boldsymbol{z}}_{0}}{\sqrt{1-\bar{\alpha}_t}},
    \end{aligned}
\end{equation}
where $\hat{\boldsymbol{z}}_{0} = g_\theta(\boldsymbol{z}_t, t, \boldsymbol{x},\boldsymbol{c})$.

\textbf{Conditional patch pyramid denoising.}
The patch-based inference has been used in high-resolution image generation to prevent out-of-memory issues~\cite{MultiDiffusion,Du_2024_CVPR,lee2023syncdiffusion}.
However, these methods rely on fixed patch sizes. In this work, we extend their fixed-size patches to multiscale patches by taking advantage of the patch pyramid, enabling high-fidelity restoration of text elements across varying sizes.

As shown in Fig.~\ref{fig:fig_framework_full} (d), we utilize patch-based inference by incorporating structural information and corrupted images, ensuring a better understanding of the text structures. 
For the denoising step $t$ with a denoised document image $\mathbf{z}_t^S \in \mathbb{R}^{3\times (S\times u)\times (S\times v)}$, a corrupted document image $\mathbf{x}^S \in \mathbb{R}^{3\times (S\times u)\times (S\times v)}$, and a predicted structure map $\hat{\mathbf{c}}^S \in \mathbb{R}^{1\times (S\times u)\times (S\times v)}$ with the scale $S$, we first partition these images into partially overlapped patches  as:
\begin{equation}\label{equ:multi_partition}
        \begin{aligned}
        \mathcal{Z}_t^{(k)} &= [\boldsymbol{z}^{(k)}_{1,t},\ldots,\boldsymbol{z}^{(k)}_{l,t},\ldots,\boldsymbol{z}^{(k)}_{L,t}] = \mathcal{S}^{(k)}_{split}(\mathbf{z}_t^S), \\
        \mathcal{X}^{(k)} &= [\boldsymbol{x}^{(k)}_1,\ldots,\boldsymbol{x}^{(k)}_l,\ldots,\boldsymbol{x}^{(k)}_L] = \mathcal{S}^{(k)}_{split}(\mathbf{x}^S),\\
        \hat{\mathcal{C}}^{(k)} &= [\hat{\boldsymbol{c}}^{(k)}_1,\ldots,\hat{\boldsymbol{c}}^{(k)}_l,\ldots,\hat{\boldsymbol{c}}^{(k)}_L] = \mathcal{S}^{(k)}_{split}(\hat{\mathbf{c}}^S), \\
        \end{aligned}
\end{equation}
where $k \in [1,\ldots,K]$; $k$ is the patch scaling factor to control the patch size for cropping;
$\boldsymbol{z}^{(k)}_{l,t} \in \mathbb{R}^{3\times  (k\times a)\times (k\times b)}$, $\boldsymbol{x}^{(k)}_l \in \mathbb{R}^{3\times (k\times a)\times (k\times b)}$, and $\hat{\boldsymbol{c}}^{(k)}_l \in \mathbb{R}^{1\times (k\times a)\times (k\times b)}$, $L = \left(\left\lfloor \frac{(S\times u)-(k\times a)}{k \times d_a} \right\rfloor+ 1\right) \times \left(\left\lfloor\frac{(S\times v)-(k\times b)}{k \times d_b} \right\rfloor + 1\right)$, $k \times d_a$ and $k \times d_b$ represent the stride in the vertical and horizontal directions, respectively. 
The corresponding denoised patches for each time step $t$ can be expressed as: 
\begin{equation}\label{equ:patch_restore}
        \begin{aligned}
        \mathcal{Z}^{(k)}_{t-1} &=  \mathcal{P}(\mathcal{Z}^{(k)}_{t-1} | \mathcal{Z}^{(k)}_t,\mathcal{X}^{(k)},\hat{\mathcal{C}}^{(k)})\\
        &= [\boldsymbol{z}^{(k)}_{1,t-1},\ldots,\boldsymbol{z}^{(k)}_{l,t-1},\ldots,\boldsymbol{z}^{(k)}_{L,t-1}]\\
         &= [\ldots,p_{\theta}(\boldsymbol{z}^{(k)}_{l,t-1}|\boldsymbol{z}^{(k)}_{l,t},\boldsymbol{x}^{(k)}_l,\hat{\boldsymbol{c}}^{(k)}_l),\ldots], l \in [1,L]. \\
        \end{aligned}
\end{equation}
To get the denoised image at the patch scaling factor $k$ and step $t$, the merge operation is applied again to reconstruct the large-scale document image, as $\bar{\mathbf{z}}^{(k)}_{t-1} = \mathcal{R}(\mathcal{Z}^{(k)}_{t-1})$.

\textbf{Document image reconstruction.}
As shown in Fig.~\ref{fig:fig_framework_full} (b), after inference of all the patch scales, we can get $K$ denoised maps, $\bar{\mathbf{z}}^{(k)}_{t-1}, k \in [1,\ldots, K]$. Each patch scale can capture different sizes of text information. We fuse these maps to get the reconstructed image at time step $t$, which is defined as ${\mathbf{z}}^S_{t-1}= \frac{1}{K} \sum_{k=1}^{K} \bar{\mathbf{z}}^{(k)}_{t-1}$.
After completing $T$ steps of patch pyramid denoising, the denoised image ${\mathbf{z}}^S_{0}$ is generated. 
The inpainted image is obtained by resizing it back to its original dimensions, obtained as $\hat{\mathbf{X}} \in \mathbb{R}^{3\times H\times W} = \mathbf{z}^{\bar{s}}_{0} \in \mathbb{R}^{3\times (\bar{s}\times 
u)\times (\bar{s}\times v)} = \operatorname{IP}_{\bar{s}}({\mathbf{z}}^S_{0})$.

\section{Experiments and results}
This section presents qualitative and quantitative results, ablation studies, and applications. More implementation details, experimental results, and Wilcoxon signed-rank tests are available in the supplementary document.
\subsection{Experimental setting}
\textbf{Datasets.} We conducted experimental evaluations on seven document image datasets consisting of various document forms, including, FUNSD~\cite{jaume2019} (tabular documents), BCSD~\cite{khan2021novel} (cheque images), and five document subsets from the Roboflow 100 dataset~\cite{ciaglia2022roboflow}, such as Twitter posts, Twitter profiles, activity diagrams, signatures, and tabular documents.
We combined the training sets of the TII-ST and TII-HT datasets~\cite{zhu2024gsdm} as our text image patch dataset for training our TextDoctor.

\textbf{Compared methods.} We compared against SOTA image inpainting (AOT-GAN (AOT)~\cite{Zeng2021} and GRIG~\cite{lu2023grig}), text image inpainting (GSDM~\cite{zhu2024gsdm}), document image restoration (DE-GAN~\cite{DEGAN}, DocDiff~\cite{DocDiff}, DocRes~\cite{DocRes}), and the large-scale text image editing methods (AnyText~\cite{tuo2024anytext}). 
GSDM and TextDoctor were trained on the same text image patch dataset, while the other methods were trained on document datasets, following their default settings. 
All methods were compared on the testing sets of document image datasets.

\begin{figure*}[!ht]
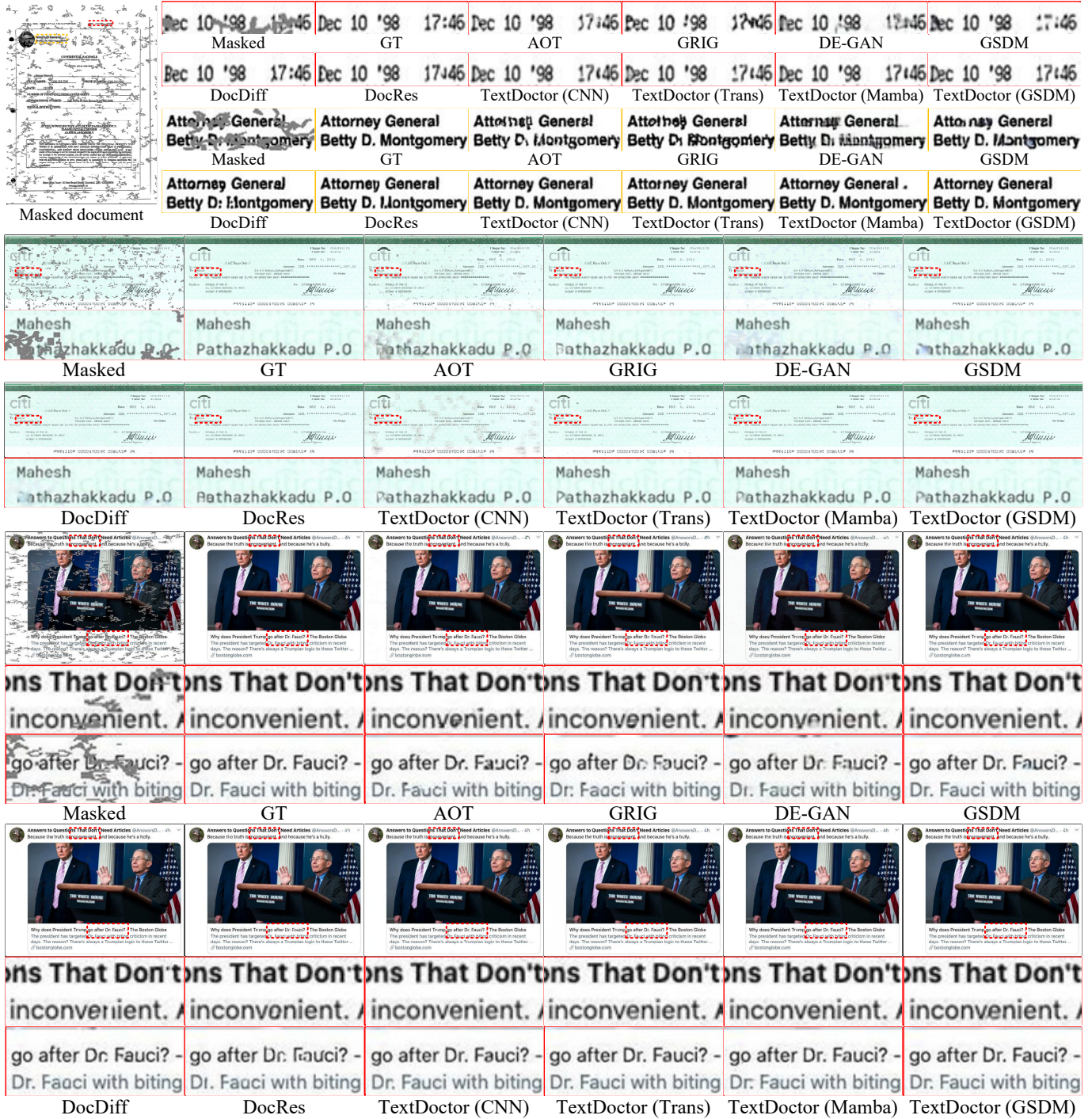

	\centering
	\includegraphics[width=1.0\textwidth]{fig_vis_compare_funsd}
        \includegraphics[width=1.0\textwidth]{fig_vis_compare_BCSD}
        \includegraphics[width=1.0\textwidth]{fig_vis_compare_tweet_profile}
	\caption{Visual comparisons to the SOTA document restoration and inpainting methods. }\label{fig:fig_vis_compare_funsd}  
\end{figure*}


\textbf{Implementation.}
We implemented our patch structure prediction model $f_{\bar{\theta}}(\cdot)$ and our inpainting model $g_{\theta}(\cdot)$ using existing networks, including CNN-based (AOT), Transformer-based (GRIG, GSDM), and Mamba-based~\cite{ruan2024vm} networks. For each implementation, we used the same network for both models to obtain a variant of our TextDoctor.
We used additional time step embeddings~\cite{StableDiffusion} for the inpainting model.
Unless specified, we used an upsampling scale factor of $m=4$ to capture detailed text information, while limiting the resolution to not exceed 4K to maintain efficient inference.
For the structure pyramid prediction, we set $u=256$, $v=256$, $h=256$, $w=256$, $d_h=128$, and $d_w=128$.
For the patch pyramid diffusion models, we set $a=128$, $b=128$, $d_a=64$, $d_b=64$, $K=2$, and $T=1$ for better efficiency.
We used $64$ and $16$ mini-batch patches for the patches sized $128^2$ ($k=1$) and $256^2$ ($k=2$), respectively.

\textbf{Evaluation metrics.} 
We used Peak Signal-to-Noise Ratio (PSNR) (dB) and Structural Similarity (SSIM) to assess pixel-level similarity. For evaluating perceptual similarity in high-resolution images, we employed the patch-based FID score~\cite{Heusel2017} (FID$_{\text{crop}}$) and patch-based U-IDS~\cite{Zhao2021} (U-IDS$_{\text{crop}}$). We utilized CRNN~\cite{CRNN}, MORAN~\cite{MORAN}, and PaddleOCR~\cite{PaddleOCR} to perform text detection and recognition, assessing the character-level (Char Acc) and word-level (Word Acc)~\cite{zhu2024gsdm} recognition accuracy of the inpainted images.

\subsection{Comparison to state-of-the-art algorithms}
\textbf{Qualitative comparison.}
Fig.~\ref{fig:fig_vis_compare_funsd} visualizes the cropped inpainting results of the compared methods. AOT and GRIG show good performance in inpainting background images but lack focus on text edges and structures, resulting in suboptimal outcomes. 
DocDiff and DocRes obtain clearer text information but still struggle with accurately inpainting text details due to the lack of text structure priors, despite being designed to focus on text edges in document image patches. 
GSDM and TextDoctor, trained on aligned text image patches, exhibit superior performance in capturing the structure of inpainted text. 
However, GSDM faces limitations when directly applied to high-resolution images, as the input size is restricted, and various text sizes increase the complexity of inpainting.
Our TextDoctor effectively inpaints patches with varying text sizes. 
We demonstrate that all the TextDoctor variants can achieve high-quality unified inpainting on unseen documents.


\begin{figure}[!ht]
	\centering
	\includegraphics[width=0.49\textwidth]{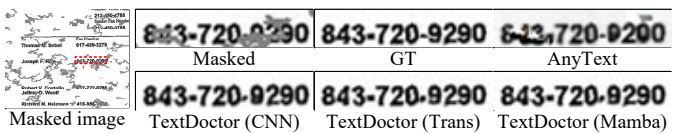}
	\caption{Visual comparisons to AnyText~\cite{tuo2024anytext} (Text prompts: A text line in a document with the words ``843-720-9290''). }\label{fig:fig_vis_compare_anytext}  
\end{figure}

\begin{table*}[!htbp]
    \centering
    \caption{Comparison with state-of-the-art methods on FUNSD dataset. \textbf{Bold} indicates our methods outperform SOTA methods.}
    \label{tab:FUNSD_comparison}
    \resizebox{\textwidth}{!}{%
        \begin{tabular}{c|cc|cc|cc|cc|cc}
            \hline
            \multirow{2}{*}{Method} &  \multicolumn{2}{c|}{Pixel-level}   & \multicolumn{2}{c|}{Perceptual-level} 
 & \multicolumn{2}{c|}{CRNN} & \multicolumn{2}{c|}{MORAN} & \multicolumn{2}{c}{PaddleOCR} \\
            \cline{2-11}
              & PSNR$^{\uparrow}$ & SSIM$^{\uparrow}$ & FID$_{\text{crop}}^{\downarrow}$ & U-IDS$_{\text{crop}}^{\uparrow}$ &  Char Acc$^{\uparrow}$ & Word Acc$^{\uparrow}$ &  Char Acc$^{\uparrow}$ & Word Acc$^{\uparrow}$ &  Char Acc$^{\uparrow}$ & Word Acc$^{\uparrow}$ \\
            \hline
            Input&16.22&0.7572&152.98&0.00\%&50.94\%&13.68\%&43.87\%&12.35\%&73.84\%&30.24\%\\
            \hline
            AOT (TVCG 2023) &27.75&0.9858&9.77&1.00\%&52.72\%&14.79\%&46.63\%&13.68\%&80.27\%&37.16\%\\
            GRIG (ArXiv 2023)&28.44&0.9889&3.17&12.40\%&52.79\%&15.27\%&46.79\%&13.89\%&80.70\%&37.68\%\\
            DE-GAN (TPAMI 2022)&29.03&0.9719&32.90&0.00\%&52.13\%&14.41\%&46.29\%&13.98\%&80.00\%&37.53\%\\
            DocDiff (ACM MM 2023)&31.60&0.9938&6.34&0.43\%&55.33\%&17.62\%&49.27\%&15.44\%&84.08\%&45.24\%\\
            DocRes (CVPR 2024)&32.37&0.9941&6.14&0.29\%&54.73\%&16.72\%&48.40\%&14.54\%&83.31\%&43.63\%\\
            GSDM (AAAI 2024) & 28.43 & 0.9855 & 7.00 & 0.00\% & 51.28\% & 14.16\% & 45.72\% & 13.21\% & 80.63\% & 39.63\% \\
            \hline
            TextDoctor (CNN)&29.88&0.9301&78.70&0.00\%&\textbf{55.94\%}&\textbf{17.98\%}&\textbf{49.80\%}&\textbf{16.42\%}&\textbf{86.74\%}&\textbf{53.03\%}\\
            TextDoctor (Trans)&\textbf{34.20}&\textbf{0.9958}&7.44&0.00\%&\textbf{56.23\%}&\textbf{18.14\%}&\textbf{49.86\%}&\textbf{16.04\%}&\textbf{86.14\%}&\textbf{52.18\%}\\
            TextDoctor (Mamba)&\textbf{35.04}&\textbf{0.9966}&\textbf{0.78}&\textbf{21.37\%}&\textbf{56.48\%}&\textbf{18.10\%}&\textbf{49.98\%}&\textbf{16.04\%}&\textbf{86.66\%}&\textbf{53.61\%}\\
            TextDoctor (GSDM)&\textbf{35.44}&0.9907&11.84&0.00\%&\textbf{56.97\%}&\textbf{19.17\%}&\textbf{50.28\%}&\textbf{16.64\%}&\textbf{86.92\%}&\textbf{55.38\%}\\
            \hline
        \end{tabular}
    }
\end{table*}

\begin{figure*}[htbp]
    \centering
    \includegraphics[width=1.0\textwidth]{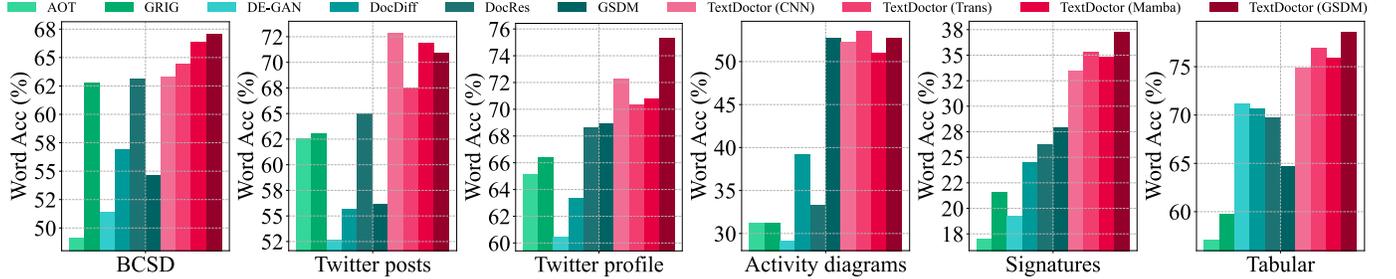}
    \caption{Quantitative comparisons on the other six document image datasets.}
    \label{fig:fig_six_vis_res}
\end{figure*} 

Fig.~\ref{fig:fig_vis_compare_anytext} clearly shows the superior performance of TextDoctor to the large-scale text image editing method AnyText. Our approach can precisely restore text information in the entire masked document image, while AnyText shows limitations despite being given clear text prompts for generation.

\textbf{Quantitative comparison.}
Table~\ref{tab:FUNSD_comparison} presents the quantitative results of the compared methods on the FUNSD dataset. AOT and GRIG achieve good perceptual-level similarity. For instance, GRIG obtains notable results with a FID$_\text{crop}$ of 3.17. 
However, as non-blind methods, requiring the position of masked regions hinders their practical application in real-world scenarios.
DocDiff and DocRes excel in pixel-level similarity but have limitations in understanding the semantic structure of text information.
GSDM and TextDoctor have more text prior knowledge. 
However, direct inference on whole documents does not guarantee good performance due to large domain gaps between trained patches and high-resolution images. 
The results show that TextDoctor is compatible with applied network architectures. 
TextDoctor variants show superior performance by large margins. 
For instance, TextDoctor (GSDM) outperforms the second-best method (DocDiff) significantly in terms of Word ACC (PaddleOCR), with a 10.14\% absolute improvement (55.38\% vs. 45.24\%).

Fig.~\ref{fig:fig_six_vis_res} shows the Word ACC (\%) of compared methods on the other six datasets. 
All the methods can achieve good performance.
However, they have to train each model when applied to another format of document images.
In contrast, TextDoctor variants achieve superior performance on these datasets.
Once trained on the text patch dataset, TextDoctor generalizes well across various unseen document datasets without requiring fine-tuning.
We conducted a Wilcoxon signed-rank test for these seven datasets, validating that our method performs significantly better.


\begin{table}[ht]
    \centering
    \caption{Comparison of inference times (in second) at various resolutions. \textbf{OOM}: Out-of-memory.}
    \resizebox{0.49\textwidth}{!}{%
    \begin{tabular}{c|c|c|c|c|c}
    \hline
    \multirow{2}{*}{Method} &  \multicolumn{5}{|c}{Resolution}\\
   \cline{2-6}
        &  $512^2$    &  $1024^2$   &  $2048^2$   &  $4096^2$ &  $8192^2$ \\ 
           \hline
    AOT       & 0.0329  & 0.1190    & 0.4651   & \textbf{OOM}   & \textbf{OOM}   \\ 
    GRIG      & 0.0337  & 0.0730   & 0.2717   & 1.0753 & \textbf{OOM}   \\ 
    DE-GAN    & 0.0085 & 0.0396  & 0.1645   & \textbf{OOM}   & \textbf{OOM}   \\ 
    DocDiff  & 0.0142   & 0.0719   & 0.3030    & \textbf{OOM}   & \textbf{OOM}   \\ 
    DocRes     & 0.1250      & 0.5208   & \textbf{OOM}     & \textbf{OOM}   & \textbf{OOM}   \\ 
    GSDM      & 0.0950  & 0.3559    & \textbf{OOM}     & \textbf{OOM}   & \textbf{OOM}   \\ \hline
    TextDoctor (CNN)  & 0.2703 & 1.3889    &  6.2458     &  26.5908 &    109.8388    \\ 
    TextDoctor (Trans)  &  0.1852   & 0.8333  &    3.5714    &  14.2857 &  59.8417       \\ 
    TextDoctor (Mamba) & 0.2933 &  1.4925  &   6.6667     &  28.6230 &  117.9581    \\ 
    TextDoctor (GSDM)   & 0.4367 & 2.1277 &  9.0909 &   39.8066    &    163.3987      \\ \hline
    \end{tabular}
    }
\label{tab:resource_comparison}
\end{table}


\textbf{Comparison on efficiency and supported resolutions.}
Table~\ref{tab:resource_comparison} shows the comparison of inference times to SOTA methods at various resolutions on an NVIDIA RTX 3090 GPU (24GB).
While SOTA methods process the entire document image at once, TextDoctor splits the image into patches, enabling ultra high-resolution inpainting (up to 8K) with limited GPU memory.
Furthermore, our structure pyramid prediction and patch pyramid diffusion models are utilized to capture global structures and local details.
Consequently, our TextDoctor effectively fulfills the task of inpainting on high-resolution document images with a little latency.

\subsection{Ablation study}
To evaluate the impact of each component in TextDoctor, we used a Mamba-based network for both structure prediction (ST) and inpainting models (IN) and tested it on the FUNSD dataset.
We progressively added each component to models and assessed their performance. 
We used PaddleOCR to evaluate Char Acc and Word Acc for all ablation studies.

\begin{table}[t]
    \centering
    \caption{Ablation study of the patch-based inference. \ding{55} and \ding{51} indicate the unplugging and plugging of this scheme on the structure prediction (ST) and inpainting (IN) models, respectively. \textbf{Bold}: Top-1 and \underline{Underline}: Top-2. }
    \label{tab:patch_infer_ablation}
        \resizebox{0.49\textwidth}{!}{%
        \begin{tabular}{c|c|c|c|c|c|c|c}
            \hline
             Scale &Config. & ST  & IN  & PSNR$^{\uparrow}$ & FID$_{\text{crop}}^{\downarrow}$&  Char Acc$^{\uparrow}$ & Word Acc$^{\uparrow}$  \\
            \hline
            \multirow{4}{*}{{$m=1$}}&A &  \ding{55}  &\ding{55} & 27.45 & 20.66 & 78.77\% & 36.42\%   \\
            &B &  \ding{51}  & \ding{55} & 27.63 & 27.40  & 77.46\% & 34.59\%     \\
            &C & \ding{55}  & \ding{51}&  \underline{29.02} &   \textbf{6.49} & \textbf{79.47\%}  &\textbf{37.09\%} \\
            &D & \ding{51} & \ding{51} & \textbf{29.06} & \underline{6.57} & \underline{79.42\%}  & \underline{36.70\%}   \\
            \hline
           \multirow{4}{*}{{$m=2$}}& A &  \ding{55}  &\ding{55} &  27.46 & 20.81 & 78.46\% & 35.97\%  \\
            &B &  \ding{51}  & \ding{55} & 32.02 & 17.82  & 83.81\% & \underline{46.08\%}   \\
            &C & \ding{55}  & \ding{51}&  \underline{32.41} &   \underline{1.41} & \underline{83.86\%}  & 45.69\% \\
            &D & \ding{51} & \ding{51} & \textbf{32.89} & \textbf{1.27} & \textbf{84.51\%}  & \textbf{47.48\%} \\
            \hline
        \end{tabular}
        }
\end{table}

\textbf{Patch-based inference.} 
Table~\ref{tab:patch_infer_ablation} shows the quantitative performance of the structure prediction and inpainting models with and without patch-based inference. 
We utilized the patch size with $256^2$, using upsampling scale factors $m=1$ and $m=2$, respectively.
Without patch-based inference for both models, setting (A) shows limited performance. 
Setting (B) (using PSPP) indicates that directly applying patch-based inference to the structure prediction model might worsen performance when the image size is small.
In this case, the inability to capture multiscale text sizes leads to unrelated background noise affecting the inpainting process. 
When we applied it to upscaled images ($m=2$), setting (B) shows some performance gains.
Setting (C) (using CPPD) demonstrates a significant boost in quantitative scores with patch-based inference for the inpainting model. 
Applying this scheme to both models (D) results in competitive performance similar to (C). 
The advantage of (D) is its ability to handle higher-resolution document images, as the model processes each small patch individually. On the higher resolutions, setting (D) shows higher quantitative scores.
Therefore, we chose setting (D), which applies patch-based inference to both models.

\begin{table}[t]
    \centering
    \caption{Ablation study of the structure pyramid (SP) for PSPP. \ding{55} and \ding{51} indicate the unplugging and plugging of this scheme on the structure prediction. \textbf{Bold}: Top-1.  }
    \label{tab:multiscale_ablation}
        \resizebox{0.49\textwidth}{!}{%
        \begin{tabular}{c|c|c|c|c|c|c}
            \hline
            Scale & Config. &  SP  & PSNR$^{\uparrow}$ & FID$_{\text{crop}}^{\downarrow}$&  Char Acc$^{\uparrow}$ & Word Acc$^{\uparrow}$ \\
            \hline
            \multirow{2}{*}{$m=1$}& D & \ding{55} & \textbf{29.06} & \textbf{6.57} & {79.42\%}  & {36.70\%}   \\
            &E &  \ding{51}   & {28.97} & {6.80} &   \textbf{79.74\%} & \textbf{37.33\%} \\
          \hline
            \multirow{2}{*}{$m=2$}&D & \ding{55} & {32.90}  &  {1.27}  &  {84.51\%}   & {47.47\%} \\
            &E &  \ding{51}  & \textbf{32.92} & \textbf{1.24} &   \textbf{84.85\%} & \textbf{47.87\%} \\
            \hline
        \end{tabular}
        }
\end{table}

\begin{table}[t]
    \centering
    \caption{Ablation study of the pyramid patches (PP) for CPPD. \ding{55} and \ding{51} indicate the unplugging and plugging of this scheme on the inpainting model. \textbf{Bold}: Top-1. }
    \label{tab:MLP_ablation}
        \resizebox{0.49\textwidth}{!}{%
        \begin{tabular}{c|c|c|c|c|c|c}
            \hline
            Scale &Config.  & PP  & PSNR$^{\uparrow}$ & FID$_{\text{crop}}^{\downarrow}$&  Char Acc$^{\uparrow}$ & Word Acc$^{\uparrow}$ \\
            \hline
            \multirow{2}{*}{{$m=1$}}& E &  \ding{55}  & {28.97} & {6.80} &   {79.74\%} & {37.33\%}   \\
            &F & \ding{51}  &  \textbf{29.16} &   \textbf{5.49}&  \textbf{80.10\%}  & \textbf{37.61\%}\\
            \hline
            \multirow{2}{*}{{$m=2$}}& E &  \ding{55} & {32.92} & {1.24} &   {84.85\%} & {47.87\%}  \\
            &F & \ding{51} &  \textbf{33.03} &   \textbf{1.21} & \textbf{84.91\%}  & \textbf{48.67\%}\\
            \hline
        \end{tabular}
        }
\end{table}

\textbf{Structure pyramid scheme for PSPP.} 
Table~\ref{tab:multiscale_ablation} illustrates the impact of the structure pyramid scheme on the PSPP of structure pyramid prediction.
We utilized the patch size with $256^2$, using upsampling scale factors $m=1$ and $m=2$, respectively.
We used setting (D) as the baseline.
Without this scheme, the baseline achieves competitive pixel and perceptual level similarities but struggles with varying text sizes due to the fixed patch size.
Setting (E) incorporates this scheme into the structure pyramid prediction, leading to improved metric scores. 
The enhancement further increases with the larger upsampling scale factor $m=2$.
Thus, we apply setting (E), which employs the structure pyramid scheme for the structure prediction model.

\begin{table}[t]
    \centering
    \caption{Effects of the upsampling scale factor $m$ on the Config.~(F). \textbf{Bold}: Top-1. }
    \label{tab:Upsampling_m_ablation}
        \resizebox{0.48\textwidth}{!}{%
        \begin{tabular}{c|c|c|c|c}
            \hline
            Scale   & PSNR$^{\uparrow}$ & FID$_{\text{crop}}^{\downarrow}$&  Char Acc$^{\uparrow}$ & Word Acc$^{\uparrow}$ \\
            \hline
            \multirow{1}{*}{{$m=1$}}   &  {29.16} &   {5.49}&  {80.10\%}  & {37.61\%} \\
            \multirow{1}{*}{{$m=2$}}   & {33.03} &   {1.21} & {84.91\%}  & {48.67\%}  \\
            \multirow{1}{*}{{$m=3$}}   & {34.59} &   {0.87} & {86.13\%}  & {52.26\%}  \\
            \multirow{1}{*}{{$m=4$}}   & \textbf{35.04} &   \textbf{0.78} & \textbf{86.66\%}  & {53.61\%}  \\
            \multirow{1}{*}{{$m=5$}}   & {35.03} &   {0.98} & {86.46\%}  & \textbf{54.93\%}  \\
            \multirow{1}{*}{{$m=6$}}   & {35.03} &   {0.99} & {86.57\%}  & {54.42\%}  \\
            \multirow{1}{*}{{$m=7$}}  & {35.03} &   {0.98} & {86.46\%}  & \textbf{54.93\%}  \\
            \multirow{1}{*}{{$m=8$}}  & {35.03} &   {0.98} & {86.46\%}  & \textbf{54.93\%}  \\
            \hline
        \end{tabular}
        }
\end{table}

\textbf{Pyramid patches for CPPD.}
Table~\ref{tab:MLP_ablation} shows quantitative results obtained by using pyramid patches for CPPD. We used setting (E) as the baseline. 
In setting (F, full model), we applied pyramid patches in the patch pyramid diffusion model. We integrated various patch sizes, including $128^2$ and $256^2$. For upsampling scale factors $m=1$ and $m=2$, setting (F) improved all metric scores. By capturing both local and global information through various patch sizes, the setting (F) achieves significantly better performance compared to using only a single patch of size $256^2$ in (E).

\textbf{Upsampling scale factor $m$.}
Table~\ref{tab:Upsampling_m_ablation} presents quantitative results for various upsampling scale factors $m \in {1, 2, \ldots, 8}$. The results show that increasing the scale factor gradually improves scores, but the gains are saturated at $m=4$.
Higher upsampling factors capture finer details and smaller text, which our TextDoctor leverages through the structure pyramid prediction and patch pyramid diffusion processes. However, since the images must be downsampled back to the original resolution for consistent evaluation, the benefits diminish as resolution increases. Additionally, higher scale factors result in significantly larger input images, increasing computational demands. Thus, we adopt $m=4$ as the default scale.

\section{Conclusion}
In this paper, we present TextDoctor, a novel unified method inspired by human reading behavior to tackle generalization and high-memory consumption challenges in high-resolution document image inpainting.
We propose structure pyramid prediction and patch pyramid diffusion models to capture various text sizes and aggregate multiple inpainting results. 
Extensive experiments on seven document datasets show that TextDoctor effectively handles various text sizes, processes on high-resolution documents, and generalizes well across different styles. 
Our method achieved competitive or superior performance compared with SOTA methods, even without fine-tuning or training on these unseen document datasets.

\textbf{Limitations and future work.}
Although our method excels in document image inpainting for unseen domains, it struggles with large masks, such as those covering entire words or sentences. 
TextDoctor may not accurately reconstruct missing numbers with minimal contextual information. We plan to address this issue in future work. Moreover, the performance of TextDoctor is influenced by the structure prediction and inpainting models, 
where the Mamba-based TextDoctor outperforms its Transformer-based variants. A more sophisticated model can further improve performance. Ultra-high-resolution document image inpainting is supported by utilizing small-scale patches in TextDoctor, but it costs more computational resources. 
Using more mini-batch patches in parallel can speed up the inference.

\bibliographystyle{IEEEtran}
\bibliography{TextDoctor-bib}

\section*{Appendix}

\section{More implementation details of TextDoctor}

\subsection{Training of structure prediction model}
To train the structure prediction model~\cite{Sun_TSINIT,zhu2024gsdm}, the L1 loss $\mathcal{L}_{{pix}}$, binary-cross entropy loss $\mathcal{L}_{{seg}}$, perceptual content loss $\mathcal{L}_{{con}}$, and style loss $\mathcal{L}_{{sty}}$ are applied to constrain the pixel similarity and semantic similarity, as following:
\begin{equation}\label{equ:l1_loss}
    \begin{aligned}
        \mathcal{L}_{{pix}} = \| \boldsymbol{c} - \tilde{\boldsymbol{c}} \|_1,
    \end{aligned}
\end{equation}
\begin{equation}\label{equ:structre_pix_loss}
        \begin{aligned}
            \mathcal{L}_{{seg}} = -\frac{1}{N} \sum_{i=1}^{N} \left( 2 \cdot \tilde{\boldsymbol{c}}_i \log \boldsymbol{c}_i + (1 - \tilde{\boldsymbol{c}}_i) \log(1 - \boldsymbol{c}_i) \right),
        \end{aligned}
\end{equation}
\begin{equation}\label{equ:content_loss}
        \begin{aligned}
            \mathcal{L}_{{con}} = \| \phi(\boldsymbol{c}) - \phi(\tilde{\boldsymbol{c}}) \|_1,
        \end{aligned}
\end{equation}
\begin{equation}\label{equ:style_loss}
        \begin{aligned}
        \mathcal{L}_{{sty}} = \| {Gram}(\boldsymbol{c}) - {Gram}(\tilde{\boldsymbol{c}}) \|_1,
        \end{aligned}
\end{equation}
where, $\tilde{\boldsymbol{c}} \in \mathbb{R}^{1\times h\times w}$ is the ground-truth binary map of a given text patch image;
$N$ is the total number of pixels of a given text image patch; $i$ is the corresponding index.
We use the pre-trained CRNN text recognition model $\phi(\cdot)$~\cite{CRNN} as the feature extraction model.
${Gram}(\cdot)$ represents the Gram matrix~\cite{gatys2015neural}.
The total loss can be expressed as:
\begin{equation}\label{equ:total_loss}
        \begin{aligned}
        \mathcal{L}_{st} = \lambda_{pix} \mathcal{L}_{{pix}} + \lambda_{seg}\mathcal{L}_{{seg}} + \lambda_{con} \mathcal{L}_{{con}} + \lambda_{sty} \mathcal{L}_{{sty}}.
        \end{aligned}
\end{equation}
We empirically set $\lambda_{pix} = 1$, $\lambda_{seg} = 1$, $\lambda_{con} = 1$, and $\lambda_{sty} = 1$.

\subsection{Training of patch pyramid diffusion model}


\textbf{Diffusion process.}
The diffusion process incrementally adds Gaussian noise to the clean image $\boldsymbol{z}_0 \in \mathbb{R}^{3\times h\times w} $ and corrupts it into an approximately pure Gaussian noise $\boldsymbol{z}_T$ using a variance schedule $\beta_1,\ldots, \beta_T$: $q(\boldsymbol{z}_{1:T}|\boldsymbol{z}_0) = \prod_{t=1}^{T} q(\boldsymbol{z}_t|\boldsymbol{z}_{t-1},\boldsymbol{z}_0 )$. Each step is expressed as:
\begin{equation}\label{equ:diffusion_process2}
	\begin{aligned}
        q(\boldsymbol{z}_t|\boldsymbol{z}_{t-1},\boldsymbol{z}_0) &= \mathcal{N}(\boldsymbol{z}_t; \sqrt{1 - \beta_{t}} \boldsymbol{z}_{t-1}, \beta_t \mathbf{I}).
    \end{aligned}
\end{equation}
$\boldsymbol{z}_t$ can be directly approximated by $\boldsymbol{z}_t=\sqrt{\bar{\alpha}_t} \boldsymbol{z}_0+ \sqrt{1-\bar{\alpha}_t} \boldsymbol{\epsilon}$, with $\bar{\alpha}_t=\prod_{s=1}^t \alpha_s$, $\alpha_t=1-\beta_t$, and $\boldsymbol{\epsilon} \sim \mathcal{N}(\mathbf{0}, \mathbf{I})$. 

\textbf{Deterministic reverse process.}
We utilized the deterministic reverse process~\cite{song2020denoising,DocDiff,liu2023textdiff}.
Based on the diffusion process, we can derive it to the reverse diffusion step:
\begin{equation}\label{equ:reverse_diffusion}
	\begin{aligned}
        q(\boldsymbol{z}_{t-1}|\boldsymbol{z}_{t},\boldsymbol{z}_0) = \mathcal{N}(\boldsymbol{z}_{t-1}; \mu_{t}(\boldsymbol{z}_{t},\boldsymbol{z}_{0}), \sigma_{t}(\boldsymbol{z}_{t},\boldsymbol{z}_{0}) \mathbf{I}),
    \end{aligned}
\end{equation}
where $\mu_{t}(\boldsymbol{z}_{t},\boldsymbol{z}_{0})$ and $\sigma_{t}(\boldsymbol{z}_{t},\boldsymbol{z}_{0})$ are the mean and variance, respectively.
We thus can perform the deterministic reverse process $q(\boldsymbol{z}_{t-1}|\boldsymbol{z}_{t},\boldsymbol{z}_0)$ with zero variance ($\sigma_{t}(\boldsymbol{z}_{t},\boldsymbol{z}_{0})=0$) and the mean can be implemented as:
\begin{equation}\label{equ:denoise_process1}
    \begin{aligned}
       \mu_{t}(\boldsymbol{z}_{t},\boldsymbol{z}_{0}) = \sqrt{\bar{\alpha}_{t-1}}{\boldsymbol{z}}_{0}+ \sqrt{1 - \bar{\alpha}_{t-1}} \underbrace{\frac{\boldsymbol{z}_{t} - \sqrt{\bar{\alpha}_t} {\boldsymbol{z}}_{0}}{\sqrt{1-\bar{\alpha}_t}}}_{ \boldsymbol{\epsilon}}.
    \end{aligned}
\end{equation}
We thus can parameterize the posterior $q(\boldsymbol{z}_{t-1}|\boldsymbol{z}_{t},\boldsymbol{z}_0)$ as:
\begin{equation}\label{equ:denoise_process2}
    \begin{aligned}
        p_{\theta}(\boldsymbol{z}_{t-1}|\boldsymbol{z}_t,\boldsymbol{x},\boldsymbol{c}) &= q(\boldsymbol{z}_{t-1}|\boldsymbol{z}_t,g_\theta(\boldsymbol{z}_t, t, \boldsymbol{x},\boldsymbol{c})).
    \end{aligned}
\end{equation}
Given the trained  $g_{\theta}$, the denoising step is implemented as:
\begin{equation}\label{equ:dm_infer2}
	\begin{aligned}
        \boldsymbol{z}_{t-1} &= \sqrt{\bar{\alpha}_{t-1}}\hat{\boldsymbol{z}}_{0}+ \sqrt{1 - \bar{\alpha}_{t-1}} \frac{\boldsymbol{z}_{t} - \sqrt{\bar{\alpha}_t} \hat{\boldsymbol{z}}_{0}}{\sqrt{1-\bar{\alpha}_t}},
    \end{aligned}
\end{equation}
where $\hat{\boldsymbol{z}}_{0} = g_\theta(\boldsymbol{z}_t, t, \boldsymbol{x},\boldsymbol{c})$.

\begin{algorithm}[!ht]
\caption{Training of the structure prediction model}
\label{alg:train_spm}
\begin{algorithmic}[1]
\While {$f_{\bar{\theta}}$ has not converged}
\State Sample batch images $\boldsymbol{z}_{0}$ from training data
\State Sample batch text structure maps $\tilde{\boldsymbol{c}}$ 
\State Create random masks $\boldsymbol{m}$ 
\State Get batch corrupted images $\boldsymbol{x} \leftarrow  \boldsymbol{z}_{0} \odot (\mathbf{1}- \boldsymbol{m}) $
\State $\boldsymbol{c} = f_{\bar{\theta}}(\boldsymbol{x})$  \hfill \textcolor{gray}{\# Structure prediction}
\State Update $f_{\bar{\theta}}$ with $\mathcal{L}_{st}$
\EndWhile
\end{algorithmic}
\end{algorithm}

\begin{algorithm}[!ht]
\caption{Training of the patch pyramid diffusion model}
\label{alg:train_inpainting}
\begin{algorithmic}[1]
\While {$g_{\theta}$ has not converged}
\State Sample batch images $\boldsymbol{z}_{0}$ from training data
\State Create random masks $\boldsymbol{m}$ 
\State Get batch corrupted images $\boldsymbol{x} \leftarrow  \boldsymbol{z}_{0} \odot (\boldsymbol{1}- \boldsymbol{m}) $
\State $t \sim \operatorname{Uniform}(1, ..., T), \boldsymbol{\epsilon} \sim \mathcal{N}(\mathbf{0}, \mathbf{I})$
\State $\boldsymbol{c} = f_{\bar{\theta}}(\boldsymbol{x})$  \hfill \textcolor{gray}{\# Structure prediction}
\State $\boldsymbol{z}_t = \sqrt{\bar{\alpha}_t} \boldsymbol{z}_{0} + \sqrt{1-\bar{\alpha}_t} \boldsymbol{\epsilon}$ \hfill \textcolor{gray}{\#Forward diffusion}
\State $\hat{\boldsymbol{z}}_{0} =g_\theta(\boldsymbol{z}_t, t, \boldsymbol{x},\boldsymbol{c})$ \hfill \textcolor{gray}{\#Patch denoising}
\State Update $g_\theta$ with $\mathcal{L}_{dm}$
\EndWhile
\end{algorithmic}
\end{algorithm}

\begin{algorithm}[!ht]
\caption{Inference of TextDoctor}
\label{alg:multi_inference}
    \begin{algorithmic}[1]
    \State \textcolor{gray}{\#\#\#\#\#\#\#\#\# Structure pyramid prediction phase \#\#\#\#\#\#\#\#\#}
    \State $\mathbf{x}^S \leftarrow \operatorname{IP}_S(\mathbf{X}) $ \hfill \textcolor{gray}{\# Input upsampling}
    \For{$s = 1$ to $S$}
            \State $\mathbf{x}^s \leftarrow \operatorname{IP}_s(\mathbf{x}^S) $ \hfill \textcolor{gray}{\# Image pyramid generation}
            \State $\mathcal{X}^s \leftarrow \mathcal{S}_{split}(\mathbf{x}^s)$ \hfill \textcolor{gray}{\# Partition}
            \State $\mathcal{C}^s \leftarrow  \mathcal{D}(\mathcal{X}^s)$\hfill \textcolor{gray}{\# PSPP}
            \State $\bar{\mathbf{c}}^s \leftarrow  \mathcal{R}(\mathcal{C}^s)$\hfill \textcolor{gray}{\# Merge}
    \EndFor
    \State   $\hat{\mathbf{c}}^S \leftarrow \frac{1}{S} \sum_{s=1}^S \operatorname{IP}_S(\bar{\mathbf{c}}^s)$  \hfill \textcolor{gray}{\# Structure pyramid fusion}
    \State \textcolor{gray}{\#\#\#\#\#\#\#\#\#\# Patch pyramid denoising phase \#\#\#\#\#\#\#\#\#\#\#}
    \State $\mathbf{z}_T^S \sim \mathcal{N}(0, \mathbf{I})$ \hfill \textcolor{gray}{\# Random initialization}
    \For{$t = T$ to $1$}
        \For{$k = 1$ to $K$}
            \State $\mathcal{Z}_t^{(k)} \leftarrow  \mathcal{S}^{(k)}_{split}(\mathbf{z}_t^{S})$  \hfill \textcolor{gray}{\# Partition}
            \State $\mathcal{X}^{(k)} \leftarrow \mathcal{S}^{(k)}_{split}(\mathbf{x}^{S})$ \hfill \textcolor{gray}{\# Partition}
            \State $\hat{\mathcal{C}}^{(k)} \leftarrow \mathcal{S}^{(k)}_{split}(\hat{\mathbf{c}}^{S})$ \hfill \textcolor{gray}{\# Partition}
            \State $\mathcal{Z}^{(k)}_{t-1} \leftarrow \mathcal{P}(\mathcal{Z}^{(k)}_{t-1} | \mathcal{Z}^{(k)}_t,\mathcal{X}^{(k)},\hat{\mathcal{C}}^{(k)})$  \hfill \textcolor{gray}{\# CPPD}
            \State $\bar{\mathbf{z}}^{(k)}_{t-1} \leftarrow \mathcal{R}(\mathcal{Z}^{(k)}_{t-1})$ \hfill \textcolor{gray}{\# Merge}
        \EndFor
        \State   ${\mathbf{z}}^S_{t-1} \leftarrow \frac{1}{K} \sum_{k=1}^{K} \bar{\mathbf{z}}^{(k)}_{t-1}$  \hfill \textcolor{gray}{\# Document image reconstruction}
        \EndFor
    \State $\hat{\mathbf{X}} \leftarrow \operatorname{IP}_{\bar{s}}({\mathbf{z}}^S_{0})$  \hfill \textcolor{gray}{\# Resizing to original dimensions}
    \State \textbf{return} $\hat{\mathbf{X}}$ \hfill \textcolor{gray}{\# Final result}
    \end{algorithmic}
\end{algorithm}

\textbf{Training.}
Our patch-based diffusion inpainting model is trained to predict the original data $\boldsymbol{z}_0$ which aligns with the goals of document image inpainting.
Since we utilize the deterministic reverse process~\cite{song2020denoising} with zero variance, 
the training loss function ensures that the learned conditional distribution $p_{\theta}(\boldsymbol{z}_{t-1}|\boldsymbol{z}_t,\boldsymbol{x},\boldsymbol{c})$ approximates the true reverse diffusion step $q(\boldsymbol{z}_{t-1}|\boldsymbol{z}_{t},\boldsymbol{z}_{0})$ as closely as possible. 
The $\mathcal{L}_{dm}$ can be defined as:
\begin{equation}\label{equ:DM_loss}
	\begin{aligned}
        \mathcal{L}_{dm} = \mathbb{E} \left\| \boldsymbol{z}_0 - g_\theta(\boldsymbol{z}_t, t, \boldsymbol{x},\boldsymbol{c}) \right\|_2.
    \end{aligned}
\end{equation}


\subsection{Implementation} We utilized Python and PyTorch to build the proposed framework. 
We conducted all the experiments on the NVIDIA GeForce RTX 3090 GPU with 24 GB memory.
We provide the pseudo-code for the training procedures in Algorithm~\ref{alg:train_spm}, Algorithm~\ref{alg:train_inpainting} and inference in the Algorithm~\ref{alg:multi_inference}.

We trained the structure prediction model and optimized the patch pyramid diffusion model (inpainting model) independently. 
Initially, the structure prediction model was trained on the text patch image dataset for 50 epochs. 
We used the Adam optimizer with the first momentum coefficient $\beta_1=0.5$, the second momentum coefficient $\beta_2=0.999$, a learning rate of $1\times 10^{-4}$, and the batch size of 32. 
All the input image is resized to the resolution of $64 \times 256$ during training.
It will take around one day to train the structure prediction model.

Next, using the trained structure prediction model, we trained the patch pyramid diffusion model (inpainting model) on the same dataset for $1000,000$ iterations, utilizing the same Adam optimizer with a learning rate of $1\times 10^{-3}$ and the batch size of 8.
During training, the time step was set to $T=2000$. 
The forward process variances to constants increasing linearly from $\beta_1=1\times 10^{-4}$ to  $\beta_T=0.02$. 
All the input image is resized to the resolution of $64 \times 320$.
It will take around one week to train the inpainting model.
During testing, we empirically set $T=1$ following the approach in GSDM~\cite{zhu2024gsdm}. 
We found that this setting achieves impressive performance with high computational efficiency.

\textbf{Details of network architectures.}
We employed the AOT generator~\cite{Zeng2021} for TextDoctor (CNN), GRIG generator~\cite{lu2023grig} for TextDoctor (Trans), and Mamba-based Unet~\cite{ruan2024vm} for TextDoctor (Mamba). 
For each variant, both the structure prediction and patch pyramid diffusion (inpainting) models use the same network architecture, and each inpainting model incorporated additional time step embeddings~\cite{StableDiffusion} before the downsampling layers in the encoder and after the upsampling layers in the decoder. 
For TextDoctor (GSDM), we utilized the structure prediction module and reconstruction module from GSDM~\cite{zhu2024gsdm} as the structure prediction and inpainting models, respectively.
We retrained these models from scratch on the text image patch dataset.

\begin{table}[!htbp]
	\centering
    \caption{The number of images used for training and testing, and height and width ranges, for the seven document image datasets.  }
	\resizebox{1\linewidth}{!}{$
		\begin{tabular}{c|cc|c|c}
			\hline
			 Dataset & \# Training & \# Test & Height range & Width range\\
            \hline
            FUNSD   &  149 & 50 & $[1000, 1000]$ & $[754, 863]$ \\ 
            \hline
             BCSD     & 129    & 29& $[750, 1168]$ & $[2240, 2240]$ \\
            \hline
             Twitter posts    &  87   &    9 & $[1520, 1520]$ & $[720, 720]$   \\
            \hline
             Twitter profiles    & 425    & 61 & $[122, 11688]$ & $[236, 2048]$\\
            \hline
		Activity diagrams   &  259   &    45  & $[85, 2811]$ & $[125, 2740]$  \\ 
            \hline
             Signatures    & 257    & 37 & $[131, 4200]$ & $[184, 3506]$\\
            \hline
		Tabular documents   &  3,251   &    206  & $[70, 2697]$ & $[176, 3012]$  \\ 
            \hline
	\end{tabular}
 $}
	\label{tab:datasets}
\end{table}

\section{More experimental results}

\subsection{Details of benchmark datasets}
\textbf{Datasets.} 
We combined the training sets from TII-ST and TII-HT datasets~\cite{zhu2024gsdm} as our text image patch dataset. It contains $118,478$ images. Each image is with a resolution of $64 \times 256$.
Each image contains several characters or words and is paired with a corresponding binary segmentation map that outlines the text's foreground, as well as a corrupted mask.
Table~\ref{tab:datasets} shows the details of FUNSD~\cite{jaume2019}, BCSD~\cite{khan2021novel}, and the five document subsets from the Roboflow 100 dataset~\cite{ciaglia2022roboflow} (Twitter posts, Twitter profiles, activity diagrams, signatures, and tabular documents).

\textbf{Masks for training and testing.}
For training models (TextDoctor-based variants, and other compared models) on text patch images, we applied a combined binary mask to each image.
The combined mask was created using a pixel-level $\operatorname{OR}(\cdot)$ operation to merge a corrupted mask from the dataset with a free-form mask.
The free-form mask was generated by simulating random brush strokes and rectangles. We used the mask generation algorithm from GConv~\cite{Yu2019}, with parts of 20, a maximum vertex count of 20, a maximum length of 20, a maximum brush width of 24, and a maximum angle of 360 degrees. 
Additionally, the number of rectangles sized $100\times 100$ and $50\times 50$ were uniformly sampled within $[0, 5]$ and $[0, 10]$, respectively.
For each image, we generate the masked images, as expressed in the following equation:
\begin{equation}\label{equ:masked_gen}
	\begin{aligned}
        \boldsymbol{x} = \boldsymbol{z}_0 \odot (\boldsymbol{1} - \boldsymbol{m}),
    \end{aligned}
\end{equation}
where $\boldsymbol{x}$ is the corrupted image; $\boldsymbol{z}_0$ is the ground-truth image; $\boldsymbol{m}$ is the generated binary mask; $\odot$ denotes the Hadamard product, respectively. 


For training and testing the document images, we directly applied the mask generation algorithm from GConv~\cite{Yu2019} to create masks sized $4096 \times 4096$, with the following parameters: parts of 200, a maximum vertex count of 40, a maximum length of 60, a maximum brush width of 24, and a maximum angle of 360 degrees. 
Additionally, the number of rectangles sized $128\times 128$ and $64\times 64$ were uniformly sampled within the ranges $[0, 50]$ and $[0, 120]$, respectively.
For each image, we resized these masks to align the image size and created the masked images using Eq.~\ref{equ:masked_gen}.

For testing on document images, we first created a masked image for each ground-truth image in the test sets of the seven document image datasets. These masked images were then used to evaluate the compared methods. We provided binary masks for the non-blind methods (AOT and GRIG), but did not supply masks for the other blind inpainting methods.


\subsection{Details of evaluation metrics}
We used Peak Signal-to-Noise Ratio (PSNR) (dB) and Structural Similarity (SSIM) to evaluate pixel-level similarity. 
For the patch-based FID (FID$_{\text{crop}}$) and patch-based U-IDS (U-IDS$_{\text{crop}}$), images were cropped into overlapping $256^2$ patches with strides of 128 pixels both vertically and horizontally, and the FID~\cite{Heusel2017} and U-IDS~\cite{Zhao2021} were calculated, respectively, based on these patches~\cite{Du_2024_CVPR}.

For character-level (Char Acc) and word-level (Word Acc) recognition accuracy~\cite{zhu2024gsdm}, we used PaddleOCR~\cite{PaddleOCR} to detect text bounding boxes in both the ground-truth and inpainted images, followed by text recognition using CRNN~\cite{CRNN}, MORAN~\cite{MORAN}, and PaddleOCR's recognition function~\cite{PaddleOCR}. 
The FUNSD dataset provides ground-truth bounding boxes and text labels, which we used directly. 
For the other six datasets, we detected text bounding boxes from the test sets using PaddleOCR and treated them as ground-truth text labels.

\begin{figure*}[!ht]
	\centering
	\includegraphics[width=1.0\textwidth]{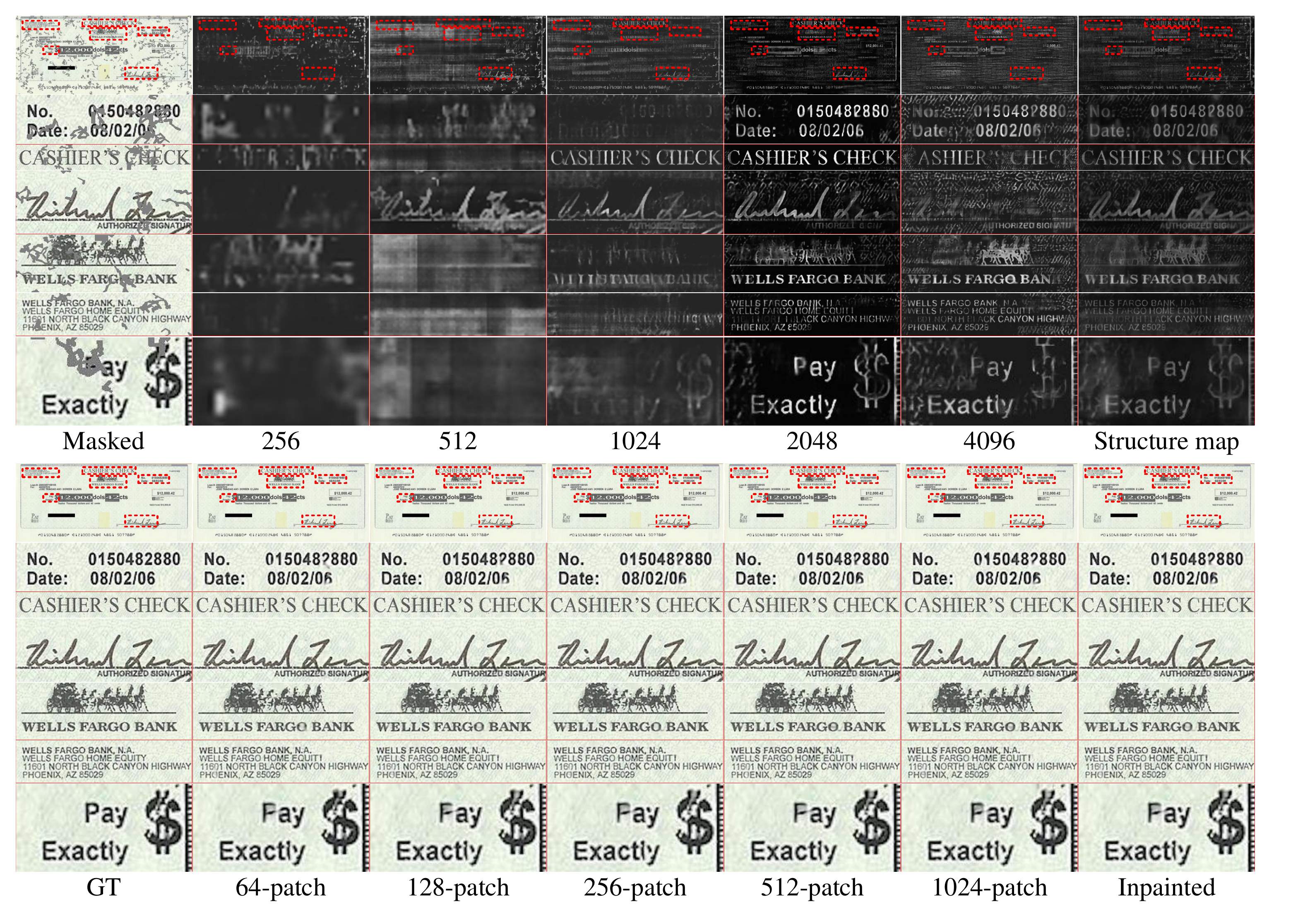}
	\caption{The intermediate visual results from the structure pyramid prediction (top) and patch pyramid diffusion model (bottom). The first column displays the masked images (top) alongside the ground-truth images (bottom). The top row shows the predicted structure maps from various multiscale inputs, with the numbers indicating the minimum side lengths (256, 512, 1024, 2048, 4096). The bottom row presents the inpainted results using patches of different sizes, where $y$-patch refers to a patch size of $y^2$.  The ``Structure map'' refers to the result of structure pyramid fusion, while the ``Inpainted'' refers to the final document image reconstruction using all patches. Zoom in for a closer review.}\label{fig:fig_vis_pyramids_res}  
\end{figure*}

\begin{figure*}[!ht]
	\centering
	\includegraphics[width=1.0\textwidth]{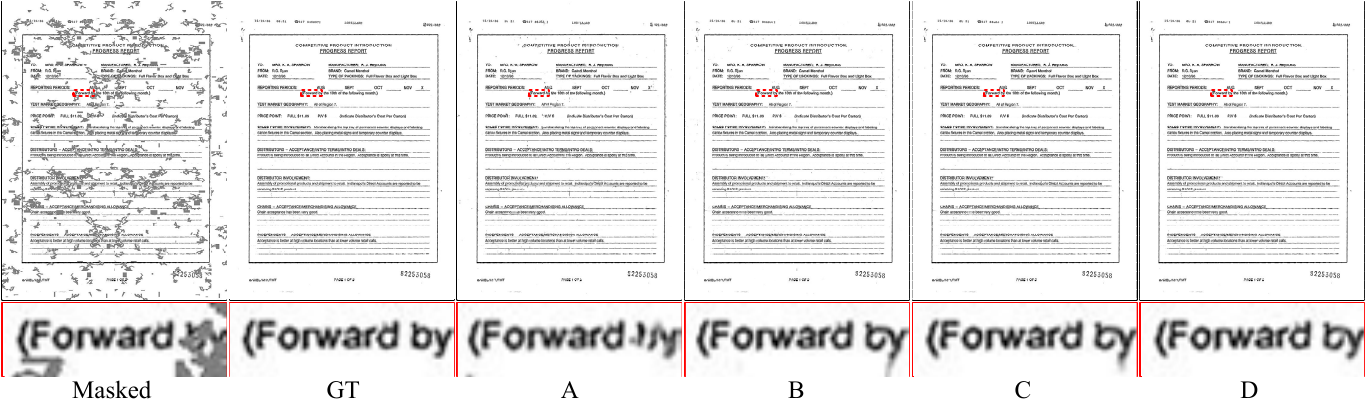}
	\caption{Visual comparisons of the ablation study for the patch-based inference.}\label{fig:fig_ablation_abcd}  
\end{figure*}

\begin{figure*}[!ht]
	\centering
        \includegraphics[width=0.66\textwidth]{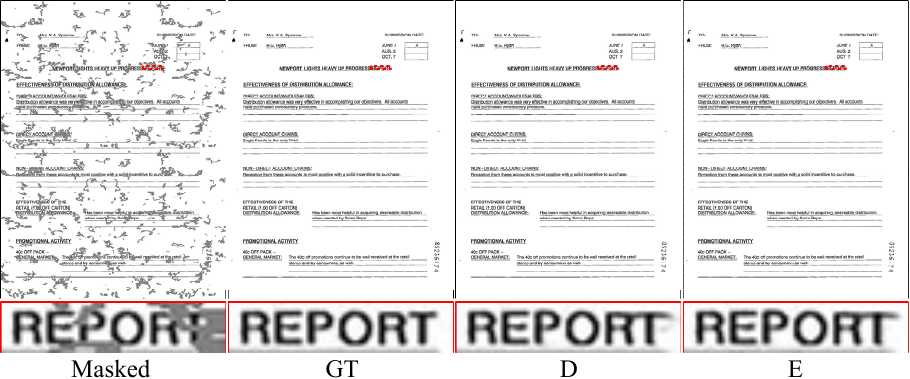}
	\caption{Visual comparisons of the ablation study on the structure pyramid scheme for PSPP.}\label{fig:fig_ablation_defg} 
\end{figure*}

\subsection{More results of ablation study}

\textbf{More visual results.} Fig.~\ref{fig:fig_vis_pyramids_res} presents additional intermediate visual results from the structure pyramid prediction and patch pyramid diffusion model (using  TextDoctor (Mamba)). 
We utilized pyramid inputs with various minimum side lengths (256, 512, 1024, 2048, 4096) and patch sizes ($64^2$, $128^2$, $256^2$, $512^2$, $1024^2$).
As shown in Fig.~\ref{fig:fig_vis_pyramids_res} (top), the structure prediction model effectively captures both local and global information, with smaller inputs providing a global layout (e.g., document titles) and higher resolutions capturing finer details like background textures.
As shown in Fig.~\ref{fig:fig_vis_pyramids_res} (bottom), the different patch sizes help capture multiscale information, ensuring generalization across diverse document styles and effectively handling text of varying sizes.

Fig.~\ref{fig:fig_ablation_abcd} presents additional visual results from the ablation study on patch-based inference discussed in the main paper. The study highlights various configurations for the structure prediction (ST) and inpainting (IN) models: (A) Baseline, (B) Baseline + patch-based inference (with ST), (C) Baseline + patch-based inference (with IN), and (D) Baseline + patch-based inference (with both ST and IN).

Fig.~\ref{fig:fig_ablation_defg} displays results from the ablation study on structure pyramid scheme in the main paper with different configurations: (D) Baseline (same as the above setting (D)), (E) Baseline + structure pyramid scheme for patch-based structure pyramid prediction (PSPP).

\begin{figure*}[t]
	\centering
          \includegraphics[width=0.66\textwidth]{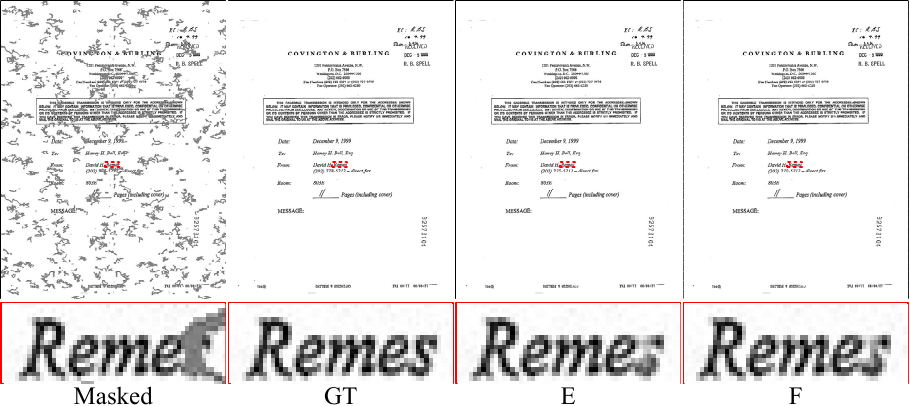}
	\caption{Visual comparisons from the ablation study on pyramid patches for CPPD. }\label{fig:fig_ablation_eh}  
\end{figure*}

\begin{figure}[!ht]
	\centering
        \includegraphics[width=0.49\textwidth]{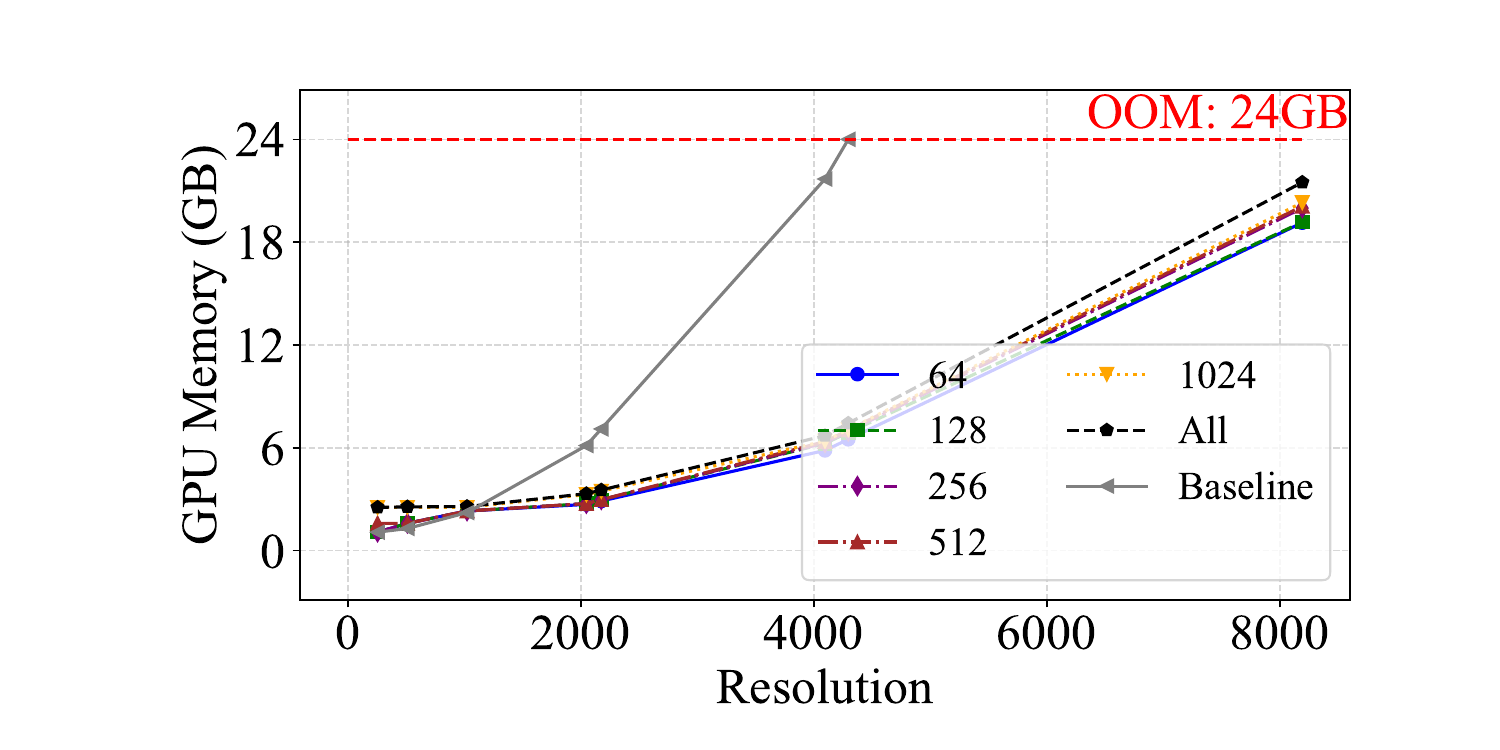}
	\caption{Memory demands of TextDoctor (Mamba) using various patches and the baseline method (without TextdDoctor).}\label{fig:plot_GPU_memory_textdoctor} 
\end{figure}

\begin{figure}[!ht]
	\centering
    \includegraphics[width=0.49\textwidth]{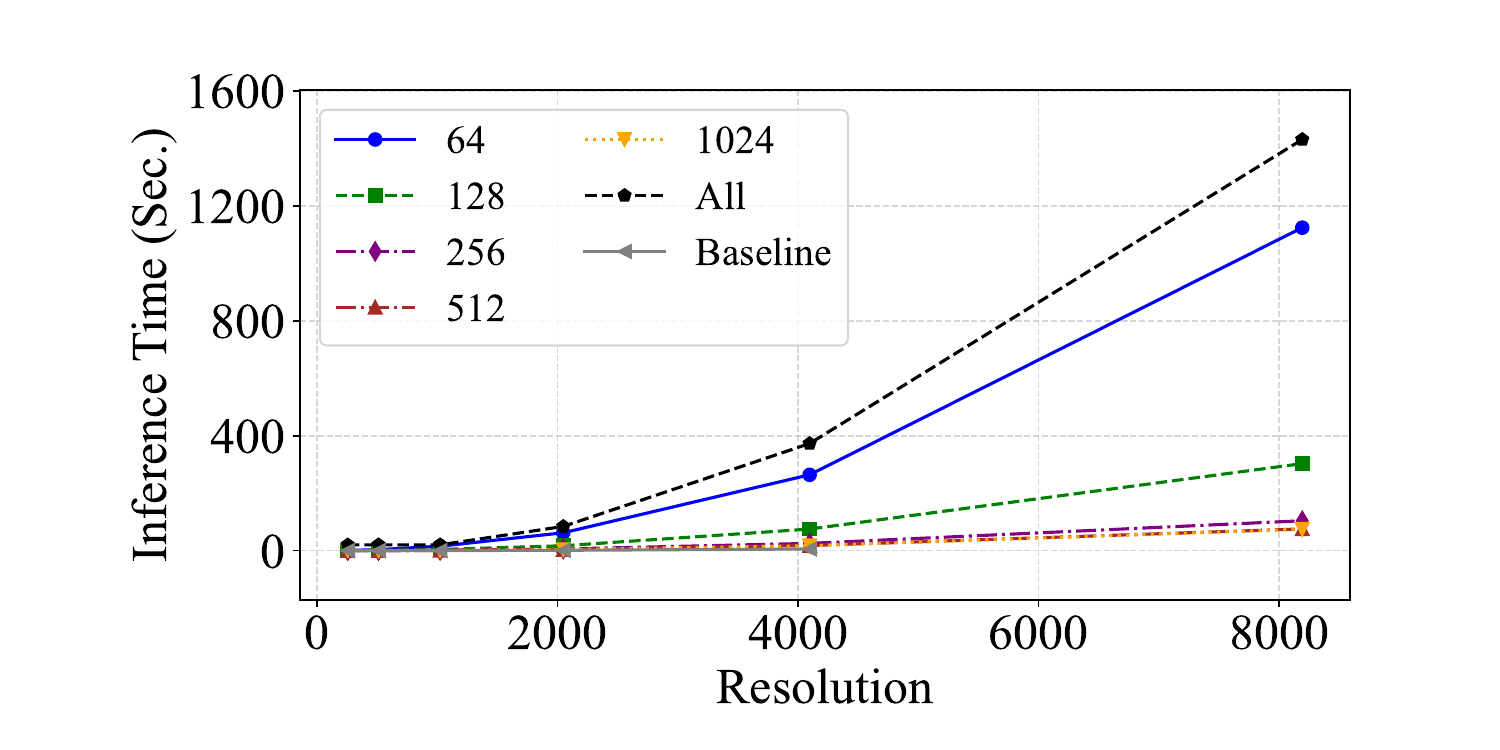}
	\caption{Inference time of TextDoctor (Mamba) using various patches and the baseline method (without TextdDoctor).}\label{fig:plot_speed_textdoctor} 
\end{figure}


Fig.~\ref{fig:fig_ablation_eh} illustrates results from the ablation study on pyramid patches in the main paper with different configurations: (E) Baseline (same as the above setting (E)), (F) Baseline + pyramid patches for conditional patch pyramid denoising (CPPD).

\textbf{Ablation study on patch sizes.}
We analyzed the resource demands of TextDoctor (Mamba). As shown in Fig.~\ref{fig:plot_GPU_memory_textdoctor}, the GPU memory requirements for different patch sizes are similar, as patch-based inference processes one patch at a time. 
When combining all patch sizes (All), our method requires slightly more GPU memory at higher resolutions. 
TextDoctor can handle resolutions above 8K, whereas the baseline Mamba model is limited to 4K resolution, demonstrating our method's capability to process high-resolution document images for high-quality inpainting.

Fig.~\ref{fig:plot_speed_textdoctor} illustrates the inference time for different patch sizes. We used a batch size of 1 and a single patch for each size. Smaller patch sizes result in higher computational time costs. Utilizing all patch sizes (All) accumulates computational time across the five different patch sizes. Note that the use of only one patch per size shifts the bottleneck to CPU processing speed rather than GPU performance. Increasing the number of mini-batch patches processed in parallel for each size can significantly enhance inference speed.

\begin{table*}[!htbp]
    \centering
    \caption{Ablation study for the inference phase of patch pyramid diffusion model using various patch sizes on the FUNSD dataset (upsampling scale factor of $m=1$). \ding{55} and \ding{51} indicate the unplugging and plugging of the patch on the inpainting model. \textbf{Bold} indicates the best result, and \underline{Underline} represents the best result within each group of the same patch number.}
    \label{tab:abla_patch_size_FUNSD_1}
    \resizebox{\textwidth}{!}{%
        \begin{tabular}{c|ccccc|cc|cc|cc|cc|cc}
            \hline
             &\multicolumn{5}{c|}{Patch size} &  \multicolumn{2}{c|}{Pixel-level}   & \multicolumn{2}{c|}{Perceptual-level} 
 & \multicolumn{2}{c|}{CRNN} & \multicolumn{2}{c|}{MORAN} & \multicolumn{2}{c}{PaddleOCR} \\
            \hline
             &64&128 &256 &512 &1024  & PSNR$^{\uparrow}$ & SSIM$^{\uparrow}$ & FID$_{\text{crop}}^{\downarrow}$ & U-IDS$_{\text{crop}}^{\uparrow}$ &  Char Acc$^{\uparrow}$ & Word Acc$^{\uparrow}$ &  Char Acc$^{\uparrow}$ & Word Acc$^{\uparrow}$ &  Char Acc$^{\uparrow}$ & Word Acc$^{\uparrow}$ \\
            \hline
            \multirow{5}{*}{\rotatebox{90}{one patch}}&\ding{51}&\ding{55}&\ding{55}&\ding{55}&\ding{55}&28.02&0.9814&5.95&1.11\%&51.81\%&14.37\%&\underline{45.81\%}&12.82\%&79.62\%&37.49\%\\
            &\ding{55}&\ding{51}&\ding{55}&\ding{55}&\ding{55}&28.65&0.9843&\underline{\textbf{4.97}}&\underline{\textbf{2.23\%}}&\underline{51.86\%}&\underline{14.88\%}&45.51\%&\underline{13.55\%}&79.78\%&\underline{37.80\%}\\
            &\ding{55}&\ding{55}&\ding{51}&\ding{55}&\ding{55}&\underline{28.96}&\underline{0.9844}&6.69&1.83\%&51.71\%&14.49\%&45.69\%&13.21\%&\underline{79.88\%}&37.20\%\\
            &\ding{55}&\ding{55}&\ding{55}&\ding{51}&\ding{55}&28.60&0.9789&16.71&0.06\%&51.64\%&14.19\%&45.68\%&12.95\%&78.77\%&35.61\%\\
            &\ding{55}&\ding{55}&\ding{55}&\ding{55}&\ding{51}&27.67&0.9726&25.68&0.00\%&51.36\%&14.34\%&45.14\%&12.78\%&77.79\%&34.45\%\\
            \hline
            \multirow{10}{*}{\rotatebox{90}{two patches}}&\ding{51}&\ding{51}&\ding{55}&\ding{55}&\ding{55}&28.64&0.9842&\underline{\textbf{4.97}}&1.37\%&52.05\%&14.88\%&\underline{46.28\%}&\underline{13.68\%}&80.05\%&\underline{38.53\%}\\
            &\ding{51}&\ding{55}&\ding{51}&\ding{55}&\ding{55}&28.85&0.9844&6.14&1.00\%&\underline{52.38\%}&\underline{15.09\%}&46.04\%&13.46\%&\underline{80.23\%}&37.98\%\\
            &\ding{51}&\ding{55}&\ding{55}&\ding{51}&\ding{55}&28.88&0.9819&12.98&0.09\%&52.27\%&14.97\%&46.06\%&13.59\%&80.03\%&37.82\%\\
            &\ding{51}&\ding{55}&\ding{55}&\ding{55}&\ding{51}&28.58&0.9784&18.07&0.03\%&52.03\%&14.71\%&45.91\%&13.34\%&79.63\%&37.61\%\\
            &\ding{55}&\ding{51}&\ding{51}&\ding{55}&\ding{55}&29.16&\underline{\textbf{0.9857}}&5.49&\underline{1.66\%}&52.21\%&14.67\%&45.94\%&13.29\%&80.10\%&37.61\%\\
            &\ding{55}&\ding{51}&\ding{55}&\ding{51}&\ding{55}&29.10&0.9826&12.06&0.20\%&52.34\%&15.05\%&45.93\%&13.59\%&79.86\%&38.15\%\\
            &\ding{55}&\ding{51}&\ding{55}&\ding{55}&\ding{51}&28.82&0.9793&17.57&0.03\%&52.00\%&14.63\%&45.68\%&12.91\%&79.36\%&36.99\%\\
            &\ding{55}&\ding{55}&\ding{51}&\ding{51}&\ding{55}&\underline{29.24}&0.9828&13.22&0.17\%&52.21\%&14.67\%&45.83\%&13.34\%&79.96\%&36.81\%\\
            &\ding{55}&\ding{55}&\ding{51}&\ding{55}&\ding{51}&28.92&0.9795&18.31&0.00\%&51.97\%&14.75\%&45.84\%&13.25\%&79.72\%&36.98\%\\
            &\ding{55}&\ding{55}&\ding{55}&\ding{51}&\ding{51}&28.82&0.9770&23.32&0.00\%&52.07\%&14.79\%&46.00\%&13.25\%&79.24\%&36.22\%\\
            \hline
            \multirow{10}{*}{\rotatebox{90}{three patches}}&\ding{51}&\ding{51}&\ding{51}&\ding{55}&\ding{55}&29.01&\underline{0.9854}&\underline{5.37}&\underline{1.20\%}&52.28\%&15.01\%&46.23\%&13.42\%&80.15\%&38.29\%\\
            &\ding{51}&\ding{51}&\ding{55}&\ding{51}&\ding{55}&29.02&0.9838&10.63&0.26\%&52.10\%&14.67\%&45.99\%&13.21\%&80.06\%&38.37\%\\
            &\ding{51}&\ding{51}&\ding{55}&\ding{55}&\ding{51}&28.90&0.9812&14.82&0.06\%&52.16\%&14.59\%&46.31\%&\underline{13.64\%}&79.95\%&38.11\%\\
            &\ding{51}&\ding{55}&\ding{51}&\ding{51}&\ding{55}&29.15&0.9838&10.95&0.20\%&52.26\%&15.01\%&\underline{\textbf{46.46\%}}&13.46\%&79.85\%&37.92\%\\
            &\ding{51}&\ding{55}&\ding{51}&\ding{55}&\ding{51}&29.00&0.9812&15.73&0.06\%&\underline{\textbf{52.42\%}}&\underline{\textbf{15.23\%}}&45.94\%&13.21\%&\underline{80.27\%}&38.35\%\\
            &\ding{51}&\ding{55}&\ding{55}&\ding{51}&\ding{51}&29.02&0.9798&19.66&0.00\%&52.16\%&14.84\%&46.10\%&13.59\%&80.04\%&37.86\%\\
            &\ding{55}&\ding{51}&\ding{51}&\ding{51}&\ding{55}&\underline{29.28}&0.9843&10.29&0.17\%&52.19\%&14.88\%&46.23\%&13.46\%&79.95\%&37.53\%\\
            &\ding{55}&\ding{51}&\ding{51}&\ding{55}&\ding{51}&29.18&0.9820&14.95&0.03\%&52.32\%&15.22\%&46.17\%&13.46\%&80.19\%&\underline{38.41\%}\\
            &\ding{55}&\ding{51}&\ding{55}&\ding{51}&\ding{51}&29.15&0.9801&19.08&0.06\%&52.29\%&15.01\%&46.02\%&13.51\%&79.88\%&37.76\%\\
            &\ding{55}&\ding{55}&\ding{51}&\ding{51}&\ding{51}&29.23&0.9801&20.03&0.03\%&52.29\%&14.92\%&45.88\%&13.42\%&79.93\%&37.42\%\\
            \hline
            \multirow{5}{*}{\rotatebox{90}{four patches}}&\ding{51}&\ding{51}&\ding{51}&\ding{51}&\ding{55}&29.18&\underline{0.9846}&\underline{9.43}&\underline{0.11\%}&52.18\%&14.84\%&\underline{46.39\%}&\underline{\textbf{13.81}}\%&79.96\%&38.15\%\\
            &\ding{51}&\ding{51}&\ding{51}&\ding{55}&\ding{51}&29.13&0.9829&13.32&0.06\%&\underline{52.29\%}&14.84\%&46.13\%&13.42\%&\underline{80.24\%}&38.08\%\\
            &\ding{51}&\ding{51}&\ding{55}&\ding{51}&\ding{51}&29.13&0.9818&16.87&0.06\%&52.27\%&14.88\%&45.84\%&13.42\%&80.23\%&\underline{38.62\%}\\
           &\ding{51}&\ding{55}&\ding{51}&\ding{51}&\ding{51}&29.21&0.9817&17.30&0.06\%&52.20\%&\underline{14.93\%}&46.21\%&13.55\%&80.20\%&37.86\%\\
            &\ding{55}&\ding{51}&\ding{51}&\ding{51}&\ding{51}&\underline{\textbf{29.29}}&0.9821&16.38&0.06\%&52.27\%&14.84\%&46.31\%&13.68\%&79.95\%&37.71\%\\
            \hline
            &\ding{51}&\ding{51}&\ding{51}&\ding{51}&\ding{51}&29.25&0.9829&15.23&0.03\%&52.41\%&15.05\%&46.13\%&13.51\%&\textbf{80.41\%}&\textbf{38.83\%}\\
            \hline
        \end{tabular}
    }
\end{table*}

\begin{table*}[!htbp]
    \centering
    \caption{Ablation study for the inference phase of patch pyramid diffusion model using various patch sizes on the FUNSD dataset (upsampling scale factor of $m=2$). \ding{55} and \ding{51} indicate the unplugging and plugging of the patch on the inpainting model. \textbf{Bold} indicates the best result, and \underline{Underline} represents the best result within each group of the same patch number.}
    \label{tab:abla_patch_size_FUNSD_2}
    \resizebox{\textwidth}{!}{%
        \begin{tabular}{c|ccccc|cc|cc|cc|cc|cc}
            \hline
             &\multicolumn{5}{c|}{Patch size} &  \multicolumn{2}{c|}{Pixel-level}   & \multicolumn{2}{c|}{Perceptual-level} 
 & \multicolumn{2}{c|}{CRNN} & \multicolumn{2}{c|}{MORAN} & \multicolumn{2}{c}{PaddleOCR} \\
            \hline
             &64&128 &256 &512 &1024  & PSNR$^{\uparrow}$ & SSIM$^{\uparrow}$ & FID$_{\text{crop}}^{\downarrow}$ & U-IDS$_{\text{crop}}^{\uparrow}$ &  Char Acc$^{\uparrow}$ & Word Acc$^{\uparrow}$ &  Char Acc$^{\uparrow}$ & Word Acc$^{\uparrow}$ &  Char Acc$^{\uparrow}$ & Word Acc$^{\uparrow}$ \\
            \hline
            \multirow{5}{*}{\rotatebox{90}{one kernel}}&\ding{51}&\ding{55}&\ding{55}&\ding{55}&\ding{55}&32.65&0.9947&1.32&13.80\%&\underline{54.87\%}&16.47\%&48.55\%&15.01\%&84.67\%&\underline{48.19\%}\\                        &\ding{55}&\ding{51}&\ding{55}&\ding{55}&\ding{55}&32.87&\underline{0.9949}&\underline{1.22}&15.03\%&54.74\%&\underline{16.81\%}&48.69\%&15.18\%&\underline{84.94\%}&48.17\%\\
            &\ding{55}&\ding{55}&\ding{51}&\ding{55}&\ding{55}&\underline{32.92}&\underline{0.9949}&\underline{1.22}&\underline{\textbf{15.17\%}}&54.82\%&16.34\%&\underline{48.85\%}&15.01\%&84.73\%&47.93\%\\
            &\ding{55}&\ding{55}&\ding{55}&\ding{51}&\ding{55}&32.84&0.9936&3.36&6.43\%&54.64\%&16.30\%&48.80\%&\underline{\textbf{15.22\%}}&84.62\%&47.89\%\\
            &\ding{55}&\ding{55}&\ding{55}&\ding{55}&\ding{51}&32.45&0.9904&7.69&3.37\%&54.70\%&16.68\%&48.55\%&14.84\%&84.62\%&47.08\%\\
            \hline
            \multirow{10}{*}{\rotatebox{90}{two patches}}&\ding{51}&\ding{51}&\ding{55}&\ding{55}&\ding{55}&32.93&0.9949&1.27&14.11\%&54.83\%&16.25\%&48.69\%&\underline{\textbf{15.22\%}}&84.88\%&48.72\%\\
            &\ding{51}&\ding{55}&\ding{51}&\ding{55}&\ding{55}&32.97&0.9950&1.27&14.06\%&55.00\%&16.77\%&48.49\%&14.75\%&\underline{85.17\%}&\underline{48.76\%}\\
            &\ding{51}&\ding{55}&\ding{55}&\ding{51}&\ding{55}&32.95&0.9944&2.45&8.60\%&54.87\%&16.47\%&48.74\%&\underline{\textbf{15.22\%}}&84.84\%&48.32\%\\
            &\ding{51}&\ding{55}&\ding{55}&\ding{55}&\ding{51}&32.85&0.9926&5.47&3.43\%&55.02\%&16.90\%&48.74\%&14.97\%&84.82\%&48.04\%\\
            &\ding{55}&\ding{51}&\ding{51}&\ding{55}&\ding{55}&\underline{33.03}&\underline{\textbf{0.9951}}&\underline{\textbf{1.21}}&\underline{14.91\%}&55.06\%&16.51\%&48.52\%&14.67\%&84.91\%&48.67\%\\
            &\ding{55}&\ding{51}&\ding{55}&\ding{51}&\ding{55}&33.00&0.9944&2.54&7.77\%&55.00\%&16.77\%&48.69\%&15.01\%&84.67\%&48.04\%\\
            &\ding{55}&\ding{51}&\ding{55}&\ding{55}&\ding{51}&32.91&0.9928&5.41&3.57\%&\underline{\textbf{55.15\%}}&\underline{16.98\%}&\underline{\textbf{48.94\%}}&15.01\%&84.73\%&48.41\%\\
            &\ding{55}&\ding{55}&\ding{51}&\ding{51}&\ding{55}&33.02&0.9945&2.54&8.37\%&54.86\%&16.60\%&48.65\%&14.88\%&84.95\%&48.15\%\\
            &\ding{55}&\ding{55}&\ding{51}&\ding{55}&\ding{51}&32.93&0.9932&4.51&4.94\%&54.95\%&16.90\%&48.64\%&14.97\%&84.89\%&48.28\%\\
            &\ding{55}&\ding{55}&\ding{55}&\ding{51}&\ding{51}&32.90&0.9925&6.50&1.60\%&54.79\%&16.64\%&48.66\%&15.05\%&84.69\%&47.63\%\\
            \hline
            \multirow{10}{*}{\rotatebox{90}{three patches}}&\ding{51}&\ding{51}&\ding{51}&\ding{55}&\ding{55}&33.03&\underline{0.9950}&\underline{1.23}&\underline{14.29\%}&55.04\%&16.64\%&48.73\%&\underline{\textbf{15.22\%}}&84.94\%&48.72\%\\
            &\ding{51}&\ding{51}&\ding{55}&\ding{51}&\ding{55}&33.02&0.9947&2.00&9.74\%&55.01\%&16.72\%&48.75\%&15.14\%&84.93\%&48.24\%\\
            &\ding{51}&\ding{51}&\ding{55}&\ding{55}&\ding{51}&33.00&0.9939&3.76&6.20\%&55.04\%&16.64\%&48.81\%&15.14\%&84.99\%&48.24\%\\
            &\ding{51}&\ding{55}&\ding{51}&\ding{51}&\ding{55}&33.04&0.9947&2.10&9.49\%&54.98\%&16.55\%&48.78\%&15.09\%&\underline{85.19\%}&\underline{\textbf{49.06\%}}\\
            &\ding{51}&\ding{55}&\ding{51}&\ding{55}&\ding{51}&33.01&0.9939&3.93&5.14\%&54.88\%&16.60\%&\underline{48.83\%}&15.09\%&84.91\%&48.09\%\\
            &\ding{51}&\ding{55}&\ding{55}&\ding{51}&\ding{51}&33.00&0.9934&5.45&2.03\%&54.99\%&\underline{16.90\%}&48.79\%&14.88\%&84.87\%&48.00\%\\
            &\ding{55}&\ding{51}&\ding{51}&\ding{51}&\ding{55}&\underline{33.06}&0.9948&2.06&9.17\%&54.95\%&16.51\%&48.59\%&15.01\%&84.84\%&48.65\%\\
            &\ding{55}&\ding{51}&\ding{51}&\ding{55}&\ding{51}&33.03&0.9938&4.04&5.23\%&\underline{55.09\%}&16.72\%&48.77\%&15.05\%&84.99\%&48.37\%\\
            &\ding{55}&\ding{51}&\ding{55}&\ding{51}&\ding{51}&33.01&0.9935&4.93&2.31\%&54.88\%&16.60\%&48.62\%&14.84\%&84.80\%&48.50\%\\
            &\ding{55}&\ding{55}&\ding{51}&\ding{51}&\ding{51}&33.02&0.9936&4.81&2.66\%&54.89\%&16.30\%&48.79\%&15.18\%&84.83\%&47.87\%\\
            \hline
            \multirow{5}{*}{\rotatebox{90}{four patches}}&\ding{51}&\ding{51}&\ding{51}&\ding{51}&\ding{55}&\underline{\textbf{33.07}}&\underline{0.9949}&\underline{1.82}&\underline{10.46\%}&54.83\%&16.47\%&48.72\%&\underline{15.14\%}&84.87\%&48.72\%\\
            &\ding{51}&\ding{51}&\ding{51}&\ding{55}&\ding{51}&33.05&0.9941&3.90&5.57\%&54.99\%&16.81\%&48.69\%&14.88\%&84.99\%&48.69\%\\
            &\ding{51}&\ding{51}&\ding{55}&\ding{51}&\ding{51}&33.06&0.9939&4.08&3.09\%&54.88\%&16.34\%&48.77\%&\underline{15.14\%}&\underline{85.18\%}&48.80\%\\
            &\ding{51}&\ding{55}&\ding{51}&\ding{51}&\ding{51}&33.05&0.9941&3.96&3.83\%&\underline{55.13\%}&16.77\%&\underline{48.86\%}&\underline{15.14\%}&84.87\%&\underline{48.93\%}\\
            &\ding{55}&\ding{51}&\ding{51}&\ding{51}&\ding{51}&\underline{\textbf{33.07}}&0.9939&4.29&3.46\%&55.00\%&\underline{\textbf{17.11\%}}&48.82\%&14.97\%&85.03\%&48.54\%\\
            
            \hline
            &\ding{51}&\ding{51}&\ding{51}&\ding{51}&\ding{51}&\textbf{33.07}&0.9942&4.08&3.49\%&55.12\%&16.94\%&48.61\%&14.84\%&\textbf{85.27\%}&48.78\%\\   
            \hline
        \end{tabular}
    }
\end{table*}

We further tested all 32 combinations (subsets) of five patch sizes for the inference phase of the patch pyramid diffusion model, including $64^2$, $128^2$, $256^2$, $512^2$, and $1024^2$. 
As shown in Table~\ref{tab:abla_patch_size_FUNSD_1} and Table~\ref{tab:abla_patch_size_FUNSD_2}, we observed that using more patch sizes leads to improved metric scores. 
For instance, utilizing all five patch sizes yields the highest Char ACC and Word ACC of $80.41\%$ and $38.83\%$ (in Table~\ref{tab:abla_patch_size_FUNSD_1}), respectively. 
The results also indicate that configurations using a higher upsampling factor ($m=2$) provide better inpainting quality than those with a lower upsampling factor ($m=1$). 
It highlights the importance of upsampling for high-quality image inpainting despite increased computational costs. 

The configuration using $128^2$ and $256^2$ patch sizes demonstrates robust performance. 
Considering both inpainting quality and computational efficiency based on the above analyses, we selected the configuration with $128^2$ and $256^2$ patch sizes for our patch pyramid diffusion models.







\subsection{Implementation of compared methods} 
We used the authors' codes for AOT~\cite{Zeng2021}, GRIG~\cite{lu2023grig}, DocDiff~\cite{DocDiff}, DocRes~\cite{DocRes}, and GSDM~\cite{zhu2024gsdm}. 
For DE-GAN~\cite{DEGAN}, we implemented a Pytorch version based on the authors' code.
For testing on document images, we resized unsupported images while maintaining their aspect ratios to accommodate the resolution limits of each method.

\begin{table*}[!ht]
    \centering
    \caption{Comparison to state-of-the-art methods on the other six datasets (BCSD).  \textbf{Bold} indicates our methods outperform SOTA methods or the best-performing method.}
    \label{tab:ori_comparison_all_1}
    \resizebox{\textwidth}{!}{%
        \begin{tabular}{c|c|cc|cc|cc|cc|cc}
            \hline
            \multirow{2}{*}{Dataset} &\multirow{2}{*}{Method} &  \multicolumn{2}{c|}{Pixel-level}   & \multicolumn{2}{c|}{Perceptual-level} 
 & \multicolumn{2}{c|}{CRNN} & \multicolumn{2}{c|}{MORAN} & \multicolumn{2}{c}{PaddleOCR} \\
            \cline{3-12}
             & & PSNR$^{\uparrow}$ & SSIM$^{\uparrow}$ & FID$_{\text{crop}}^{\downarrow}$ & U-IDS$_{\text{crop}}^{\uparrow}$ &  Char Acc$^{\uparrow}$ & Word Acc$^{\uparrow}$ &  Char Acc$^{\uparrow}$ & Word Acc$^{\uparrow}$ &  Char Acc$^{\uparrow}$ & Word Acc$^{\uparrow}$ \\
            \hline
            \multirow{11}{*}{\rotatebox{90}{BCSD}}&Input &18.60&0.8288&143.33&0.00\%&45.68\%&11.42\%&35.43\%&12.56\%&72.16\%&26.26\%\\
            \cline{2-12}
            &AOT (TVCG 2023) & 29.14 & 0.9309 & 69.10  & 0.00\% & 51.11\% & 16.96\% & 41.52\% & 16.31\% & 84.55\% & 49.10\% \\
            &GRIG (ArXiv 2023) & 34.82 & 0.9848 & 5.61  & 12.60\% & 54.93\% & 20.72\% & 43.88\% & 18.92\% & 89.27\% & 62.81\% \\
            &DE-GAN (TPAMI 2022) & 29.83 & 0.9349 & 56.46 & 0.00\% & 54.02\% & 20.06\% & 42.69\% & 18.59\%  & 86.23\% & 51.38\% \\
            &DocDiff (ACM MM 2023) & 33.13 & 0.9772 & 12.54 & 0.94\% & 54.16\% & 20.39\% & 43.48\% & 18.11\% & 87.53\% & 56.93\% \\
            &DocRes (CVPR 2024) & {37.48} & {0.9886} & {3.79} & 20.43\% & {55.91\%} & \textbf{23.33\%} & {44.77\%} & {19.58\%} & 88.71\% & 63.13\% \\
            &GSDM (AAAI 2024) & 33.04 & 0.9693 & 8.88 & 3.27\% & 53.93\% & 20.72\% & 43.01\% & 17.94\% & 87.55\% & 54.65\% \\
            \cline{2-12}
            &TextDoctor (CNN)&33.28&0.9639&26.38&0.00\%&{55.73\%}&{22.84\%}&{44.15\%}&\textbf{19.74\%}&\textbf{89.99\%}&{61.83\%}\\
            &TextDoctor (Trans)&\textbf{38.12}&\textbf{0.9894}&4.18&17.44\%&55.34\%&22.02\%&44.75\%&\textbf{20.23\%}&\textbf{89.73\%}&\textbf{65.74\%}\\
            &TextDoctor (Mamba)&\textbf{38.73}&\textbf{0.9908}&\textbf{2.25}&\textbf{25.96\%}&55.43\%&22.19\%&44.36\%&\textbf{20.55\%}&89.26\%&\textbf{64.44\%}\\
            &TextDoctor (GSDM)&\textbf{39.12}&\textbf{0.9902}&7.86&{11.66\%}&\textbf{56.10\%}&{22.84\%}&\textbf{45.16\%}&\textbf{20.55\%}&\textbf{89.73\%}&\textbf{66.23\%}\\
            \hline
        \end{tabular}
    }
\end{table*}

\begin{table*}[!ht]
    \centering
    \caption{Comparison to state-of-the-art methods on the other six datasets (Twitter posts).  \textbf{Bold} indicates our methods outperform SOTA methods or the best-performing method.}
    \label{tab:ori_comparison_all_2}
    \resizebox{\textwidth}{!}{%
        \begin{tabular}{c|c|cc|cc|cc|cc|cc}
            \hline
            \multirow{2}{*}{Dataset} &\multirow{2}{*}{Method} &  \multicolumn{2}{c|}{Pixel-level}   & \multicolumn{2}{c|}{Perceptual-level} 
 & \multicolumn{2}{c|}{CRNN} & \multicolumn{2}{c|}{MORAN} & \multicolumn{2}{c}{PaddleOCR} \\
            \cline{3-12}
             & & PSNR$^{\uparrow}$ & SSIM$^{\uparrow}$ & FID$_{\text{crop}}^{\downarrow}$ & U-IDS$_{\text{crop}}^{\uparrow}$ &  Char Acc$^{\uparrow}$ & Word Acc$^{\uparrow}$ &  Char Acc$^{\uparrow}$ & Word Acc$^{\uparrow}$ &  Char Acc$^{\uparrow}$ & Word Acc$^{\uparrow}$ \\
            \hline
            \multirow{11}{*}{\rotatebox{90}{Twitter posts}}&Input&16.78&0.7705&193.93&0.00\%&50.66\%&19.51\%&40.71\%&16.10\%&79.14\%&36.95\%\\
            \cline{2-12}
             &AOT (TVCG 2023) & 30.60 & 0.9828 & 28.18 & 0.61\% & 55.43\% & 25.85\% & 45.77\% & 19.02\% & 87.23\% & 62.56\% \\
            &GRIG (ArXiv 2023) & 31.32 & 0.9881 &17.93 & 0.00\% & 55.12\% & 25.85\% & 45.52\% & 19.02\% & 86.92\% & 63.05\% \\
            &DE-GAN (TPAMI 2022) & 25.51 & 0.7621 &113.72 & 0.00\% & 53.24\%  & 23.41\%  & 44.64\%  &  16.58\%  & 83.31\%  & 52.71\% \\
            &DocDiff (ACM MM 2023) & 32.84  &  0.9832  &  28.89  &  0.00\%  & 54.44\% & 25.85\% & 45.29\% & 18.53\% & 84.99\% &  55.66\% \\
            &DocRes (CVPR 2024) & {37.31} & \textbf{0.9933} & \textbf{6.20} & \textbf{0.91\%} & 55.33\% & 25.85\% & 45.84\% & 19.51\% & 87.31\% & 65.02\% \\
            &GSDM (AAAI 2024) & 28.33 & 0.9561 & 41.44 & 0.00\% & 53.07\% & 22.93\% & 43.42\% & 16.10\% & 84.16\% & 56.16\% \\
            \cline{2-12}
            &TextDoctor (CNN)&31.20&0.8784&93.59&0.00\%&\textbf{55.84\%}&\textbf{27.80\%}&\textbf{46.76\%}&\textbf{21.46\%}&\textbf{88.26\%}&\textbf{71.92\%}\\
            &TextDoctor (Trans)&35.83&0.9769&25.35&0.00\%&\textbf{55.76\%}&\textbf{26.34\%}&\textbf{46.20\%}&\textbf{20.00\%}&\textbf{88.00\%}&\textbf{67.49\%}\\
            &TextDoctor (Mamba)&36.97&0.9165&12.73&0.00\%&\textbf{56.03\%}&\textbf{27.80\%}&\textbf{46.71\%}&\textbf{20.49\%}&\textbf{87.84\%}&\textbf{70.94\%}\\
            &TextDoctor (GSDM)&\textbf{38.04}&0.9811&41.01&0.00\%&\textbf{55.64\%}&\textbf{27.80\%}&\textbf{46.88\%}&\textbf{21.46\%}&\textbf{87.72\%}&\textbf{74.88\%}\\
            \hline
        \end{tabular}
    }
\end{table*}

\begin{table*}[!ht]
    \centering
    \caption{Comparison to state-of-the-art methods on the other six datasets (Twitter profiles).  \textbf{Bold} indicates our methods outperform SOTA methods or the best-performing method.}
    \label{tab:ori_comparison_all_3}
    \resizebox{\textwidth}{!}{%
        \begin{tabular}{c|c|cc|cc|cc|cc|cc}
            \hline
            \multirow{2}{*}{Dataset} &\multirow{2}{*}{Method} &  \multicolumn{2}{c|}{Pixel-level}   & \multicolumn{2}{c|}{Perceptual-level} 
 & \multicolumn{2}{c|}{CRNN} & \multicolumn{2}{c|}{MORAN} & \multicolumn{2}{c}{PaddleOCR} \\
            \cline{3-12}
             & & PSNR$^{\uparrow}$ & SSIM$^{\uparrow}$ & FID$_{\text{crop}}^{\downarrow}$ & U-IDS$_{\text{crop}}^{\uparrow}$ &  Char Acc$^{\uparrow}$ & Word Acc$^{\uparrow}$ &  Char Acc$^{\uparrow}$ & Word Acc$^{\uparrow}$ &  Char Acc$^{\uparrow}$ & Word Acc$^{\uparrow}$ \\
            \hline
            \multirow{11}{*}{\rotatebox{90}{Twitter profiles}}&Input&17.42&0.7464&88.09&21.42\%&44.39\%&9.87\%&36.09\%&10.99\%&83.13\%&23.25\%\\
            \cline{2-12}
             &AOT (TVCG 2023) & 30.91 & 0.9781 & 12.51 & 22.66\% & 52.71\% & 15.28\% & 42.75\% & 14.33\% & 94.24\% &65.13\% \\
            &GRIG (ArXiv 2023) & 32.40 & 0.9881 & 7.19 & 22.00\% &52.41\% & 14.49\% & 43.09\% & 14.65\% & 94.49\% &66.40\% \\
            &DE-GAN (TPAMI 2022) & 31.04 & 0.9164 & 30.15 & 21.42\% &  52.53\% & 14.64\% & 42.66\% & 14.33\% & 93.08\%  & 60.51\% \\
           & DocDiff (ACM MM 2023) & 34.82 & 0.9885 & 8.84 & 21.58\% & 52.93\% & 14.97\% & 43.63\% & 14.81\% & 94.55\% & 63.38\% \\
            &DocRes (CVPR 2024) & 38.81 & \textbf{0.9938} & 4.16 & 24.03\% & 53.04\% & 14.33\% & 43.32\% & 14.49\% & 94.73\% & 68.63\% \\
            &GSDM (AAAI 2024) & 34.98 & 0.9858 & 7.66 & 21.43\% & 53.14\% & 15.45\% & 42.86\% & {15.13\%} & 94.38\% & 68.95\% \\
            \cline{2-12}
            &TextDoctor (CNN)&33.02&0.9369&49.21&21.42\%&\textbf{53.71\%}&\textbf{15.45\%}&\textbf{44.06\%}&\textbf{15.13\%}&\textbf{95.31\%}&\textbf{73.25\%}\\
            &TextDoctor (Trans)&38.05&0.9912&7.32&21.60\%&\textbf{53.51\%}&15.13\%&\textbf{43.73\%}&14.49\%&\textbf{95.04\%}&\textbf{71.34\%}\\
            &TextDoctor (Mamba)&38.75&0.9809&\textbf{3.16}&\textbf{30.37\%}&\textbf{53.74\%}&\textbf{15.76\%}&\textbf{43.73\%}&14.97\%&\textbf{96.03\%}&\textbf{74.20\%}\\
            &TextDoctor (GSDM)&\textbf{39.36}&0.9883&19.98&21.42\%&\textbf{53.73\%}&\textbf{15.61\%}&\textbf{44.04\%}&\textbf{15.29\%}&\textbf{96.19\%}&\textbf{75.80\%}\\
            \hline
        \end{tabular}
    }
\end{table*}

\begin{table*}[!ht]
    \centering
    \caption{Comparison to state-of-the-art methods on the other six datasets (Activity diagrams).  \textbf{Bold} indicates our methods outperform SOTA methods or the best-performing method.}
    \label{tab:ori_comparison_all_4}
    \resizebox{\textwidth}{!}{%
        \begin{tabular}{c|c|cc|cc|cc|cc|cc}
            \hline
            \multirow{2}{*}{Dataset} &\multirow{2}{*}{Method} &  \multicolumn{2}{c|}{Pixel-level}   & \multicolumn{2}{c|}{Perceptual-level} 
 & \multicolumn{2}{c|}{CRNN} & \multicolumn{2}{c|}{MORAN} & \multicolumn{2}{c}{PaddleOCR} \\
            \cline{3-12}
             & & PSNR$^{\uparrow}$ & SSIM$^{\uparrow}$ & FID$_{\text{crop}}^{\downarrow}$ & U-IDS$_{\text{crop}}^{\uparrow}$ &  Char Acc$^{\uparrow}$ & Word Acc$^{\uparrow}$ &  Char Acc$^{\uparrow}$ & Word Acc$^{\uparrow}$ &  Char Acc$^{\uparrow}$ & Word Acc$^{\uparrow}$ \\
            \hline
            \multirow{11}{*}{\rotatebox{90}{Activity diagrams}}&Input&16.47&0.5312&253.85&0.00\%&51.26\%&10.55\%&47.56\%&12.24\%&62.88\%&17.72\%\\
            \cline{2-12}
             &AOT (TVCG 2023) & 32.74 & 0.9806 & 84.25 & 0.00\% & 54.39\% & 15.61\% & 53.88\% & 18.14\% & 74.11\% & 31.22\% \\
            &GRIG (ArXiv 2023) & 32.91 & 0.9862 & 93.24 & 0.00\% & 55.51\% & 15.61\% & 53.52\% & 16.03\% & 76.09\% & 31.22\% \\
            &DE-GAN (TPAMI 2022) & 31.54 & 0.9716 & 141.48 & 0.00\% & 53.63\% & 14.34\% &51.07\% & 14.76\% &72.55\% &29.11\% \\
            &DocDiff (ACM MM 2023) & 35.25 & 0.9913 & 66.43 & 0.00\% & 57.00\% & 16.88\% & 54.34\% & 16.88\% & 78.22\% & 39.24\% \\
            &DocRes (CVPR 2024) & 34.75 & 0.9909 & 30.96 & 0.00\% & 55.50\% & 15.61\% & 52.31\% & 14.35\% & 76.66\% & 33.33\% \\
            &GSDM (AAAI 2024) & 41.81 & \textbf{0.9968} & 60.68 & 0.00\% & 62.76\% & 21.94\% & 59.31\% & 25.74\% & 83.66\% & 52.74\% \\
            \cline{2-12}
            &TextDoctor (CNN)&35.02&0.9818&130.80&0.00\%&\textbf{62.97\%}&\textbf{27.43\%}&\textbf{60.82\%}&\textbf{25.74\%}&\textbf{83.77\%}&\textbf{54.01\%}\\
            &TextDoctor (Trans)&40.56&0.9956&{35.63}&0.00\%&\textbf{63.89\%}&\textbf{25.32\%}&\textbf{60.46\%}&\textbf{27.00\%}&83.29\%&\textbf{53.59\%}\\
            &TextDoctor (Mamba)&41.39&0.9948&\textbf{17.30}&0.00\%&\textbf{63.86\%}&\textbf{25.74\%}&\textbf{60.13\%}&\textbf{27.43\%}&83.33\%&\textbf{52.74\%}\\
            &TextDoctor (GSDM)&\textbf{43.66}&0.9967&74.94&0.00\%&\textbf{63.83\%}&\textbf{24.47\%}&\textbf{61.17\%}&\textbf{28.69\%}&\textbf{83.70\%}&\textbf{54.43\%}\\
            \hline
        \end{tabular}
    }
\end{table*}

\begin{table*}[!ht]
    \centering
    \caption{Comparison to state-of-the-art methods on the other six datasets (Signatures).  \textbf{Bold} indicates our methods outperform SOTA methods or the best-performing method.}
    \label{tab:ori_comparison_all_5}
    \resizebox{\textwidth}{!}{%
        \begin{tabular}{c|c|cc|cc|cc|cc|cc}
            \hline
            \multirow{2}{*}{Dataset} &\multirow{2}{*}{Method} &  \multicolumn{2}{c|}{Pixel-level}   & \multicolumn{2}{c|}{Perceptual-level} 
 & \multicolumn{2}{c|}{CRNN} & \multicolumn{2}{c|}{MORAN} & \multicolumn{2}{c}{PaddleOCR} \\
            \cline{3-12}
             & & PSNR$^{\uparrow}$ & SSIM$^{\uparrow}$ & FID$_{\text{crop}}^{\downarrow}$ & U-IDS$_{\text{crop}}^{\uparrow}$ &  Char Acc$^{\uparrow}$ & Word Acc$^{\uparrow}$ &  Char Acc$^{\uparrow}$ & Word Acc$^{\uparrow}$ &  Char Acc$^{\uparrow}$ & Word Acc$^{\uparrow}$ \\
            \hline
            \multirow{11}{*}{\rotatebox{90}{Signatures}}&Input&17.21&0.7219&151.91&0.00\%&31.33\%&4.00\%&26.41\%&4.35\%&59.31\%&13.07\%\\
            \cline{2-12}
            & AOT (TVCG 2023) & 28.55 & 0.9649 & 27.20 & 1.35\% & 31.21\% & 4.23\% & 27.79\% & 4.94\% & 66.68\% & 16.96\% \\
            &GRIG (ArXiv 2023) & 31.93 & 0.9824 & 10.55 & 1.35\% & 33.11\% & 5.52\% & 29.59\% & 6.24\% & 70.86\% & 21.55\% \\
            &DE-GAN (TPAMI 2022) & 28.06 & 0.9309 & 45.28 & 0.00\% & 32.28\% &  4.94\% & 27.34\% & 4.71\%  &69.06\% & 19.19\% \\
            &DocDiff (ACM MM 2023) & 31.56 & 0.9821 & 13.40 & 0.00\% & 33.70\% & 5.53\% & 30.32\% & 6.24\% & 73.03\% & 24.50\% \\
            &DocRes (CVPR 2024) & 33.83 & 0.9874 & 10.25 & 0.00\% & 33.87\% & 6.12\% & 30.54\% & 6.82\% & 74.37\% & 26.27\% \\
            &GSDM (AAAI 2024) & 34.04 & 0.9880 & 23.44 &  0.00\%& 34.36\% & 6.12\% & 31.00\% & 7.29\% & 74.93\% & 27.92\% \\
            \cline{2-12}
            &TextDoctor (CNN)&32.37&0.9600&86.41&0.00\%&\textbf{35.65\%}&\textbf{6.82\%}&\textbf{31.34\%}&\textbf{8.12\%}&\textbf{77.97\%}&\textbf{36.16\%}\\
            &TextDoctor (Trans)&\textbf{36.83}&\textbf{0.9935}&13.16&0.00\%&\textbf{35.61\%}&\textbf{7.29\%}&\textbf{31.92\%}&\textbf{8.71\%}&\textbf{77.93\%}&\textbf{36.04\%}\\
            &TextDoctor (Mamba)&\textbf{37.93}&\textbf{0.9949}&\textbf{3.88}&\textbf{2.14\%}&\textbf{35.83\%}&\textbf{7.41\%}&\textbf{32.00\%}&\textbf{8.12\%}&\textbf{78.13\%}&\textbf{36.75\%}\\
            &TextDoctor (GSDM)&\textbf{38.28}&\textbf{0.9924}&37.49&0.00\%&\textbf{35.94\%}&\textbf{7.76\%}&\textbf{32.47\%}&\textbf{8.00\%}&\textbf{78.65\%}&\textbf{38.75\%}\\
            \hline
        \end{tabular}
    }
\end{table*}

\begin{table*}[!ht]
    \centering
    \caption{Comparison to state-of-the-art methods on the other six datasets (Tabular).  \textbf{Bold} indicates our methods outperform SOTA methods or the best-performing method.}
    \label{tab:ori_comparison_all_6}
    \resizebox{\textwidth}{!}{%
        \begin{tabular}{c|c|cc|cc|cc|cc|cc}
            \hline
            \multirow{2}{*}{Dataset} &\multirow{2}{*}{Method} &  \multicolumn{2}{c|}{Pixel-level}   & \multicolumn{2}{c|}{Perceptual-level} 
 & \multicolumn{2}{c|}{CRNN} & \multicolumn{2}{c|}{MORAN} & \multicolumn{2}{c}{PaddleOCR} \\
            \cline{3-12}
             & & PSNR$^{\uparrow}$ & SSIM$^{\uparrow}$ & FID$_{\text{crop}}^{\downarrow}$ & U-IDS$_{\text{crop}}^{\uparrow}$ &  Char Acc$^{\uparrow}$ & Word Acc$^{\uparrow}$ &  Char Acc$^{\uparrow}$ & Word Acc$^{\uparrow}$ &  Char Acc$^{\uparrow}$ & Word Acc$^{\uparrow}$ \\
            \hline
            \multirow{11}{*}{\rotatebox{90}{Tabular}}&Input &16.42&0.6794&126.73&0.00\%&45.86\%&11.14\%&41.52\%&10.87\%&77.56\%&39.16\%\\
             \cline{2-12}       
             &AOT (TVCG 2023) & 28.23 & 0.9685 & 25.01 & 0.00\% & 52.30\% & 14.04\% & 48.07\% & 12.75\% & 86.81\% & 57.07\% \\
            &GRIG (ArXiv 2023) & 29.97 & 0.9843 & 12.06 & 0.73\% & 54.01\% & 14.76\% & 49.25\% & 13.49\% & 87.98\% & 59.79\% \\
            &DE-GAN (TPAMI 2022) & 35.24 & 0.9907 & 16.28 & 0.00\% &  57.74\% &16.98\% & 52.27\% &15.06\% &91.03\%  & 71.15\%\\
            &DocDiff (ACM MM 2023) & 34.81 & 0.9922 & 10.10 & 0.12\% & 58.02\% & 16.90\% & 52.71\% & 15.16\% & 91.12\% & 70.63\% \\
            &DocRes (CVPR 2024) & 35.80 & 0.9938 & 8.11 & 0.49\% & 57.53\% & 16.65\% & 52.11\% & 14.77\% & 90.75\% & 69.68\% \\
            &GSDM (AAAI 2024) & 31.71 & 0.9877 & 8.67 & 0.00\% & 54.60\% & 15.53\% & 49.30\% & 13.70\% & 89.20\% & 64.72\% \\
             \cline{2-12}
            &TextDoctor (CNN)&32.44&0.9693&55.83&0.00\%&57.40\%&\textbf{17.60\%}&51.93\%&\textbf{15.33\%}&\textbf{92.05\%}&\textbf{76.07\%}\\
            &TextDoctor (Trans)&\textbf{36.81}&\textbf{0.9950}&9.53&0.00\%&57.87\%&\textbf{17.51\%}&52.50\%&\textbf{15.37\%}&\textbf{92.15\%}&\textbf{76.20\%}\\
            &TextDoctor (Mamba)&\textbf{37.72}&\textbf{0.9963}&\textbf{1.46}&\textbf{5.22\%}&57.86\%&\textbf{17.60\%}&52.47\%&\textbf{15.69\%}&\textbf{92.21\%}&\textbf{76.89\%}\\
            &TextDoctor (GSDM)&\textbf{39.20}&\textbf{0.9946}&24.82&0.00\%&\textbf{58.54\%}&\textbf{18.25\%}&\textbf{53.29\%}&\textbf{16.12\%}&\textbf{92.66\%}&\textbf{79.26\%}\\
            \hline
        \end{tabular}
    }
\end{table*}

\begin{table*}[!ht]
    \centering
    \caption{Comparison with SOTA methods on the seven document image datasets (FUNSD). All the methods were trained on the same text patch image dataset. \textbf{Bold} indicates our methods outperform SOTA methods or the best-performing method.}
    \label{tab:patch_all_comparison_1}
    \resizebox{\textwidth}{!}{%
        \begin{tabular}{c|c|cc|cc|cc|cc|cc}
            \hline
             \multirow{2}{*}{Dataset} &\multirow{2}{*}{Method} &  \multicolumn{2}{c|}{Pixel-level}   & \multicolumn{2}{c|}{Perceptual-level} 
 & \multicolumn{2}{c|}{CRNN} & \multicolumn{2}{c|}{MORAN} & \multicolumn{2}{c}{PaddleOCR} \\
            \cline{3-12}
             &  & PSNR$^{\uparrow}$ & SSIM$^{\uparrow}$ & FID$_{\text{crop}}^{\downarrow}$ & U-IDS$_{\text{crop}}^{\uparrow}$ &  Char Acc$^{\uparrow}$ & Word Acc$^{\uparrow}$ &  Char Acc$^{\uparrow}$ & Word Acc$^{\uparrow}$ &  Char Acc$^{\uparrow}$ & Word Acc$^{\uparrow}$ \\
            \hline
            \multirow{10}{*}{\rotatebox{90}{FUNSD}}&AOT (TVCG 2023)&26.52&0.9760&25.08&0.00\%&51.02\%&14.37\%&45.26\%&12.95\%&77.06\%&33.61\%\\
            &GRIG (ArXiv 2023) &25.57&0.9404&61.97&0.00\%&50.69\%&13.51\%&44.69\%&12.31\%&75.99\%&32.58\%\\
            &DE-GAN (TPAMI 2022)&29.02&0.9722&22.73&0.00\%&52.16\%&14.62\%&46.35\%&13.34\%&79.89\%&37.14\%\\
            &DocDiff (ACM MM 2023)&28.07&0.9808&17.80&0.00\%&51.76\%&14.37\%&45.70\%&13.12\%&78.43\%&36.50\%\\
            &DocRes (CVPR 2024)&22.60&0.8824&89.36&0.00\%&50.29\%&13.08\%&44.05\%&12.09\%&75.90\%&31.71\%\\
            &GSDM (AAAI 2024) & 28.43 & 0.9855 & 7.00 & 0.00\% & 51.28\% & 14.16\% & 45.72\% & 13.21\% & 80.63\% & 39.63\% \\
            \cline{2-12}       
            &TextDoctor (CNN)&\textbf{29.88}&0.9301&78.70&0.00\%&\textbf{55.94\%}&\textbf{17.98\%}&\textbf{49.80\%}&\textbf{16.42\%}&\textbf{86.74\%}&\textbf{53.03\%}\\
            &TextDoctor (Trans)&\textbf{34.20}&\textbf{0.9958}&7.44&0.00\%&\textbf{56.23\%}&\textbf{18.14\%}&\textbf{49.86\%}&\textbf{16.04\%}&\textbf{86.14\%}&\textbf{52.18\%}\\
            &TextDoctor (Mamba)&\textbf{35.03}&\textbf{0.9966}&\textbf{0.80}&\textbf{21.37\%}&\textbf{56.48\%}&\textbf{18.10\%}&\textbf{49.98\%}&\textbf{16.04\%}&\textbf{86.53\%}&\textbf{53.50\%}\\
            &TextDoctor (GSDM)&\textbf{35.44}&\textbf{0.9907}&25.61&0.00\%&\textbf{56.97\%}&\textbf{19.17\%}&\textbf{50.28\%}&\textbf{16.64\%}&\textbf{86.92\%}&\textbf{55.38\%}\\
            \hline
        \end{tabular}
    }
\end{table*}

\begin{table*}[!ht]
    \centering
    \caption{Comparison with SOTA methods on the seven document image datasets (BCSD). All the methods were trained on the same text patch image dataset. \textbf{Bold} indicates our methods outperform SOTA methods or the best-performing method.}
    \label{tab:patch_all_comparison_2}
    \resizebox{\textwidth}{!}{%
        \begin{tabular}{c|c|cc|cc|cc|cc|cc}
            \hline
             \multirow{2}{*}{Dataset} &\multirow{2}{*}{Method} &  \multicolumn{2}{c|}{Pixel-level}   & \multicolumn{2}{c|}{Perceptual-level} 
 & \multicolumn{2}{c|}{CRNN} & \multicolumn{2}{c|}{MORAN} & \multicolumn{2}{c}{PaddleOCR} \\
            \cline{3-12}
             &  & PSNR$^{\uparrow}$ & SSIM$^{\uparrow}$ & FID$_{\text{crop}}^{\downarrow}$ & U-IDS$_{\text{crop}}^{\uparrow}$ &  Char Acc$^{\uparrow}$ & Word Acc$^{\uparrow}$ &  Char Acc$^{\uparrow}$ & Word Acc$^{\uparrow}$ &  Char Acc$^{\uparrow}$ & Word Acc$^{\uparrow}$ \\
            \hline
            \multirow{10}{*}{\rotatebox{90}{BCSD}}&AOT (TVCG 2023)&32.29&0.9726&21.62&0.03\%&54.11\%&21.21\%&43.76\%&20.23\%&88.20\%&60.20\%\\
            &GRIG (ArXiv 2023)&31.04&0.9539&39.07&0.00\%&53.18\%&18.76\%&42.24\%&17.78\%&86.30\%&54.32\%\\
            &DE-GAN (TPAMI 2022)&33.33&0.9726&16.91&0.01\%&54.83\%&20.88\%&43.99\%&19.74\%&87.75\%&59.87\%\\
            &DocDiff (ACM MM 2023)&32.35&0.9696&17.27&0.06\%&53.22\%&19.58\%&42.89\%&17.78\%&86.52\%&54.98\%\\
            &DocRes (CVPR 2024)&29.64&0.9241&76.64&0.00\%&51.10\%&16.80\%&42.08\%&16.31\%&83.04\%&45.02\%\\
            &GSDM (AAAI 2024) & 33.04 & 0.9693 & 8.88 & 3.27\% & 53.93\% & 20.72\% & 43.01\% & 17.94\% & 87.55\% & 54.65\% \\
            \cline{2-12}  
            &TextDoctor (CNN)&{33.28}&0.9639&26.38&0.00\%&\textbf{55.73\%}&\textbf{22.84\%}&\textbf{44.15\%}&{19.74\%}&\textbf{89.99\%}&\textbf{61.83\%}\\
            &TextDoctor (Trans)&\textbf{38.12}&\textbf{0.9894}&\textbf{4.18}&\textbf{17.44\%}&\textbf{55.34\%}&\textbf{22.02\%}&\textbf{44.75\%}&\textbf{20.23\%}&\textbf{89.73\%}&\textbf{65.74\%}\\
            &TextDoctor (Mamba)&\textbf{38.73}&\textbf{0.9908}&\textbf{2.25}&\textbf{25.96\%}&\textbf{55.43\%}&\textbf{22.19\%}&\textbf{44.36\%}&\textbf{20.55\%}&\textbf{89.26\%}&\textbf{64.44\%}\\
            &TextDoctor (GSDM)&\textbf{39.12}&\textbf{0.9902}&\textbf{7.86}&\textbf{11.66\%}&\textbf{56.10\%}&\textbf{22.84\%}&\textbf{45.16\%}&\textbf{20.55\%}&\textbf{89.73\%}&\textbf{66.23\%}\\
            \hline
        \end{tabular}
    }
\end{table*}

\begin{table*}[!ht]
    \centering
    \caption{Comparison with SOTA methods on the seven document image datasets (Twitter posts). All the methods were trained on the same text patch image dataset. \textbf{Bold} indicates our methods outperform SOTA methods or the best-performing method.}
    \label{tab:patch_all_comparison_3}
    \resizebox{\textwidth}{!}{%
        \begin{tabular}{c|c|cc|cc|cc|cc|cc}
            \hline
             \multirow{2}{*}{Dataset} &\multirow{2}{*}{Method} &  \multicolumn{2}{c|}{Pixel-level}   & \multicolumn{2}{c|}{Perceptual-level} 
 & \multicolumn{2}{c|}{CRNN} & \multicolumn{2}{c|}{MORAN} & \multicolumn{2}{c}{PaddleOCR} \\
            \cline{3-12}
             &  & PSNR$^{\uparrow}$ & SSIM$^{\uparrow}$ & FID$_{\text{crop}}^{\downarrow}$ & U-IDS$_{\text{crop}}^{\uparrow}$ &  Char Acc$^{\uparrow}$ & Word Acc$^{\uparrow}$ &  Char Acc$^{\uparrow}$ & Word Acc$^{\uparrow}$ &  Char Acc$^{\uparrow}$ & Word Acc$^{\uparrow}$ \\
            \hline
            \multirow{10}{*}{\rotatebox{90}{Twitter posts}}&AOT (TVCG 2023)&29.37&0.9516&63.03&0.00\%&53.93\%&23.41\%&44.10\%&17.56\%&83.88\%&54.19\%\\
            &GRIG (ArXiv 2023)&27.11&0.8820&116.46&0.00\%&51.44\%&21.95\%&43.37\%&15.61\%&84.57\%&49.26\%\\
            &DE-GAN (TPAMI 2022)&25.51&0.7621&113.72&0.00\%&53.24\%&23.41\%&44.64\%&16.59\%&83.32\%&52.71\%\\
            &DocDiff (ACM MM 2023)&31.12&0.9774&41.83&0.00\%&53.34\%&24.39\%&43.22\%&15.61\%&85.09\%&54.68\%\\
            &DocRes (CVPR 2024)&22.53&0.8121&170.52&0.00\%&47.24\%&18.05\%&37.90\%&12.68\%&70.64\%&33.00\%\\  
            &GSDM (AAAI 2024) & 28.33 & 0.9561 & 41.44 & 0.00\% & 53.07\% & 22.93\% & 43.42\% & 16.10\% & 84.16\% & 56.16\% \\
            \cline{2-12} 
            &TextDoctor (CNN)&\textbf{31.20}&0.8784&93.59&0.00\%&\textbf{55.84\%}&\textbf{27.80\%}&\textbf{46.76\%}&\textbf{21.46\%}&\textbf{88.26\%}&\textbf{71.92\%}\\
            &TextDoctor (Trans)&\textbf{35.83}&0.9769&\textbf{25.35}&0.00\%&\textbf{55.76\%}&\textbf{26.34\%}&\textbf{46.20\%}&\textbf{20.00\%}&\textbf{88.00\%}&\textbf{67.49\%}\\
            &TextDoctor (Mamba)&\textbf{36.97}&0.9165&\textbf{12.73}&0.00\%&\textbf{56.03\%}&\textbf{27.80\%}&\textbf{46.71\%}&\textbf{20.49\%}&\textbf{87.84\%}&\textbf{70.94\%}\\
            &TextDoctor (GSDM)&\textbf{38.04}&\textbf{0.9811}&\textbf{41.01}&0.00\%&\textbf{55.64\%}&\textbf{27.80\%}&\textbf{46.88\%}&\textbf{21.46\%}&\textbf{87.72\%}&\textbf{74.88\%}\\
            \hline
        \end{tabular}
    }
\end{table*}

\begin{table*}[!ht]
    \centering
    \caption{Comparison with SOTA methods on the seven document image datasets (Twitter profiles). All the methods were trained on the same text patch image dataset. \textbf{Bold} indicates our methods outperform SOTA methods or the best-performing method.}
    \label{tab:patch_all_comparison_4}
    \resizebox{\textwidth}{!}{%
        \begin{tabular}{c|c|cc|cc|cc|cc|cc}
            \hline
             \multirow{2}{*}{Dataset} &\multirow{2}{*}{Method} &  \multicolumn{2}{c|}{Pixel-level}   & \multicolumn{2}{c|}{Perceptual-level} 
 & \multicolumn{2}{c|}{CRNN} & \multicolumn{2}{c|}{MORAN} & \multicolumn{2}{c}{PaddleOCR} \\
            \cline{3-12}
             &  & PSNR$^{\uparrow}$ & SSIM$^{\uparrow}$ & FID$_{\text{crop}}^{\downarrow}$ & U-IDS$_{\text{crop}}^{\uparrow}$ &  Char Acc$^{\uparrow}$ & Word Acc$^{\uparrow}$ &  Char Acc$^{\uparrow}$ & Word Acc$^{\uparrow}$ &  Char Acc$^{\uparrow}$ & Word Acc$^{\uparrow}$ \\
            \hline
            \multirow{10}{*}{\rotatebox{90}{Twitter profiles}}&AOT (TVCG 2023)&30.26&0.9773&20.01&21.43\%&51.78\%&14.17\%&42.17\%&13.38\%&93.40\%&61.78\%\\
            &GRIG (ArXiv 2023)&28.29&0.9316&44.35&21.42\%&50.00\%&12.26\%&41.61\%&13.22\%&91.96\%&51.11\%\\
            &DE-GAN (TPAMI 2022)&33.68&0.9326&20.04&21.42\%&52.43\%&15.29\%&42.83\%&13.85\%&93.06\%&61.62\%\\
            &DocDiff (ACM MM 2023)&32.98&0.9821&15.79&21.43\%&52.06\%&14.33\%&42.78\%&13.85\%&93.94\%&61.31\%\\
            &DocRes (CVPR 2024)&24.27&0.8372&76.67&21.42\%&44.71\%&10.99\%&37.55\%&12.10\%&84.96\%&35.99\%\\
            &GSDM (AAAI 2024) & 34.98 & 0.9858 & 7.66 & 21.43\% & 53.14\% & 15.45\% & 42.86\% & {15.13\%} & 94.38\% & 68.95\% \\
            \cline{2-12} 
            &TextDoctor (CNN)&33.02&0.9369&49.21&21.42\%&\textbf{53.71\%}&\textbf{15.45\%}&\textbf{44.06\%}&\textbf{15.13\%}&\textbf{95.31\%}&\textbf{73.25\%}\\
            &TextDoctor (Trans)&\textbf{38.05}&\textbf{0.9912}&\textbf{7.32}&\textbf{21.60\%}&\textbf{53.51\%}&15.13\%&\textbf{43.73\%}&14.49\%&\textbf{95.04\%}&\textbf{71.34\%}\\
            &TextDoctor (Mamba)&\textbf{38.75}&0.9809&\textbf{3.16}&\textbf{30.37\%}&\textbf{53.74\%}&\textbf{15.76\%}&\textbf{43.73\%}&14.97\%&\textbf{96.03\%}&\textbf{74.20\%}\\
            &TextDoctor (GSDM)&\textbf{39.36}&\textbf{0.9883}&19.98&21.42\%&\textbf{53.73\%}&\textbf{15.61\%}&\textbf{44.04\%}&\textbf{15.29\%}&\textbf{96.19\%}&\textbf{75.80\%}\\
            \hline
        \end{tabular}
    }
\end{table*}

\begin{table*}[!ht]
    \centering
    \caption{Comparison with SOTA methods on the seven document image datasets (Activity diagrams). All the methods were trained on the same text patch image dataset. \textbf{Bold} indicates our methods outperform SOTA methods or the best-performing method.}
    \label{tab:patch_all_comparison_5}
    \resizebox{\textwidth}{!}{%
        \begin{tabular}{c|c|cc|cc|cc|cc|cc}
            \hline
             \multirow{2}{*}{Dataset} &\multirow{2}{*}{Method} &  \multicolumn{2}{c|}{Pixel-level}   & \multicolumn{2}{c|}{Perceptual-level} 
 & \multicolumn{2}{c|}{CRNN} & \multicolumn{2}{c|}{MORAN} & \multicolumn{2}{c}{PaddleOCR} \\
            \cline{3-12}
             &  & PSNR$^{\uparrow}$ & SSIM$^{\uparrow}$ & FID$_{\text{crop}}^{\downarrow}$ & U-IDS$_{\text{crop}}^{\uparrow}$ &  Char Acc$^{\uparrow}$ & Word Acc$^{\uparrow}$ &  Char Acc$^{\uparrow}$ & Word Acc$^{\uparrow}$ &  Char Acc$^{\uparrow}$ & Word Acc$^{\uparrow}$ \\
            \hline
            \multirow{10}{*}{\rotatebox{90}{Activity diagrams}}&AOT (TVCG 2023)&30.26&0.9672&49.82&\textbf{1.94\%}&52.86\%&13.08\%&51.21\%&14.35\%&72.12\%&29.54\%\\
            &GRIG (ArXiv 2023)&28.53&0.9301&169.86&0.00\%&51.35\%&12.24\%&49.61\%&13.50\%&71.05\%&23.63\%\\
            &DE-GAN (TPAMI 2022)&27.95&0.9044&169.42&0.00\%&52.96\%&11.81\%&51.06\%&15.61\%&71.53\%&24.05\%\\
            &DocDiff (ACM MM 2023)&31.25&0.9700&52.44&0.00\%&54.81\%&13.92\%&51.49\%&16.03\%&75.15\%&28.69\%\\
            &DocRes (CVPR 2024)&22.09&0.6633&204.50&0.00\%&50.09\%&11.39\%&47.01\%&12.66\%&62.47\%&18.57\%\\  
            &GSDM (AAAI 2024) & 41.81 & \textbf{0.9968} & 60.68 & 0.00\% & 62.76\% & 21.94\% & 59.31\% & 25.74\% & 83.66\% & 52.74\% \\
            \cline{2-12} 
            &TextDoctor (CNN)&35.02&0.9818&130.80&0.00\%&\textbf{62.97\%}&\textbf{27.43\%}&\textbf{60.82\%}&\textbf{25.74\%}&\textbf{83.77\%}&\textbf{54.01\%}\\
            &TextDoctor (Trans)&40.56&0.9956&\textbf{35.63}&0.00\%&\textbf{63.89\%}&\textbf{25.32\%}&\textbf{60.46\%}&\textbf{27.00\%}&83.29\%&\textbf{53.59\%}\\
            &TextDoctor (Mamba)&41.39&0.9948&\textbf{17.30}&0.00\%&\textbf{63.86\%}&\textbf{25.74\%}&\textbf{60.13\%}&\textbf{27.43\%}&83.33\%&\textbf{52.74\%}\\
            &TextDoctor (GSDM)&\textbf{43.66}&0.9967&74.94&0.00\%&\textbf{63.83\%}&\textbf{24.47\%}&\textbf{61.17\%}&\textbf{28.69\%}&\textbf{83.70\%}&\textbf{54.43\%}\\
            \hline
        \end{tabular}
    }
\end{table*}

\begin{table*}[!ht]
    \centering
    \caption{Comparison with SOTA methods on the seven document image datasets (Signatures). All the methods were trained on the same text patch image dataset. \textbf{Bold} indicates our methods outperform SOTA methods or the best-performing method.}
    \label{tab:patch_all_comparison_6}
    \resizebox{\textwidth}{!}{%
        \begin{tabular}{c|c|cc|cc|cc|cc|cc}
            \hline
             \multirow{2}{*}{Dataset} &\multirow{2}{*}{Method} &  \multicolumn{2}{c|}{Pixel-level}   & \multicolumn{2}{c|}{Perceptual-level} 
 & \multicolumn{2}{c|}{CRNN} & \multicolumn{2}{c|}{MORAN} & \multicolumn{2}{c}{PaddleOCR} \\
            \cline{3-12}
             &  & PSNR$^{\uparrow}$ & SSIM$^{\uparrow}$ & FID$_{\text{crop}}^{\downarrow}$ & U-IDS$_{\text{crop}}^{\uparrow}$ &  Char Acc$^{\uparrow}$ & Word Acc$^{\uparrow}$ &  Char Acc$^{\uparrow}$ & Word Acc$^{\uparrow}$ &  Char Acc$^{\uparrow}$ & Word Acc$^{\uparrow}$ \\
            \hline
            \multirow{10}{*}{\rotatebox{90}{Signatures}}&AOT (TVCG 2023)&29.75&0.9766&17.71&0.00\%&32.25\%&4.82\%&28.39\%&5.06\%&68.17\%&19.79\%\\
            &GRIG (ArXiv 2023)&28.52&0.9436&77.16&0.00\%&32.00\%&4.71\%&28.46\%&5.06\%&66.79\%&17.67\%\\
            &DE-GAN (TPAMI 2022)&29.61&0.9513&32.35&0.00\%&32.63\%&4.35\%&28.53\%&5.76\%&69.19\%&19.20\%\\
            &DocDiff (ACM MM 2023)&30.96&0.9747&21.21&0.00\%&32.76\%&5.18\%&28.92\%&5.88\%&69.56\%&21.20\%\\
            &DocRes (CVPR 2024)&25.63&0.8520&95.53&0.00\%&29.79\%&3.42\%&26.44\%&4.47\%&59.54\%&13.31\%\\
            &GSDM (AAAI 2024) & 34.04 & 0.9880 & 23.44 &  0.00\%& 34.36\% & 6.12\% & 31.00\% & 7.29\% & 74.93\% & 27.92\% \\
            \cline{2-12} 
            &TextDoctor (CNN)&32.37&0.9600&86.41&0.00\%&\textbf{35.65\%}&\textbf{6.82\%}&\textbf{31.34\%}&\textbf{8.12\%}&\textbf{77.97\%}&\textbf{36.16\%}\\
            &TextDoctor (Trans)&\textbf{36.83}&\textbf{0.9935}&\textbf{13.16}&0.00\%&\textbf{35.61\%}&\textbf{7.29\%}&\textbf{31.92\%}&\textbf{8.71\%}&\textbf{77.93\%}&\textbf{36.04\%}\\
            &TextDoctor (Mamba)&\textbf{37.93}&\textbf{0.9949}&\textbf{3.88}&\textbf{2.14\%}&\textbf{35.83\%}&\textbf{7.41\%}&\textbf{32.00\%}&\textbf{8.12\%}&\textbf{78.13\%}&\textbf{36.75\%}\\
            &TextDoctor (GSDM)&\textbf{38.28}&\textbf{0.9924}&37.49&0.00\%&\textbf{35.94\%}&\textbf{7.76\%}&\textbf{32.47\%}&\textbf{8.00\%}&\textbf{78.65\%}&\textbf{38.75\%}\\
            \hline
        \end{tabular}
    }
\end{table*}

\begin{table*}[!ht]
    \centering
    \caption{Comparison with SOTA methods on the seven document image datasets (Tabular). All the methods were trained on the same text patch image dataset. \textbf{Bold} indicates our methods outperform SOTA methods or the best-performing method.}
    \label{tab:patch_all_comparison_7}
    \resizebox{\textwidth}{!}{%
        \begin{tabular}{c|c|cc|cc|cc|cc|cc}
            \hline
             \multirow{2}{*}{Dataset} &\multirow{2}{*}{Method} &  \multicolumn{2}{c|}{Pixel-level}   & \multicolumn{2}{c|}{Perceptual-level} 
 & \multicolumn{2}{c|}{CRNN} & \multicolumn{2}{c|}{MORAN} & \multicolumn{2}{c}{PaddleOCR} \\
            \cline{3-12}
             &  & PSNR$^{\uparrow}$ & SSIM$^{\uparrow}$ & FID$_{\text{crop}}^{\downarrow}$ & U-IDS$_{\text{crop}}^{\uparrow}$ &  Char Acc$^{\uparrow}$ & Word Acc$^{\uparrow}$ &  Char Acc$^{\uparrow}$ & Word Acc$^{\uparrow}$ &  Char Acc$^{\uparrow}$ & Word Acc$^{\uparrow}$ \\
            \hline
            \multirow{10}{*}{\rotatebox{90}{Tabular}}&AOT (TVCG 2023)&26.30&0.9705&22.94&0.00\%&49.24\%&12.42\%&45.54\%&11.71\%&83.61\%&50.55\%\\
            &GRIG (ArXiv 2023)&25.96&0.9337&56.90&0.00\%&49.26\%&12.52\%&45.25\%&11.33\%&83.53\%&49.62\%\\
            &DE-GAN (TPAMI 2022)&28.91&0.9608&30.74&0.00\%&51.07\%&13.74\%&46.69\%&12.34\%&85.94\%&54.66\%\\
            &DocDiff (ACM MM 2023)&29.35&0.9756&23.67&0.00\%&51.31\%&13.61\%&46.59\%&12.31\%&85.52\%&54.04\%\\
            &DocRes (CVPR 2024)&22.14&0.8016&99.03&0.00\%&44.87\%&11.12\%&41.49\%&10.46\%&77.35\%&38.48\%\\
            &GSDM (AAAI 2024) & 31.71 & 0.9877 & 8.67 & 0.00\% & 54.60\% & 15.53\% & 49.30\% & 13.70\% & 89.20\% & 64.72\% \\
            \cline{2-12} 
            &TextDoctor (CNN)&\textbf{32.44}&0.9693&55.83&0.00\%&\textbf{57.40\%}&\textbf{17.60\%}&\textbf{51.93\%}&\textbf{15.33\%}&\textbf{92.05\%}&\textbf{76.07\%}\\
            &TextDoctor (Trans)&\textbf{36.81}&\textbf{0.9950}&9.53&0.00\%&\textbf{57.87\%}&\textbf{17.51\%}&\textbf{52.50\%}&\textbf{15.37\%}&\textbf{92.15\%}&\textbf{76.20\%}\\
            &TextDoctor (Mamba)&\textbf{37.72}&\textbf{0.9963}&\textbf{1.46}&\textbf{5.22\%}&\textbf{57.86\%}&\textbf{17.60\%}&\textbf{52.47\%}&\textbf{15.69\%}&\textbf{92.21\%}&\textbf{76.89\%}\\
            &TextDoctor (GSDM)&\textbf{39.20}&\textbf{0.9946}&24.82&0.00\%&\textbf{58.54\%}&\textbf{18.25\%}&\textbf{53.29\%}&\textbf{16.12\%}&\textbf{92.66\%}&\textbf{79.26\%}\\
            \hline
        \end{tabular}
    }
\end{table*}


\begin{table*}[!ht]
    \caption{Wilcoxon signed-rank test results ($p$-values) for the quantitative results from Table 1 in the main paper, Table~\ref{tab:ori_comparison_all_1}, Table~\ref{tab:ori_comparison_all_2}, Table~\ref{tab:ori_comparison_all_3}, Table~\ref{tab:ori_comparison_all_4}, Table~\ref{tab:ori_comparison_all_5}, and Table~\ref{tab:ori_comparison_all_6} across seven document image datasets. TDO: TextDoctor.}
    \label{tab:Wilcoxon_ori_comparison_all}
    \centering
    \resizebox{1.0\textwidth}{!}{$
    \begin{tabular}{c|c|c|c|c||c|c|c|c||c|c|c|c}
        \hline
        \multirow{2}{*}{Method}&\multicolumn{4}{c||}{PSNR} &\multicolumn{4}{c||}{SSIM} &\multicolumn{4}{c}{FID$_{\text{crop}}$} \\
        \cline{2-13}
        & TDO (CNN)    & TDO (Trans)  & TDO (Mamba)   & TDO (GSDM) & TDO (CNN)    & TDO (Trans)  & TDO (Mamba)   & TDO (GSDM) & TDO (CNN)    & TDO (Trans)  & TDO (Mamba)   & TDO (GSDM)  \\ \hline
        AOT        & 0.015625 & 0.015625  & 0.015625  & 0.015625 & 0.296875 & 0.031250   & 0.296875  & 0.031250 & 0.078125 & 0.015625  & 0.015625  & 0.687500   \\
        GRIG      & 0.218750 & 0.015625  & 0.015625  & 0.015625 & 0.015625 & 0.296875  & 0.468750   & 0.109375 & 0.015625 & 0.812500    & 0.015625  & 0.109375\\
        DE-GAN    & 0.078125 & 0.015625  & 0.015625  & 0.015625  & 0.468750  & 0.015625  & 0.015625  & 0.015625 & 0.375000    & 0.015625  & 0.015625  & 0.078125\\
        DocDiff   & 0.109375 & 0.015625  & 0.015625  & 0.015625 & 0.015625 & 0.156250   & 0.687500    & 0.296875 & 0.015625 & 0.078125  & 0.015625  & 0.031250 \\
        DocRes    & 0.031250 & 0.218750  & 0.078125  & 0.015625 & 0.015625 & 0.687500    & 0.937500    & 0.937500 & 0.015625 & 0.015625  & 0.015625  & 0.015625 \\
        GSDM      & 0.812500 & 0.031250  & 0.031250  & 0.015625 & 0.015625 & 0.031250   & 0.578125  & 0.031250 & 0.015625 & 0.156250   & 0.015625  & 0.078125 \\ \hline
        \hline
        \multirow{2}{*}{Method}&\multicolumn{4}{c||}{Char ACC (CRNN)} &\multicolumn{4}{c||}{Word ACC (CRNN)} &\multicolumn{4}{c}{Char ACC (MORAN)} \\
        \cline{2-13}
        & TDO (CNN)    & TDO (Trans)  & TDO (Mamba)   & TDO (GSDM) & TDO (CNN)    & TDO (Trans)  & TDO (Mamba)   & TDO (GSDM) & TDO (CNN)    & TDO (Trans)  & TDO (Mamba)   & TDO (GSDM)  \\ \hline
        AOT       & 0.015625 & 0.015625  & 0.015625  & 0.015625  & 0.015625 & 0.031250   & 0.015625  & 0.015625 & 0.015625 & 0.015625  & 0.015625  & 0.015625\\
        GRIG      & 0.015625 & 0.015625  & 0.015625  & 0.015625 & 0.015625 & 0.015625  & 0.015625  & 0.015625 & 0.015625 & 0.015625  & 0.015625  & 0.015625\\
        DE-GAN    & 0.031250  & 0.015625  & 0.015625  & 0.015625 & 0.015625 & 0.015625  & 0.015625  & 0.015625 & 0.031250  & 0.015625  & 0.015625  & 0.015625\\
        DocDiff   & 0.046875 & 0.031250   & 0.031250   & 0.015625 & 0.015625 & 0.015625  & 0.015625  & 0.015625 & 0.109375 & 0.046875  & 0.046875  & 0.015625\\
        DocRes    & 0.078125 & 0.109375  & 0.046875  & 0.015625 & 0.031250  & 0.156250   & 0.046875  & 0.031250 & 0.078125 & 0.031250   & 0.078125  & 0.015625\\
        GSDM      & 0.015625 & 0.015625  & 0.015625  & 0.015625 & 0.027708 & 0.031250   & 0.015625  & 0.015625 & 0.015625 & 0.015625  & 0.015625  & 0.015625\\ \hline
        \hline
        \multirow{2}{*}{Method}&\multicolumn{4}{c||}{Word ACC (MORAN)} &\multicolumn{4}{c||}{Char ACC (PaddleOCR)} &\multicolumn{4}{c}{Word ACC (PaddleOCR)} \\
        \cline{2-13}
        & TDO (CNN)    & TDO (Trans)  & TDO (Mamba)   & TDO (GSDM) & TDO (CNN)    & TDO (Trans)  & TDO (Mamba)   & TDO (GSDM) & TDO (CNN)    & TDO (Trans)  & TDO (Mamba)   & TDO (GSDM)  \\ \hline
        AOT       & 0.015625 & 0.015625  & 0.015625  & 0.015625 & 0.015625 & 0.015625  & 0.015625  & 0.015625   & 0.015625 & 0.015625  & 0.015625  & 0.015625 \\
        GRIG      & 0.015625 & 0.031250   & 0.015625  & 0.015625  & 0.015625 & 0.015625  & 0.031250   & 0.015625 & 0.031250  & 0.015625  & 0.015625  & 0.015625\\
        DE-GAN    & 0.015625 & 0.015625  & 0.015625  & 0.015625 & 0.015625 & 0.015625  & 0.015625  & 0.015625 & 0.015625 & 0.015625  & 0.015625  & 0.015625 \\
        DocDiff   & 0.015625 & 0.046875  & 0.015625  & 0.015625 & 0.015625 & 0.015625  & 0.015625  & 0.015625 & 0.015625 & 0.015625  & 0.015625  & 0.015625 \\
        DocRes    & 0.015625 & 0.027707 & 0.015625 & 0.015625 & 0.015625 & 0.015625  & 0.015625  & 0.015625 & 0.031250  & 0.015625  & 0.015625  & 0.015625 \\
        GSDM      & 0.043114 & 0.031250  & 0.031250  & 0.015625 & 0.015625 & 0.031250   & 0.031250   & 0.015625 & 0.015625 & 0.015625  & 0.027707  & 0.015625 \\ \hline
    \end{tabular}
    $}

    \caption{Wilcoxon signed-rank test results ($p$-values) for the quantitative results from Table~\ref{tab:patch_all_comparison_1}, Table~\ref{tab:patch_all_comparison_2}, Table~\ref{tab:patch_all_comparison_3}, Table~\ref{tab:patch_all_comparison_4}, Table~\ref{tab:patch_all_comparison_5}, Table~\ref{tab:patch_all_comparison_6}, and Table~\ref{tab:patch_all_comparison_7} across seven document image datasets. TDO: TextDoctor.}
    \label{tab:Wilcoxon_patch_all_comparison}
    \centering
    \resizebox{1.0\textwidth}{!}{$
    \begin{tabular}{c|c|c|c|c||c|c|c|c||c|c|c|c}
        \hline
        \multirow{2}{*}{Method}&\multicolumn{4}{c||}{PSNR} &\multicolumn{4}{c||}{SSIM} &\multicolumn{4}{c}{FID$_{\text{crop}}$} \\
        \cline{2-13}
        & TDO (CNN)    & TDO (Trans)  & TDO (Mamba)   & TDO (GSDM) & TDO (CNN)    & TDO (Trans)  & TDO (Mamba)   & TDO (GSDM) & TDO (CNN)    & TDO (Trans)  & TDO (Mamba)   & TDO (GSDM)  \\ \hline
        AOT       & 0.015625 & 0.015625  & 0.015625  & 0.015625 & 0.078125 & 0.015625  & 0.296875  & 0.015625  & 0.015625 & 0.015625  & 0.015625  & 0.687500 \\
        GRIG      & 0.015625 & 0.015625  & 0.015625  & 0.015625 & 0.015625 & 0.015625  & 0.015625  & 0.015625 & 0.578125 & 0.015625  & 0.015625  & 0.015625 \\
        DE-GAN    & 0.078125 & 0.015625  & 0.015625  & 0.015625 & 0.468750  & 0.015625  & 0.015625  & 0.015625 & 0.296875 & 0.015625  & 0.015625  & 0.015625\\
        DocDiff   & 0.015625 & 0.015625  & 0.015625  & 0.015625 & 0.015625 & 0.015625  & 0.015625  & 0.015625 & 0.015625 & 0.015625  & 0.015625  & 0.015625 \\
        DocRes    & 0.015625 & 0.015625  & 0.015625  & 0.015625 & 0.078125 & 0.031250   & 0.375000     & 0.015625 & 0.015625 & 0.015625  & 0.015625  & 0.218750\\
        GSDM      & 0.812500   & 0.031250   & 0.031250   & 0.015625 & 0.015625 & 0.031250   & 0.578125  & 0.031250 & 0.015625 & 0.015625  & 0.015625  & 0.078125\\ \hline
        \hline
        \multirow{2}{*}{Method}&\multicolumn{4}{c||}{Char ACC (CRNN)} &\multicolumn{4}{c||}{Word ACC (CRNN)} &\multicolumn{4}{c}{Char ACC (MORAN)} \\
        \cline{2-13}
        & TDO (CNN)    & TDO (Trans)  & TDO (Mamba)   & TDO (GSDM) & TDO (CNN)    & TDO (Trans)  & TDO (Mamba)   & TDO (GSDM) & TDO (CNN)    & TDO (Trans)  & TDO (Mamba)   & TDO (GSDM)  \\ \hline
        AOT       & 0.015625 & 0.015625  & 0.015625  & 0.015625  & 0.015625 & 0.015625  & 0.015625  & 0.015625 & 0.015625 & 0.015625  & 0.015625  & 0.015625\\
        GRIG      & 0.015625 & 0.015625  & 0.015625  & 0.015625 & 0.015625 & 0.015625  & 0.015625  & 0.015625 & 0.015625 & 0.015625  & 0.015625  & 0.015625\\
        DE-GAN    & 0.015625 & 0.015625  & 0.015625  & 0.015625 & 0.015625 & 0.031250   & 0.015625  & 0.015625 & 0.015625 & 0.015625  & 0.015625  & 0.015625\\
        DocDiff   & 0.015625 & 0.015625  & 0.015625  & 0.015625 & 0.015625 & 0.015625  & 0.015625  & 0.015620  & 0.015625 & 0.015625  & 0.015625  & 0.015625\\
        DocRes    & 0.015625 & 0.015625  & 0.015625  & 0.015625 & 0.015625 & 0.015625  & 0.015625  & 0.015625 & 0.015625 & 0.015625  & 0.015625  & 0.015625 \\
        GSDM      & 0.015625 & 0.015625  & 0.015625  & 0.015625  & 0.027707 & 0.031250  & 0.027707 & 0.015625& 0.015625 & 0.015625  & 0.015625  & 0.015625 \\ \hline
        \hline
        \multirow{2}{*}{Method}&\multicolumn{4}{c||}{Word ACC (MORAN)} &\multicolumn{4}{c||}{Char ACC (PaddleOCR)} &\multicolumn{4}{c}{Word ACC (PaddleOCR)} \\
        \cline{2-13}
        & TDO (CNN)    & TDO (Trans)  & TDO (Mamba)   & TDO (GSDM) & TDO (CNN)    & TDO (Trans)  & TDO (Mamba)   & TDO (GSDM) & TDO (CNN)    & TDO (Trans)  & TDO (Mamba)   & TDO (GSDM)  \\ \hline
        AOT       & 0.031250  & 0.027707  & 0.015625  & 0.015625 & 0.015625 & 0.015625  & 0.015625  & 0.015625 & 0.015625 & 0.015625  & 0.015625  & 0.015625 \\
        GRIG      & 0.015625 & 0.015625  & 0.015625  & 0.015625 & 0.015625 & 0.015625  & 0.015625  & 0.015625  & 0.015625 & 0.015625  & 0.015625  & 0.015625\\
        DE-GAN    & 0.027708 & 0.015625  & 0.015625  & 0.015625  & 0.015625 & 0.015625  & 0.015625  & 0.015625 & 0.015625 & 0.015625  & 0.015625  & 0.015625\\
        DocDiff   & 0.015625 & 0.015625  & 0.015625  & 0.015625 & 0.015625 & 0.015625  & 0.015625  & 0.015625 & 0.015625 & 0.015625  & 0.015625  & 0.015625 \\
        DocRes    & 0.015625 & 0.015625  & 0.015625  & 0.015625 & 0.015625 & 0.015625  & 0.015625  & 0.015625 & 0.015625 & 0.015625  & 0.015625  & 0.015625 \\
        GSDM      & 0.043114 & 0.031250   & 0.031250   & 0.015625 & 0.015625 & 0.031250   & 0.031250   & 0.015625 & 0.015625 & 0.015625  & 0.027707 & 0.015625\\ \hline
    \end{tabular}
    $}
\end{table*}

\subsection{More comparisons to state-of-the-art methods}

\textbf{More quantitative comparisons.} 
More quantitative comparisons to the state-of-the-art methods for the other six datasets (BCSD, Twitter posts, Twitter profiles, activity diagrams, signatures, and tabular documents) are provided in Table~\ref{tab:ori_comparison_all_1}, Table~\ref{tab:ori_comparison_all_2}, Table~\ref{tab:ori_comparison_all_3}, Table~\ref{tab:ori_comparison_all_4}, Table~\ref{tab:ori_comparison_all_5}, and Table~\ref{tab:ori_comparison_all_6}, respectively.
Note that the quantitative results for the FUNSD dataset were presented in the main paper, Table 1.



We evaluated document image inpainting performance when all methods were trained on text image patches. 
As shown in Table~\ref{tab:patch_all_comparison_1}, Table~\ref{tab:patch_all_comparison_2}, Table~\ref{tab:patch_all_comparison_3}, Table~\ref{tab:patch_all_comparison_4}, Table~\ref{tab:patch_all_comparison_5}, Table~\ref{tab:patch_all_comparison_6}, and Table~\ref{tab:patch_all_comparison_7} after training on the same text image patch dataset, the inpainting performance of the compared methods was assessed on seven document image datasets (FUNSD, BCSD, Twitter posts, Twitter profiles, activity diagrams, signatures, and tabular documents). 
The results reveal that state-of-the-art methods struggle to achieve optimal results due to domain gaps between the trained image patches and high-resolution document images. 

In contrast, TextDoctor's patch-based inference partitions documents into patches, maintaining consistent visual patterns between training and inference. 
Our structure pyramid prediction captures global and local structures through multiscale inputs, while the patch pyramid diffusion model leverages this information for high-quality inpainting.

\textbf{Wilcoxon signed-rank test.} 
To evaluate the performance differences between our TextDoctor variants and compared methods, we conducted the Wilcoxon signed-rank test using PSNR, SSIM, FID$_{\text{crop}}$, Char Acc (CRNN), Word ACC (CRNN), Char Acc (MORAN), Word ACC (MORAN), Char Acc (PaddleOCR), and Word ACC (PaddleOCR) scores across the seven datasets.
Since most U-IDS$_{\text{crop}}$ scores for the compared methods are zeros, we were unable to calculate $p$-values for this metric.
For each metric, we collected quantitative values from these datasets, resulting in seven paired observations per comparison.
We then analyzed the results from Table 1 of the main paper, Table~\ref{tab:ori_comparison_all_1}, Table~\ref{tab:ori_comparison_all_2}, Table~\ref{tab:ori_comparison_all_3}, Table~\ref{tab:ori_comparison_all_4}, Table~\ref{tab:ori_comparison_all_5}, and Table~\ref{tab:ori_comparison_all_6} as well as Table~\ref{tab:patch_all_comparison_1}, Table~\ref{tab:patch_all_comparison_2}, Table~\ref{tab:patch_all_comparison_3}, Table~\ref{tab:patch_all_comparison_4}, Table~\ref{tab:patch_all_comparison_5}, Table~\ref{tab:patch_all_comparison_6}, and Table~\ref{tab:patch_all_comparison_7} to calculate the $p$-values, as shown in Table~\ref{tab:Wilcoxon_ori_comparison_all} and Table~\ref{tab:Wilcoxon_patch_all_comparison}, respectively.
Most of the $p$-value scores are below than 0.05, indicating a statistically significant difference in performance between our TextDoctor variants and the compared methods at the 0.05 significance level.

\begin{figure*}[!ht]
	\centering
	\includegraphics[width=1.0\textwidth]{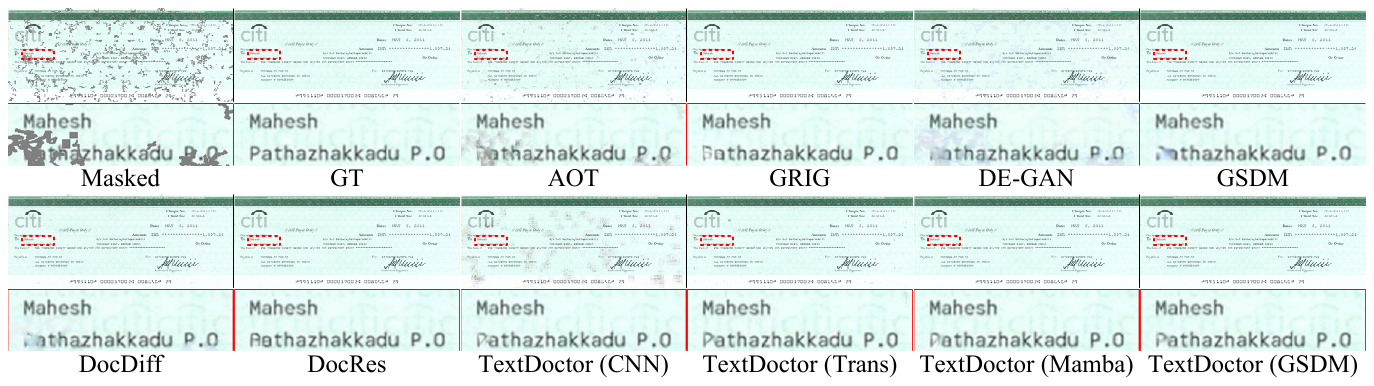}
 	\includegraphics[width=1.0\textwidth]{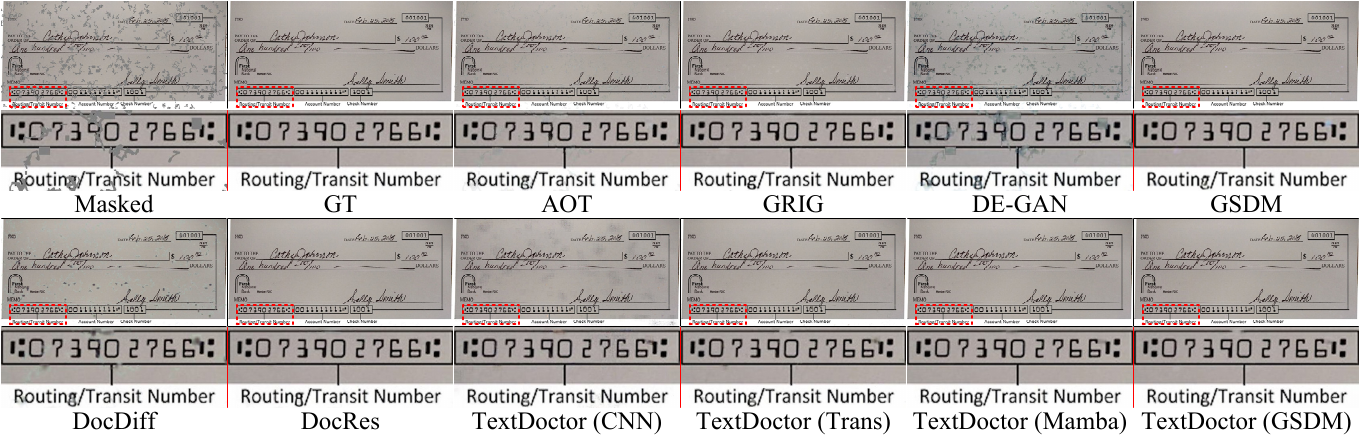}
	\caption{Visual comparison to the SOTA text document inpainting and restoration methods on the BCSD dataset. }\label{fig:fig_vis_compare_BCSD}  

	\includegraphics[width=1.0\textwidth]{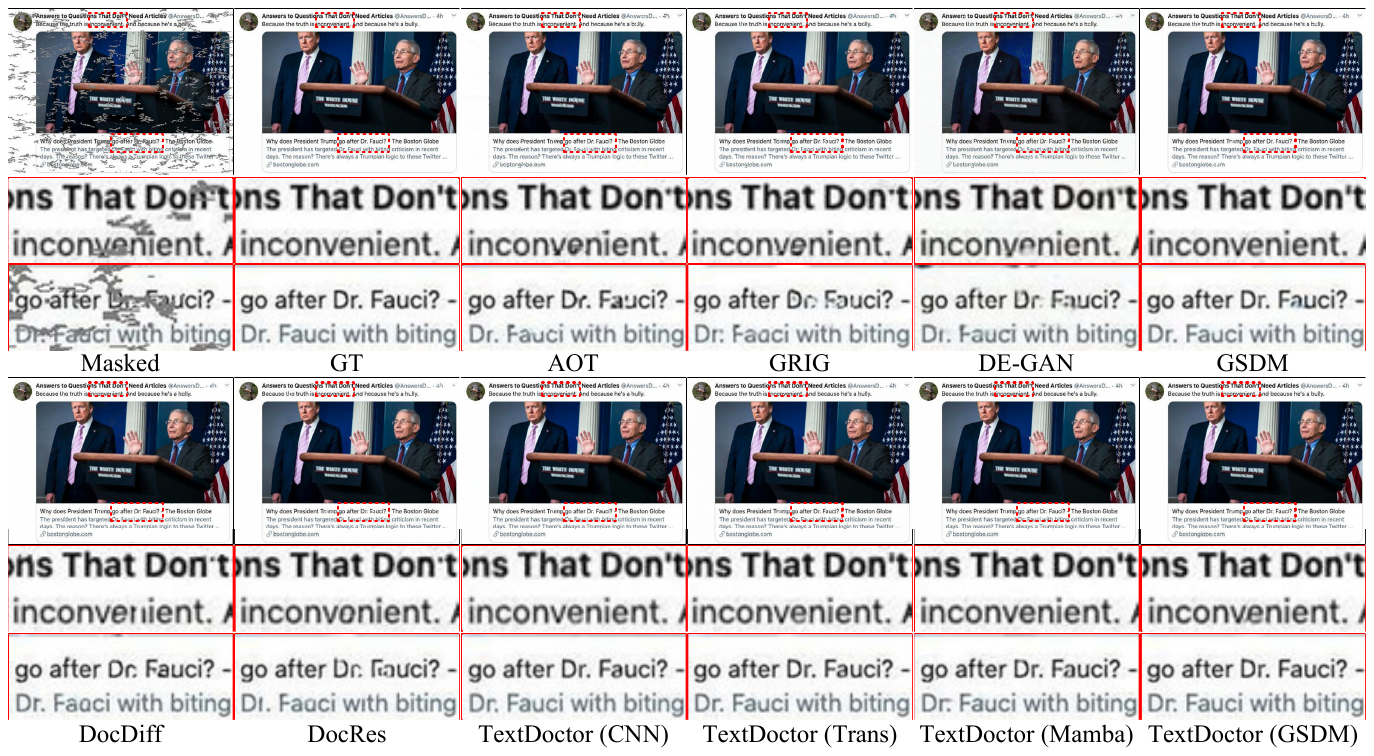}
	\caption{Visual comparison to the SOTA text document inpainting and restoration methods on the Twitter profiles dataset. }\label{fig:fig_vis_compare_tweet_profiles}  
\end{figure*}
\begin{figure*}[!ht]
	\centering
	\includegraphics[width=1.0\textwidth]{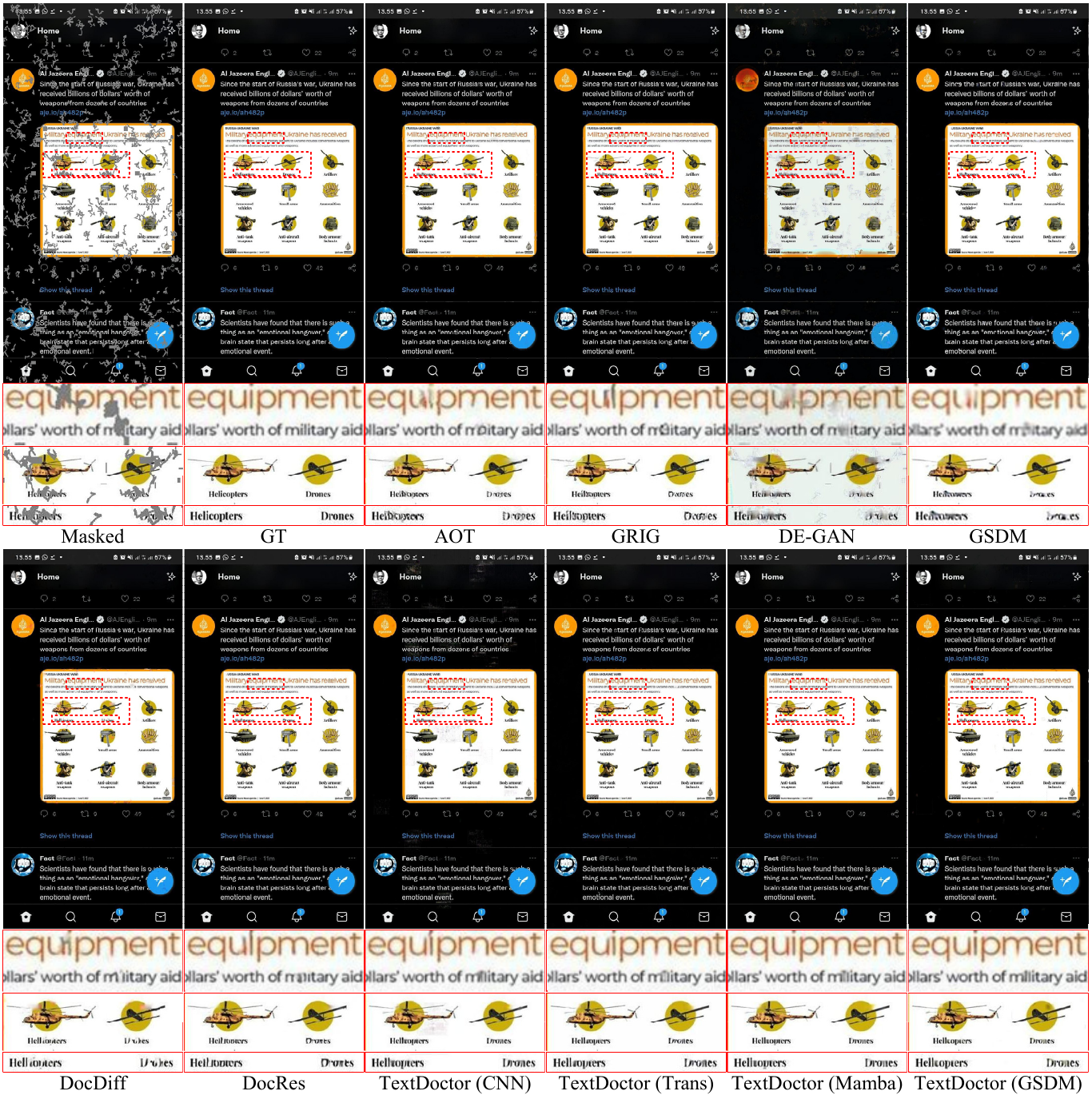}
	\caption{Visual comparison to the SOTA text document inpainting and restoration methods on the Twitter posts dataset. }\label{fig:fig_vis_compare_tweet_post}  
\end{figure*}
\begin{figure*}[!ht]
	\centering
	\includegraphics[width=1.0\textwidth]{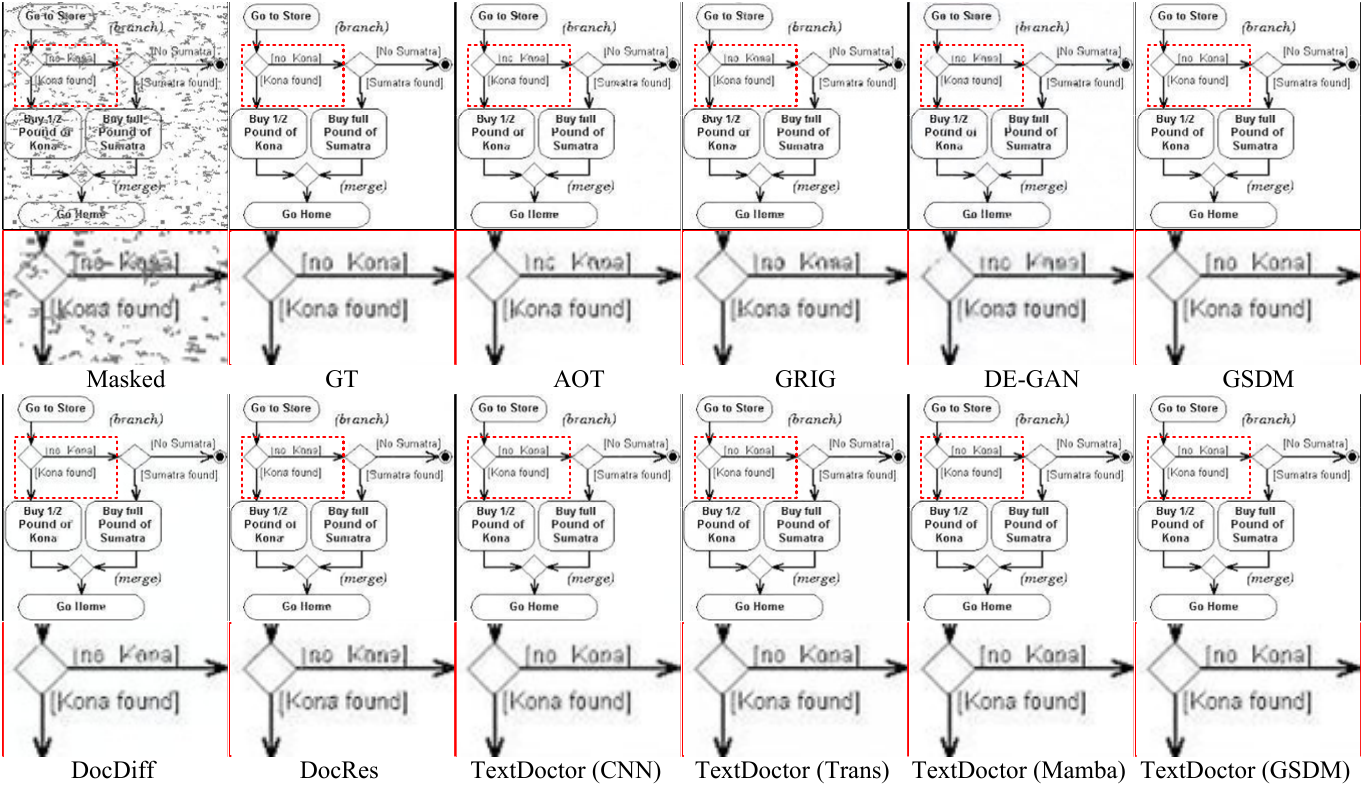}
 	\includegraphics[width=1.0\textwidth]{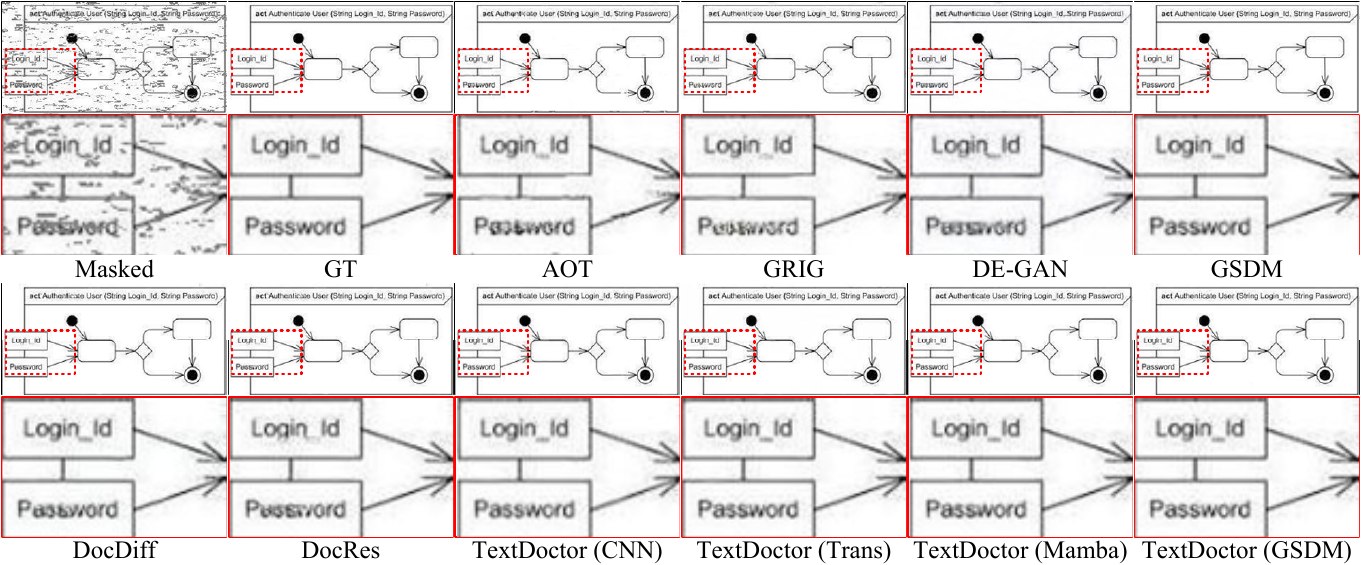}
	\caption{Visual comparison to the SOTA text document inpainting and restoration methods on the activity diagrams dataset. }\label{fig:fig_vis_compare_Diagram}  
\end{figure*}
\begin{figure*}[!ht]
	\centering
	\includegraphics[width=1.0\textwidth]{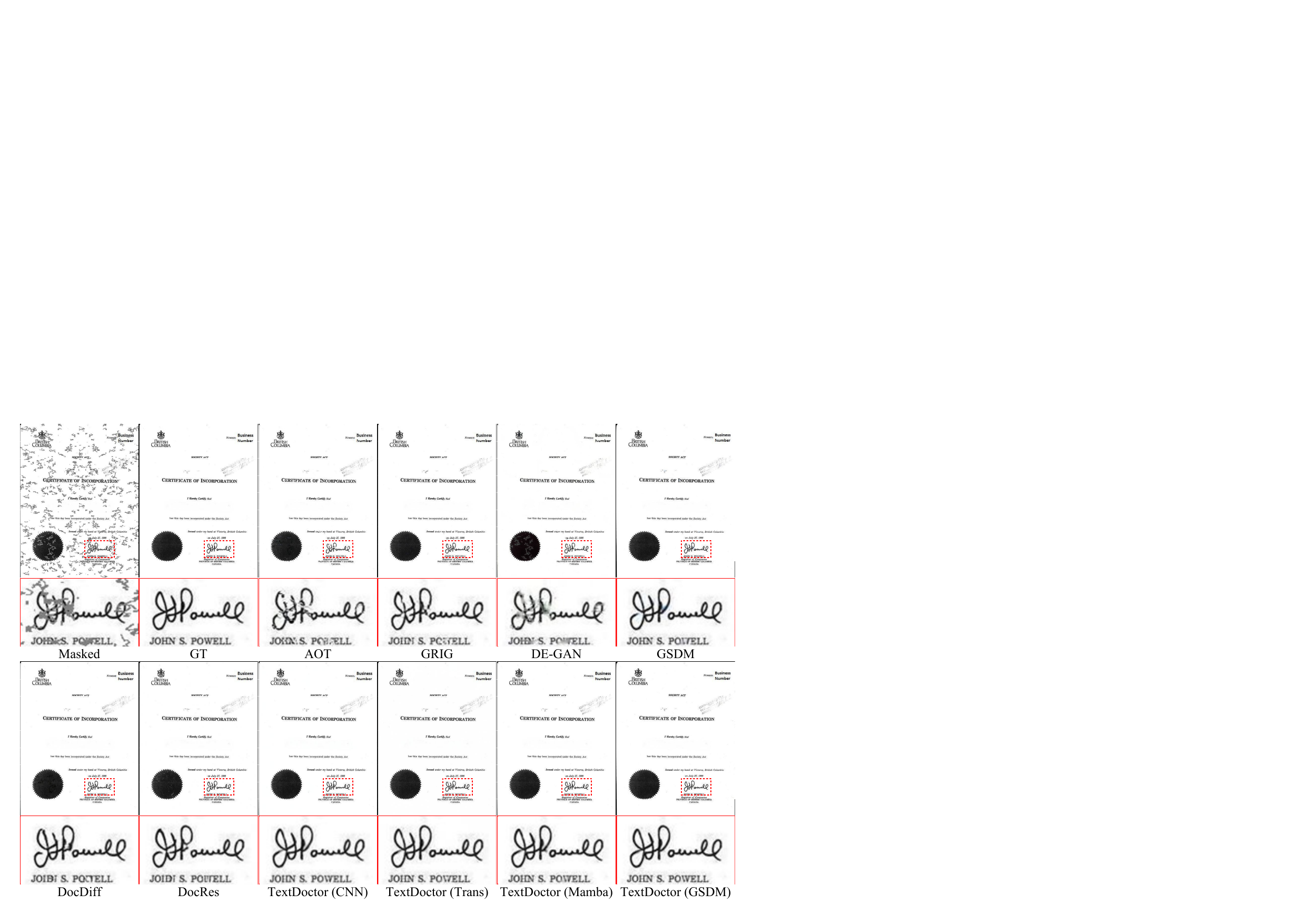}
 	\includegraphics[width=1.0\textwidth]{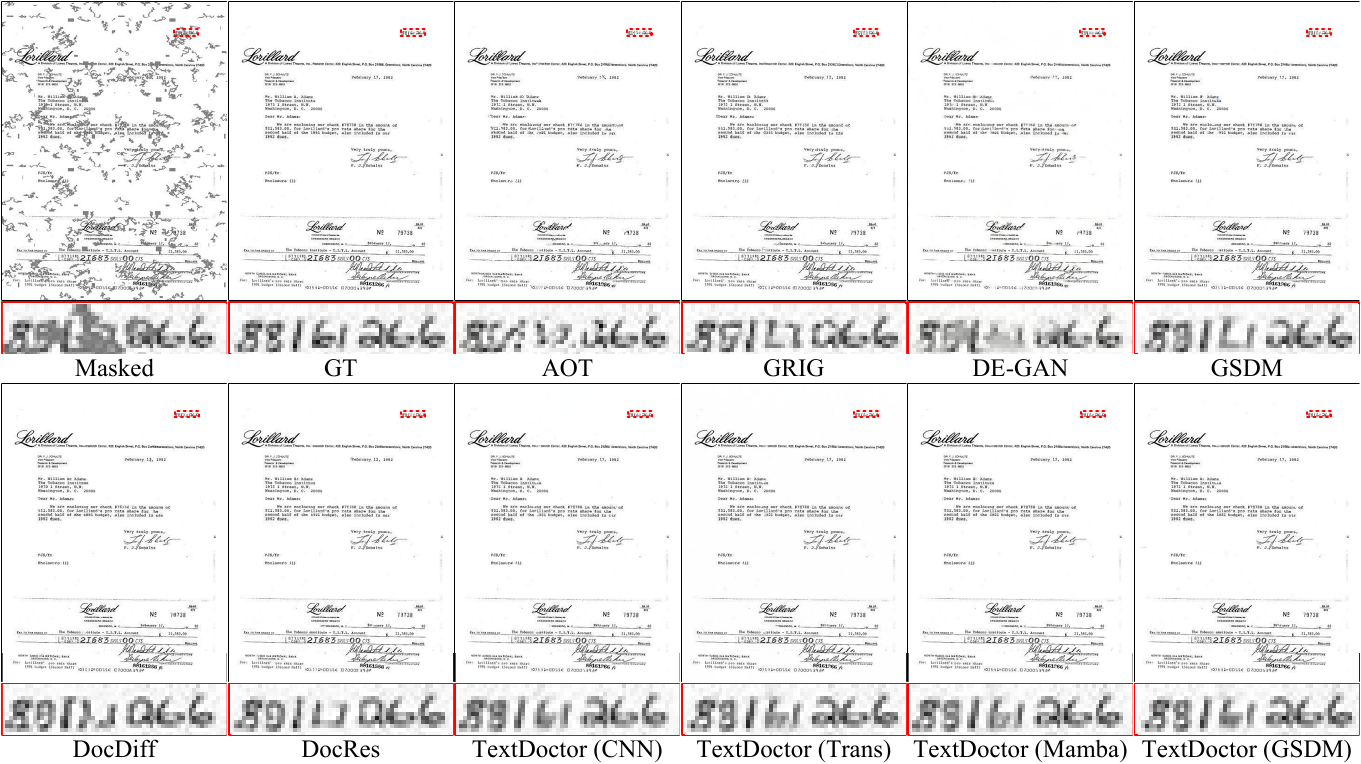}
	\caption{Visual comparison to the SOTA text document inpainting and restoration methods on the signatures dataset. }\label{fig:fig_vis_compare_signature}  
\end{figure*}
\begin{figure*}[!ht]
	\centering
	\includegraphics[width=1.0\textwidth]{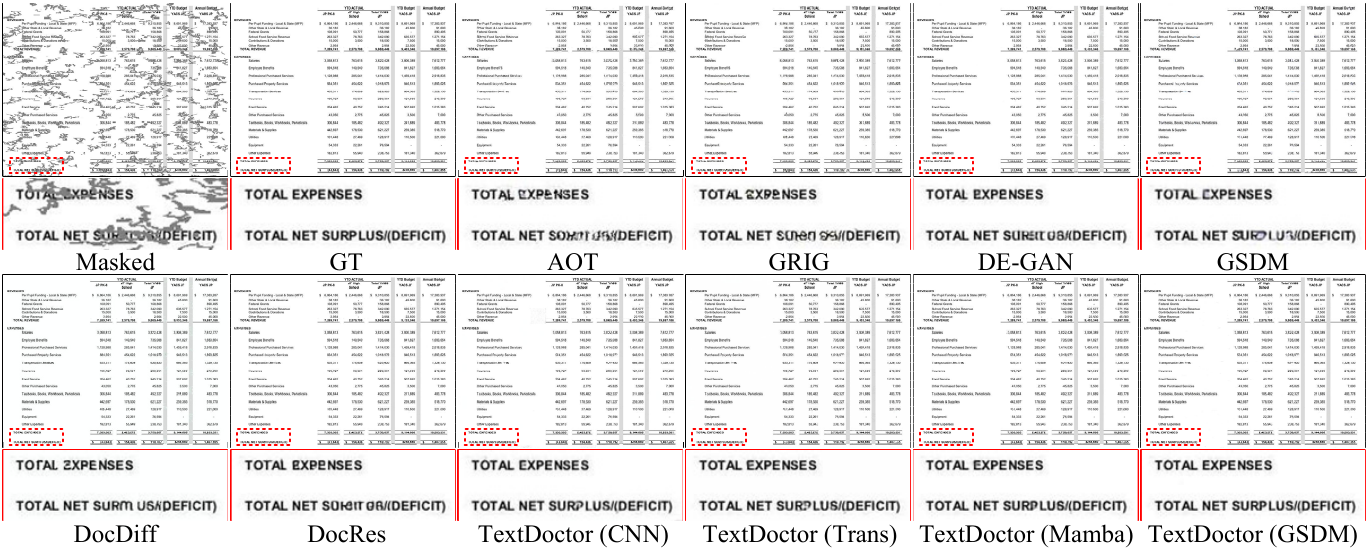}
 	\includegraphics[width=1.0\textwidth]{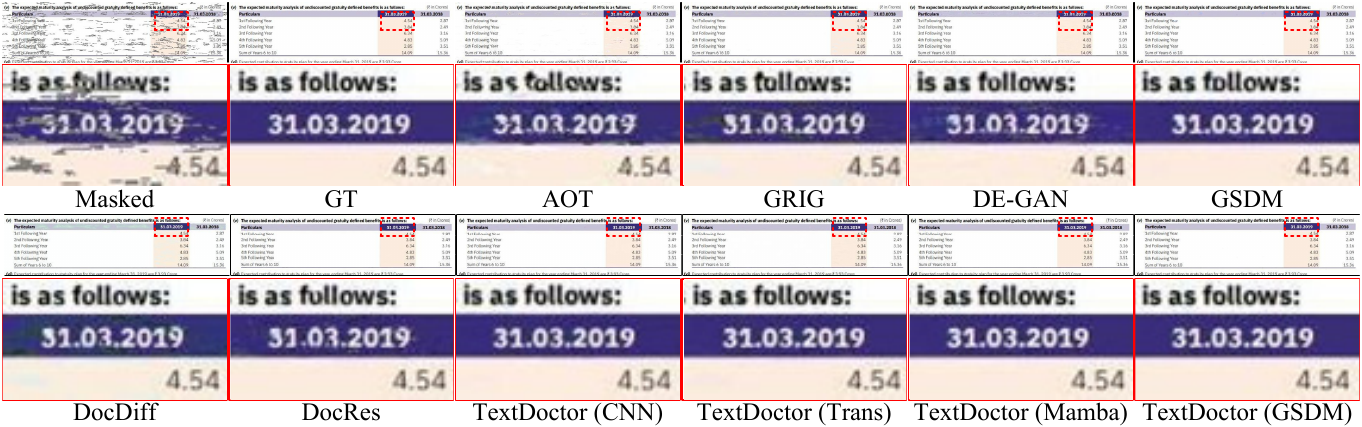}
   	\includegraphics[width=1.0\textwidth]{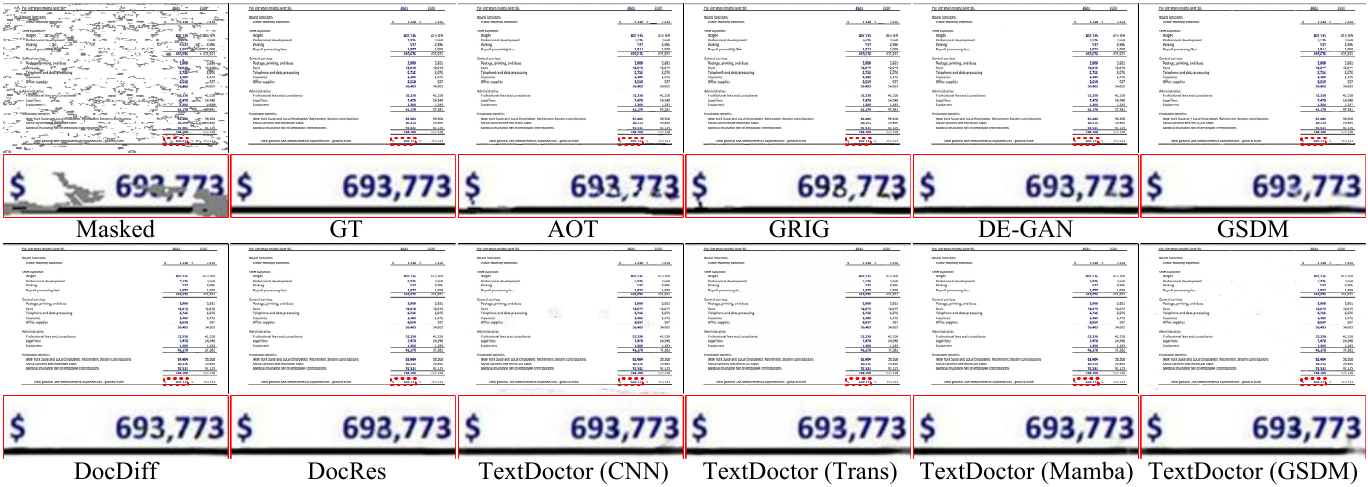}
	\caption{Visual comparison to the SOTA text document inpainting and restoration methods on the tabular dataset. }\label{fig:fig_vis_compare_tabular}  
\end{figure*}

\textbf{More qualitative comparisons.} 
More qualitative comparisons to state-of-the-art methods are shown in Fig.~\ref{fig:fig_vis_compare_BCSD}, Fig.~\ref{fig:fig_vis_compare_tweet_profiles}, 
 Fig.~\ref{fig:fig_vis_compare_tweet_post},  Fig.~\ref{fig:fig_vis_compare_Diagram}, Fig.~\ref{fig:fig_vis_compare_signature}, and Fig.~\ref{fig:fig_vis_compare_tabular}.



\begin{figure*}[!ht]
	\centering
	\includegraphics[width=1.0\textwidth]{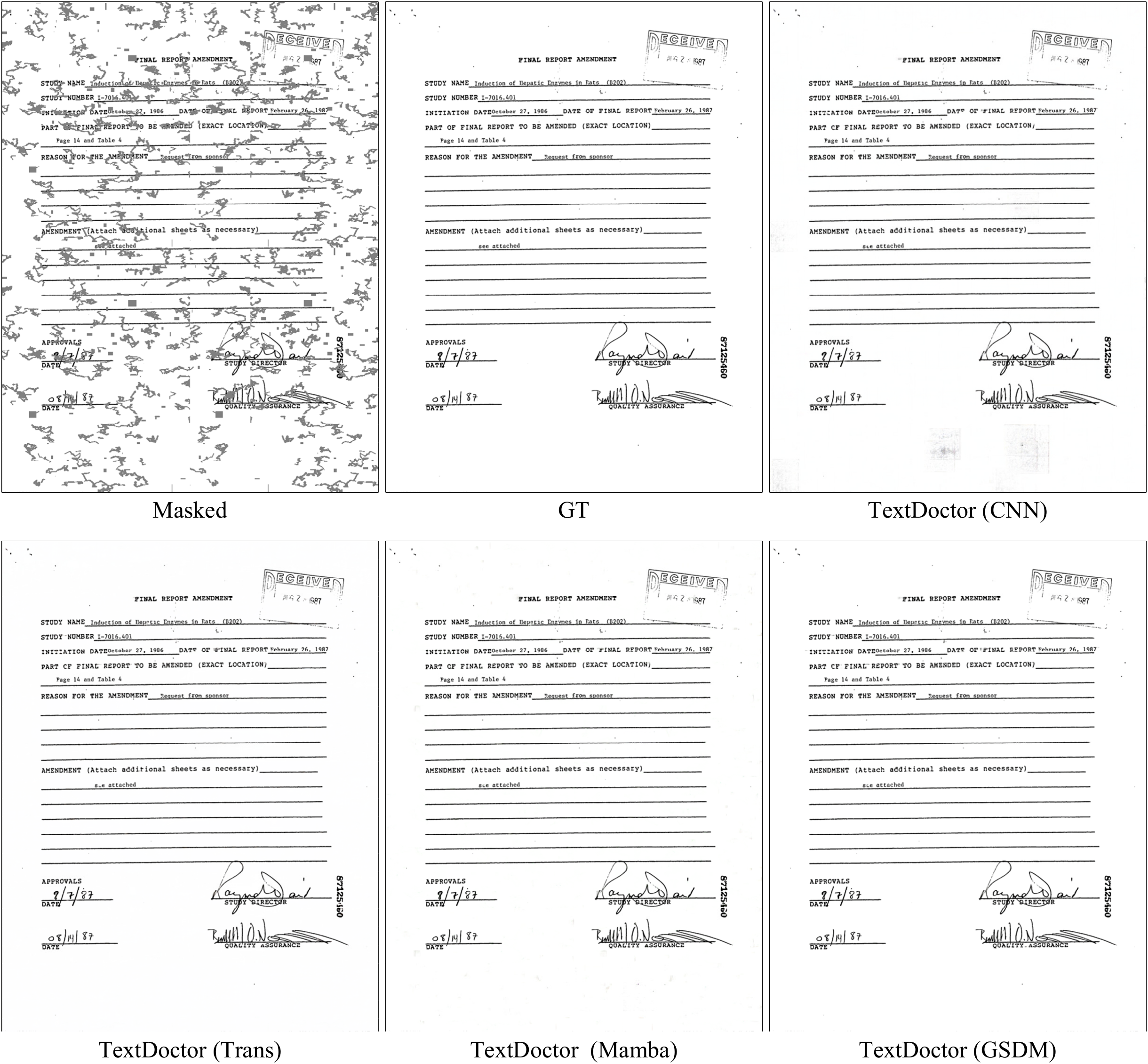}
	\caption{More visual results of TextDoctor on the FUNSD dataset (zoom in for a better review). }\label{fig:fig_vis_our_full_FUNSD}  
\end{figure*}

\begin{figure*}[!ht]
	\centering
	\includegraphics[width=1.0\textwidth]{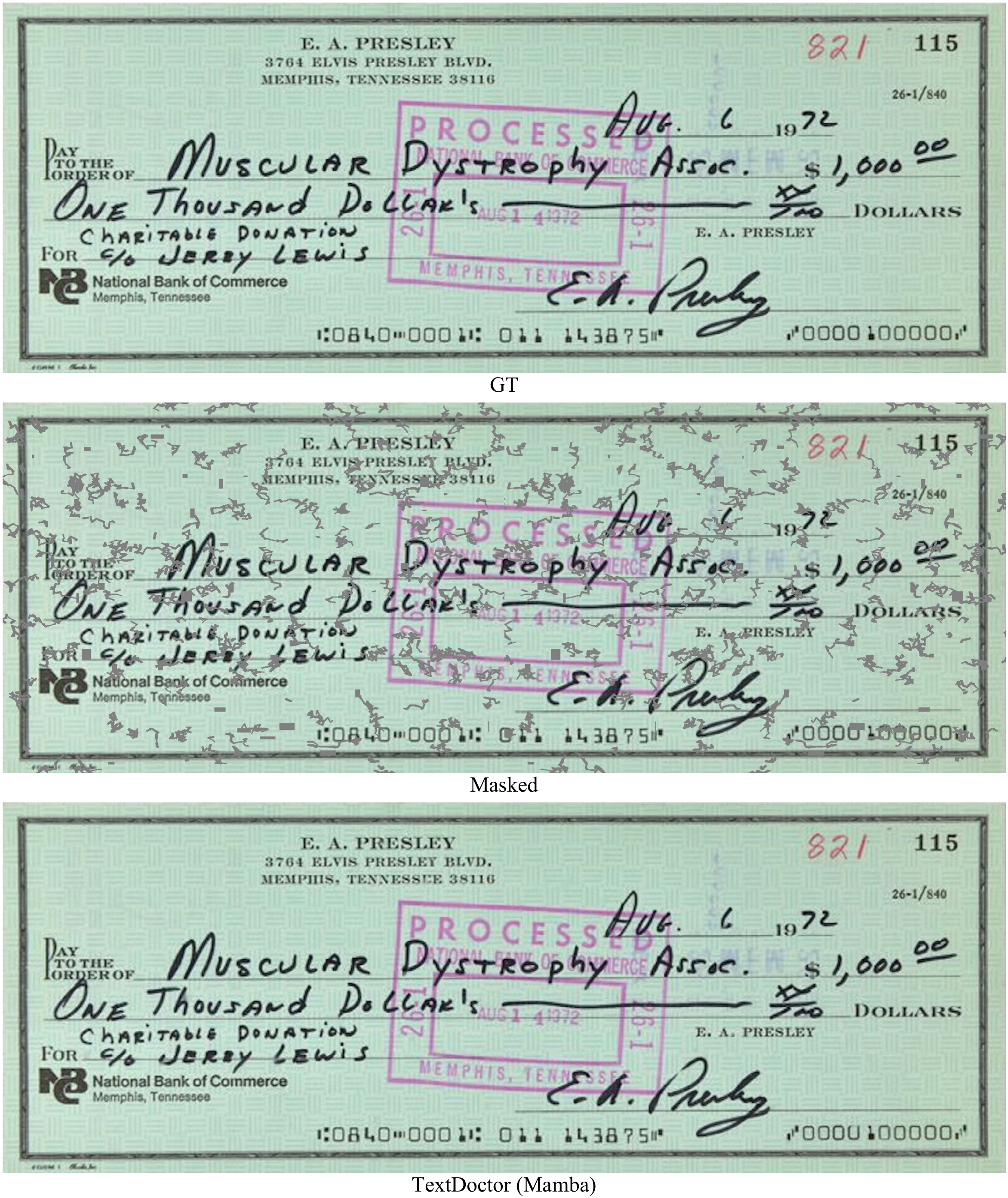}
	\caption{More visual results of TextDoctor on the BCSD dataset (zoom in for a better review). }\label{fig:fig_vis_our_full_BCSD}  
\end{figure*}

\begin{figure*}[!ht]
	\centering
	\includegraphics[width=1.0\textwidth]{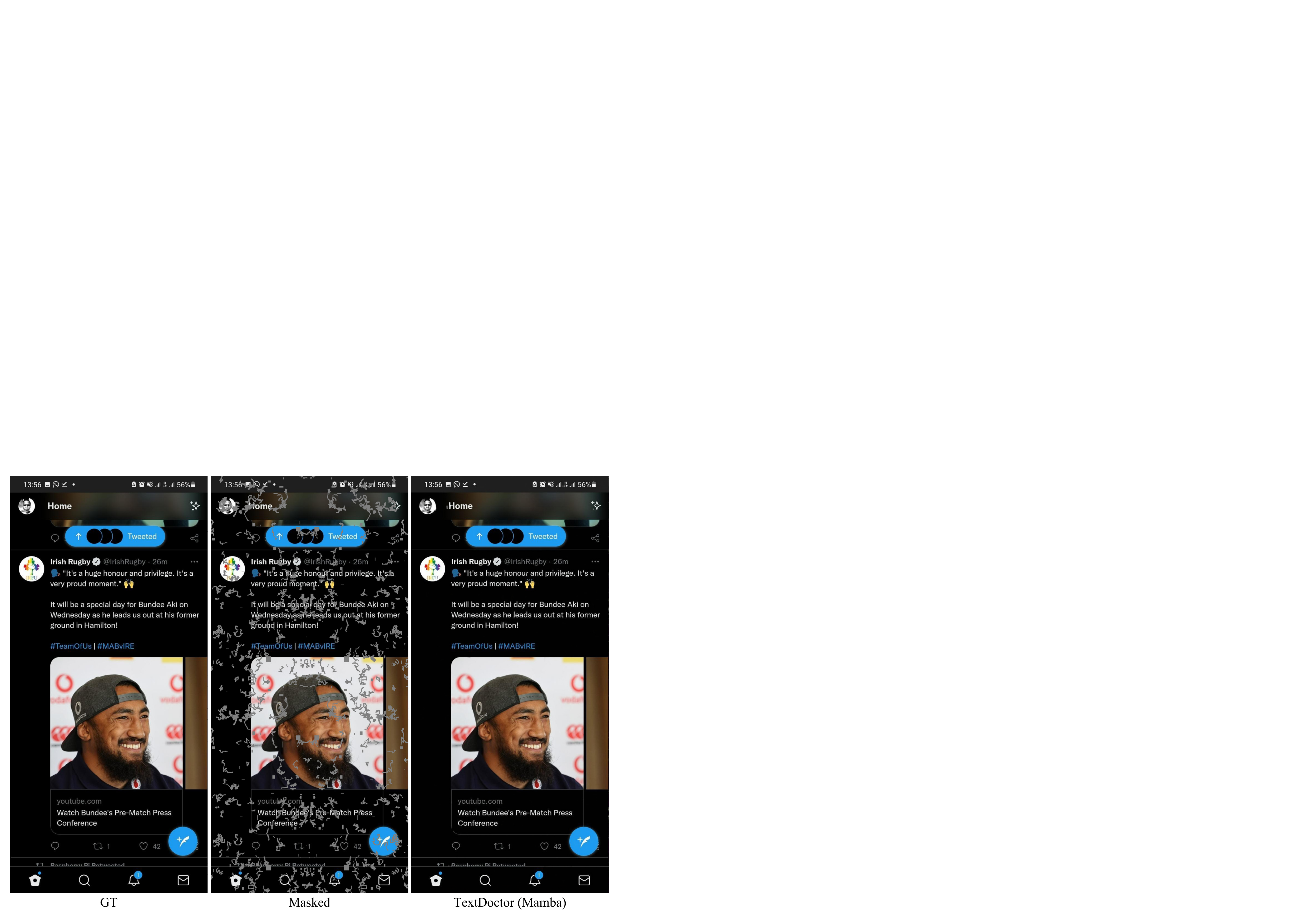}
	\caption{More visual results of TextDoctor on the Twitter posts dataset (zoom in for a better review). }\label{fig:fig_vis_our_full_tweet_post}  
\end{figure*}

\begin{figure*}[!ht]
	\centering
	\includegraphics[width=1.0\textwidth]{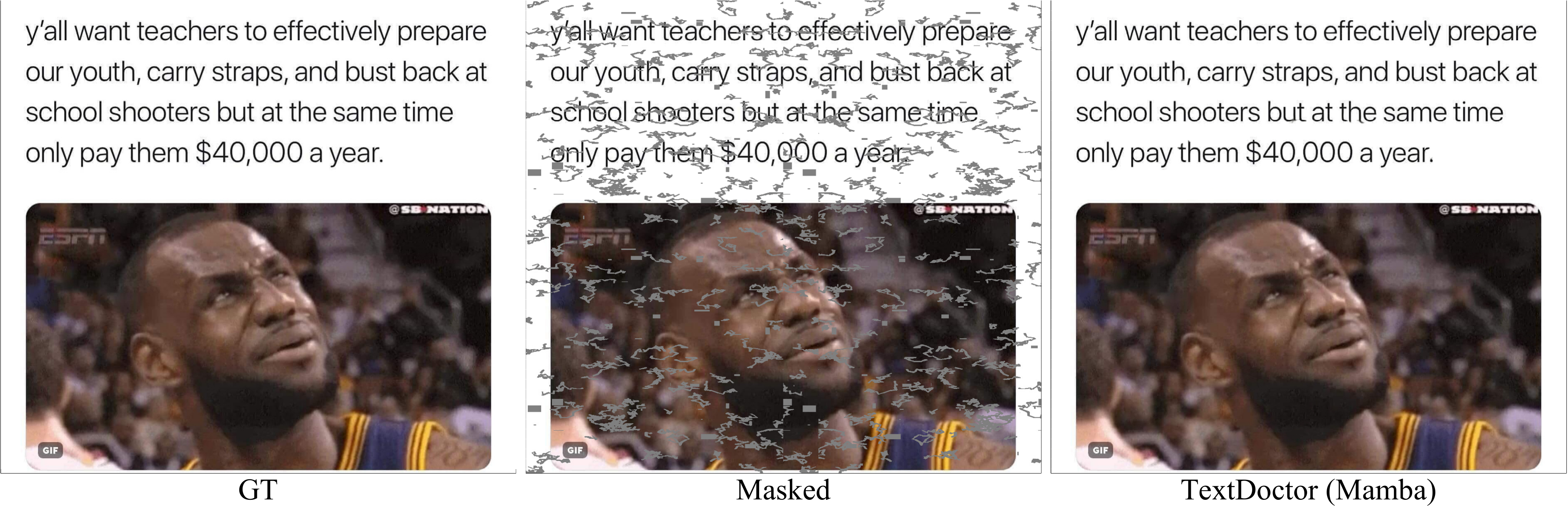}
	\caption{More visual results of TextDoctor on the Twitter profiles dataset (zoom in for a better review). }\label{fig:fig_vis_our_full_tweet_profile}  
\end{figure*}

\begin{figure*}[!ht]
	\centering
	\includegraphics[width=0.94\textwidth]{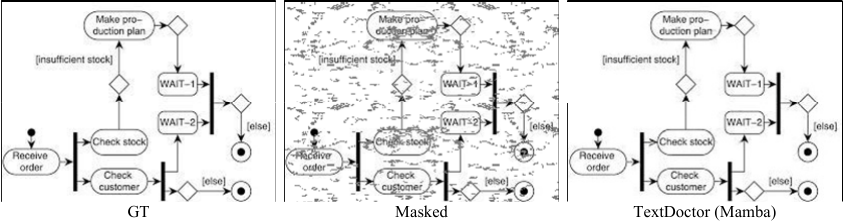}
	\caption{More visual results of TextDoctor on the activity diagrams dataset (zoom in for a better review). }\label{fig:fig_vis_our_full_Diagram}  
	\includegraphics[width=0.94\textwidth]{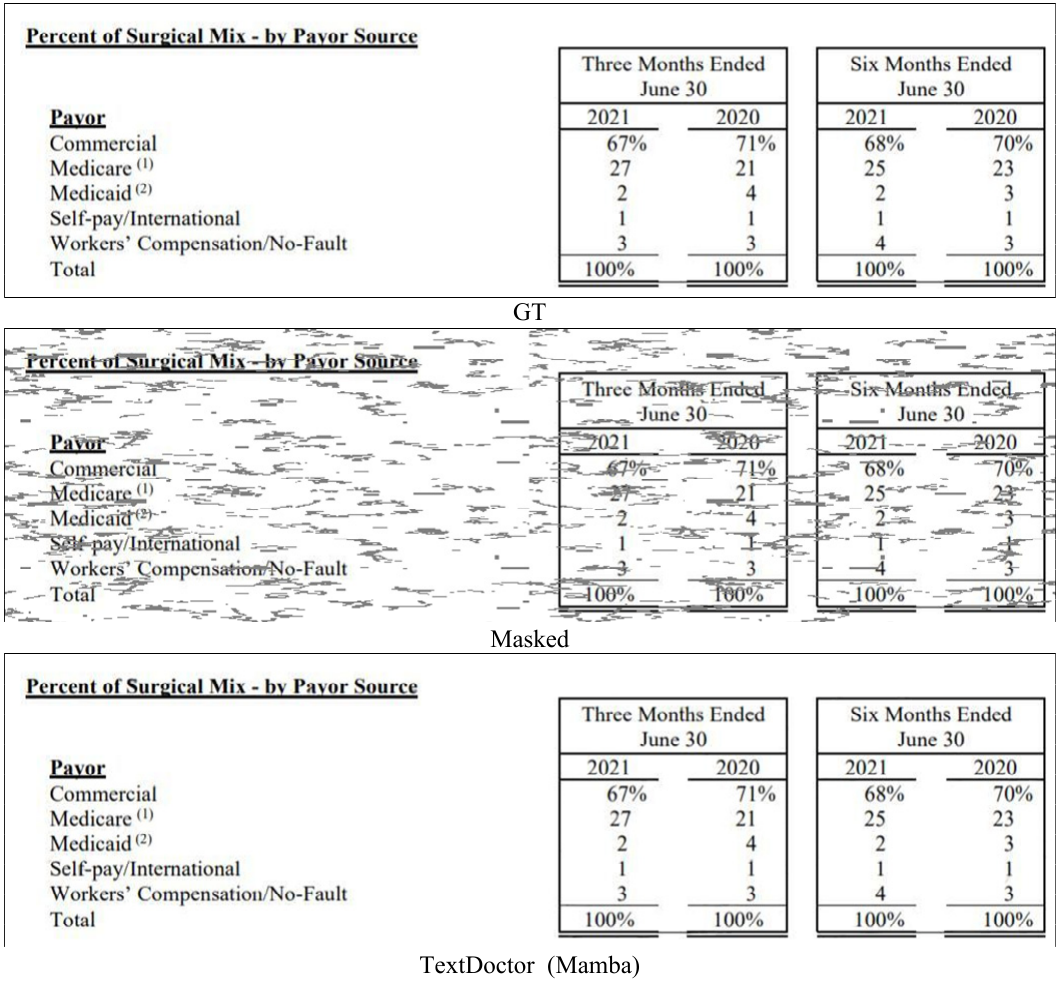}
	\caption{More visual results of TextDoctor on the tabular dataset (zoom in for a better review). }\label{fig:fig_vis_our_full_tabular}  
\end{figure*}

\begin{figure*}[!ht]
	\centering
	\includegraphics[width=1.0\textwidth]{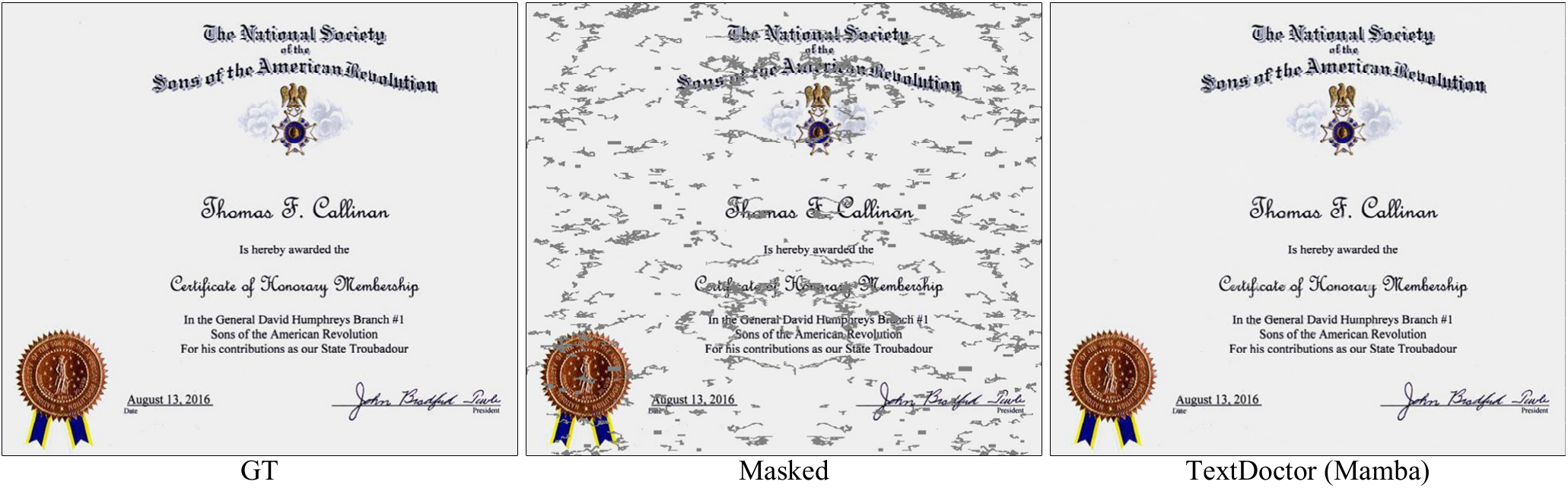}
 	\includegraphics[width=1.0\textwidth]{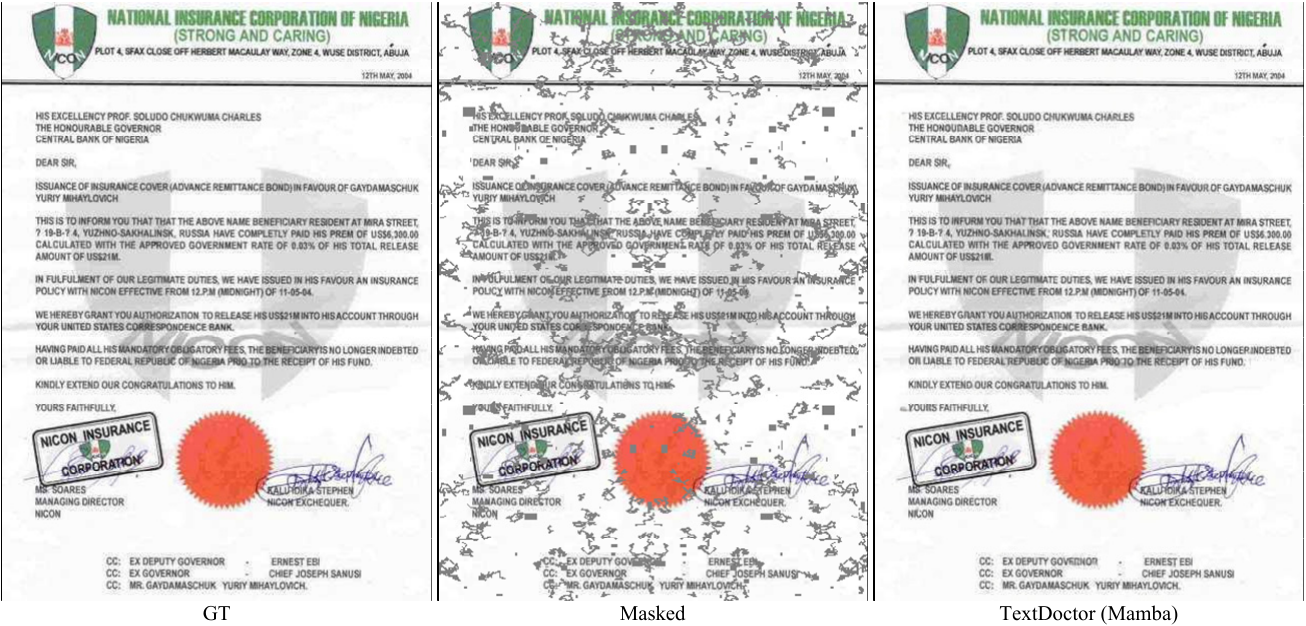}
   	\includegraphics[width=1.0\textwidth]{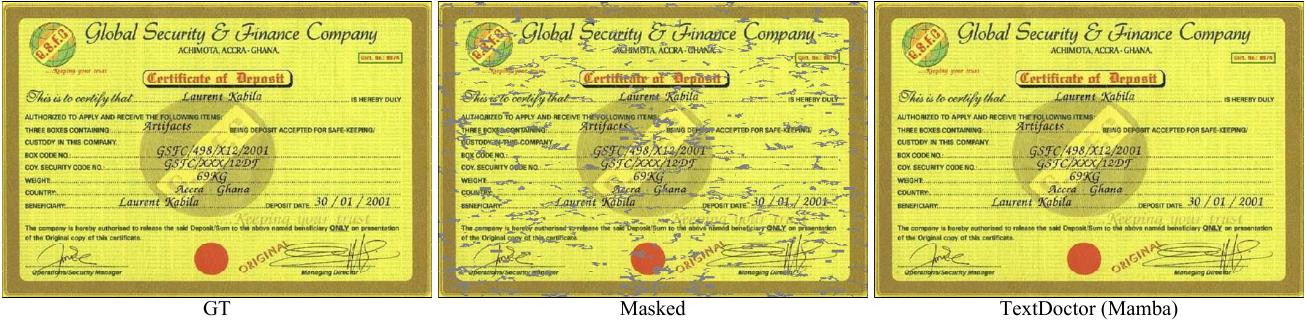}
	\caption{More visual results of TextDoctor on the signatures dataset (zoom in for a better review). }\label{fig:fig_vis_our_full_signature}  
\end{figure*}

\textbf{More visual results of TextDoctor.} 
More visual results of TextDoctor are shown in Fig.~\ref{fig:fig_vis_our_full_FUNSD}, Fig.~\ref{fig:fig_vis_our_full_BCSD}, Fig.~\ref{fig:fig_vis_our_full_tweet_post},  Fig.~\ref{fig:fig_vis_our_full_tweet_profile}, Fig.~\ref{fig:fig_vis_our_full_Diagram}, Fig.~\ref{fig:fig_vis_our_full_tabular}, and Fig.~\ref{fig:fig_vis_our_full_signature}.
Our TextDoctor performs impressively on these unseen document images.

\end{document}